\documentclass[letterpaper]{article} % DO NOT CHANGE THIS
\usepackage{aaai2026}  % DO NOT CHANGE THIS
\usepackage{times}  % DO NOT CHANGE THIS
\usepackage{helvet}  % DO NOT CHANGE THIS
\usepackage{courier}  % DO NOT CHANGE THIS
\usepackage[hyphens]{url}  % DO NOT CHANGE THIS
\usepackage{graphicx} % DO NOT CHANGE THIS
\usepackage{natbib}  % DO NOT CHANGE THIS AND DO NOT ADD ANY OPTIONS TO IT
\usepackage{caption} % DO NOT CHANGE THIS AND DO NOT ADD ANY OPTIONS TO IT
\usepackage{algorithm}
\usepackage{algorithmic}
\usepackage{amsmath}
\usepackage{amssymb}
\usepackage{amsfonts}
\usepackage{amsthm}
\usepackage{booktabs}
\usepackage{amsmath}
\usepackage{amssymb}
\usepackage{multirow}
\usepackage{siunitx}

\usepackage[table]{xcolor}
\usepackage{rotating}
\providecommand{\texorpdfstring}[2]{#1}   % hyperref is forbidden; supply fallback
\newtheorem{assumption}{Assumption}
\newtheorem{theorem}{Theorem}
\newtheorem{proposition}{Proposition}
\newtheorem{corollary}{Corollary}
\newtheorem{definition}{Definition}
\newcommand{\Ebb}{\mathbb{E}}
\newcommand{\Var}{\mathrm{Var}}
\newcommand{\Kbase}{K^{0}}
\newcommand{\Kdoob}{K^{C}}
\newcommand{\Kguid}{K^{g}}

\newcommand{\qbase}{q^{0}}
\newcommand{\qguid}{q^{g}}
\newcommand{\astar}{a^{\!*}}
\newcommand{\Vval}{V}

\newcommand{\Ssw}{\mathcal{S}_{\mathrm{sw}}}
\usepackage{newfloat}
\usepackage{listings}
\DeclareCaptionStyle{ruled}{labelfont=normalfont,labelsep=colon,strut=off} % DO NOT CHANGE THIS
\floatstyle{ruled}
\newfloat{listing}{tb}{lst}{}
\floatname{listing}{Listing}
\title{Commitment Before Realization: When Classifier-Free Guidance Becomes Unnecessary in Masked Diffusion Language Models}

\author{
Fan Zhou\textsuperscript{\rm 1},
Weitian Wang\textsuperscript{\rm 2},
Tim Van de Cruys\textsuperscript{\rm 1}
}

\affiliations{
\textsuperscript{\rm 1}KU Leuven, Belgium\\
\textsuperscript{\rm 2}Ruhr University Bochum, Germany\\
fan.zhou@kuleuven.be,
weitian.wang@edu.ruhr-uni-bochum.de,
tim.vandecruys@kuleuven.be
}

\usepackage{bibentry}
\begin{document}
\nocopyright
\maketitle

\begin{abstract}
% Classifier-free guidance (CFG) is typically used throughout masked diffusion language model decoding, although its usefulness varies across prompts and over time. To characterize when guidance is actually needed, we introduce a committor-based framework for measuring guidance necessity. For a partial output, a committor is the probability of eventually satisfying the target constraint under a specified continuation policy; comparing this probability under continued CFG and base-only continuation quantifies the remaining value of guidance. At the prompt level, these continuation values show that many prompts already succeed without CFG, while for others guidance provides no measurable benefit or is harmful. For prompts that initially benefit from CFG,
% necessity is also temporal. We define the commitment horizon $a^*$ as the earliest point at which switching all remaining decoding to the base decoder reduces final success by no more than a prespecified tolerance. Among prompts with a measurable commitment horizon, estimated horizons are usually early, revealing commitment before realization: terminal success is preserved under base continuation within the prescribed tolerance while many tokens remain unresolved. This boundary supports turning off CFG and localizes a region in which the additional success cost of increased decoding parallelism becomes small.
Classifier-free guidance (CFG) is usually kept on throughout masked diffusion language model decoding, although its benefit varies across prompts and over time. We study when CFG is actually needed by comparing, from any partial output, the probability of eventual constraint satisfaction under continued CFG and under base-only continuation. Their difference defines the remaining value of guidance. Guidance dependence is highly prompt-specific. Many prompts already succeed without CFG, while for others it provides no measurable benefit or can be harmful. For prompts that do benefit, the gain is often concentrated early. We define the commitment horizon $\astar$ as the earliest point from which switching all remaining decoding to the base model reduces final success by no more than a chosen tolerance. Under the base model, the corresponding success probability, or committor, is a martingale. To first order, CFG's per-step effect is governed by the covariance between the guidance logit direction and the successor committor. This gives a local account of when guidance can help, but it does not by itself locate the horizon. Among prompts with an observed preterminal horizon, $\astar$ is usually early and varies more within constraint families than between them. Freezing each prompt at its own cross-fitted horizon is noninferior to full CFG on all 13 subtasks at the prespecified margin, even while many tokens remain masked. This separates commitment from realization. The boundary also identifies a later region in which higher parallelism adds only a small cost in constraint success, although fluency still degrades with parallel width. For failed trajectories, reopening committed positions improves recovery in both failure modes.

\end{abstract}


\section{Introduction}

% Classifier-free guidance (CFG) \citep{ho2022cfg,dhariwal2021classifierguidance} is a general mechanism for steering conditional generation in diffusion models. Here, we study its use in masked diffusion language models (MDLMs) \citep{nie2025llada,sahoo2024mdlm}, where it provides a flexible means of controlling properties of the generated text \citep{schiff2024simple,he2025guidance}. In practice, CFG is typically treated as a global control parameter: a fixed guidance weight is applied at every denoising step, and its effect is evaluated through a terminal metric such as the constraint-satisfaction rate; even recent adaptive schemes that vary the weight across the trajectory \citep{li2025acfg,wang2024schedulers,zhou2026guidance} still tune it against that terminal signal. These approaches reveal whether guidance helps on average, but not whether it remains necessary for a particular prompt and partial decoding state, nor when that trajectory can be safely handed back to the base decoder.

Classifier-free guidance (CFG)
\citep{ho2022cfg,dhariwal2021classifierguidance} is widely used to steer
diffusion models toward a conditioning signal. In masked diffusion language models (MDLMs) \citep{nie2025llada,sahoo2024mdlm}, CFG is commonly applied at every denoising step with a fixed guidance weight
\citep{schiff2024simple,he2025guidance}. Recent methods vary the guidance weight over time \citep{li2025acfg,wang2024schedulers,zhou2026guidance}, but they still do not directly answer a basic question: for a particular prompt
and partial output, when is CFG still needed, and when can it be removed
without a meaningful loss in final constraint satisfaction?

This question is especially relevant for MDLMs because different
properties of an output need not be resolved at the same time. A model may establish a high-level semantic or structural property before revealing all tokens, much as continuous diffusion can form coarse structure before fine detail \citep{wang2023painters}. A partial output may therefore already support the target constraint even while much of the sequence remains masked. In such a state, CFG may no longer be necessary, and the base model may be able to complete the remaining text on its own.

% We test this possibility by comparing continued CFG with base-only continuation from the same partial state. We formalize this comparison using committor functions under the two continuation policies. Our objective is not to measure how much the guided and base continuation distributions differ, but whether that difference changes the probability of satisfying the target constraint. A generic divergence such as KL can be large when guidance changes lexical realization while leaving terminal success unchanged, or small while moving consequential probability mass across the success boundary. The committor, a classical object in transition-path theory \citep{evanden2010tpt}, provides a state-conditioned measure of this outcome probability. For a partially decoded state $\mathbf{x}$ at denoising step $t$, let $q_t^0(\mathbf{x})$ denote the probability of terminal constraint satisfaction under base-only continuation, and let $q_t^g(\mathbf{x})$ denote the corresponding probability under guided continuation. Their difference, $V_t(\mathbf{x})=q_t^g(\mathbf{x})-q_t^0(\mathbf{x})$, measures the remaining task-relevant value of future guidance from the current state.

We test whether CFG is still needed by comparing two ways of completing the generation from the same partial output. In one case, CFG remains active for all remaining decoding steps. In the other, CFG is turned off and the base model completes the rest of the generation. We refer to these two future decoding processes as the guided continuation and the base-only continuation. If they have similar probabilities of eventually satisfying the target constraint, then CFG has little remaining value from the current state. These probabilities are committors in the sense of transition-path theory \citep{evanden2010tpt}, and their difference measures the remaining value of guidance.

% We use the remaining guidance value to identify a trajectory-level commitment horizon $a^*$. We define $a^*$ as the earliest point after which switching all remaining decoding from continued CFG to the base model reduces terminal constraint satisfaction by no more than a prespecified tolerance. It is therefore an interventional handoff boundary, determined by comparing two future continuation policies from the same partial state rather than by a local confidence score, mask ratio, or one-step transport statistic. Importantly, a small remaining guidance value indicates that CFG has become redundant relative to the base model; it does not by itself imply that the target constraint is likely to be satisfied.

This comparison defines a prompt-specific commitment horizon $\astar$. We define $\astar$ as the earliest point from which switching all remaining decoding to the base model reduces final success by no more than a chosen tolerance at every later switch point. Because it is defined by changing the future decoding policy, $\astar$ is
identified by the effect of removing CFG on final constraint satisfaction, rather than by confidence, mask ratio, noise level, or a generic distributional measure such as KL divergence.

% Guidance necessity varies along two distinct dimensions. At the prompt level, many prompts already satisfy the target constraint under base continuation, while for others CFG provides no measurable benefit or can even be harmful. For prompts that initially benefit from guidance, necessity is also temporal: among trajectories with a measurable commitment horizon, $a^*$ typically occurs early in decoding, while a substantial fraction of tokens remains masked. Its
% timing also varies across constraint families. These results reveal commitment before realization: for trajectories with a measurable successful handoff, base continuation preserves terminal constraint
% satisfaction within the prescribed tolerance while substantial token-level realization remains incomplete.

Our results show that guidance necessity varies strongly across prompts. Some prompts already succeed without CFG, some benefit from it, some show no measurable benefit, and some are harmed by it. For prompts that do benefit, the gain is often concentrated early in decoding. Among prompts with an observed preterminal horizon, switching at each prompt's cross-fitted $\astar$ is noninferior to full CFG on all 13 subtasks at the prespecified margin, even while many tokens remain masked. This shows that constraint commitment can occur before token-level realization is complete.

% To characterize how CFG changes continuation value before this boundary, we analyze the guided transition as an exponential tilt of the base kernel. For small guidance weights, the resulting one-step change in base-continuation value is governed by the covariance between the CFG logit contrast and successor committor values. This transport view shows that a guided step can increase continuation value only when it preferentially reweights reachable successors toward higher future success probability; when successor committors are locally indistinguishable, no normalized reweighting can change the expected continuation value.

We also provide a local explanation of when a guided step can help. Under the base decoder, the probability of eventual constraint satisfaction is a martingale. To first order, the effect of one CFG step depends on the covariance between the CFG logit direction and the successor committor. CFG can increase continuation value when it shifts probability toward next states with higher future success probability. This local characterization explains when guidance can help at a particular step, but it does not by itself locate the commitment horizon.

% The commitment horizon also organizes different decoding interventions. Before $a^*$, when the target outcome is not yet stably supported under base continuation, localized remasking or reopening can redirect trajectories whose current realization is drifting away from the constraint. After $a^*$, continued CFG becomes largely redundant, so the unconditional branch can be removed and generation handed off to the base decoder. We further use matched tail interventions to test whether decoding can become more parallel in this post-commitment regime. The additional success cost of increased parallelism becomes small in the neighborhood of the estimated horizon. 
% \input{fig_failure_modes}

\begin{figure}[t]
\centering
\includegraphics[width=1\linewidth]{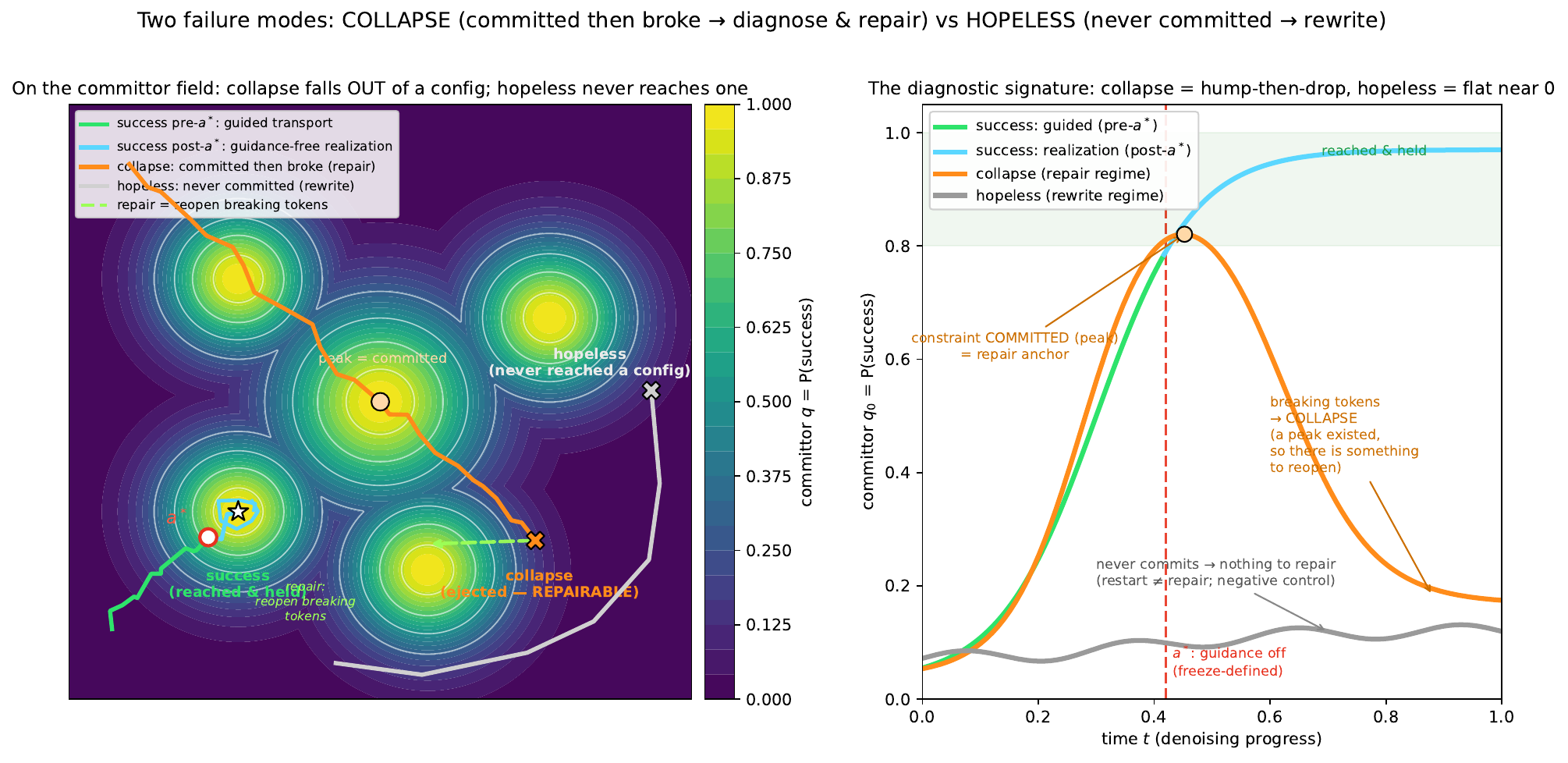}
\caption{The figure illustrates three trajectory types and is not a measurement. Reopening results are reported in
Section~\ref{sec:results:repair}.}
\label{fig:failure-modes}
\end{figure}

The horizon also helps organize later decoding decisions. Once CFG is no
longer needed, it can be removed. A later region often supports wider parallel decoding with only a small additional loss in constraint success, although this region need not coincide with $\astar$. For failed trajectories, reopening written positions improves recovery in both failure modes. Collapse refers to a trajectory that reaches a constraint-satisfying configuration and later leaves it, while hopeless failure refers to a trajectory that never reaches one. Figure~\ref{fig:failure-modes} illustrates the successful handoff and these two failure cases. % Generated by plot_failure_modes.py (figs/fig_failure_modes.pdf).
% Figure 1, Section 1.  SCHEMATIC -- nothing in it is measured, and the caption
% says so in its first sentence.  The committor field is a sum of Gaussians, the
% trajectories are hand-drawn polylines, the q0(t) curves are sigmoids.  Its job
% is to fix the vocabulary the rest of the paper uses.
%
% REMOVED 2026-07-28: the right panel used to annotate "committor > confidence"
% as the way to locate the breaking tokens.  That came from 10 sentiment prompts
% and does not replicate at n=396 (7 of 8 cells cross zero; the one significant
% cell favours confidence).  Do not reinstate it.
%
% The MEASURED counterpart of the right panel -- real q0 traces for the 196
% collapse and 599 hopeless prompts of the phase-1 classification -- is what the
% Sec 5.7 figure placeholder calls for and is a separate figure.

% Our contributions are fourfold. First, we formulate CFG necessity through continuation-specific committors, distinguishing task-relevant changes in terminal constraint satisfaction from generic differences between guided and base continuation distributions. Second, we show that guidance necessity varies at two levels: whether CFG is needed differs across prompts, and, for trajectories that initially benefit from it, how long it remains necessary is captured by an interventional commitment horizon $a^*$. Third, we derive a local transport characterization of CFG, showing that the one-step change in base-continuation value is governed, to first order, by the alignment between the CFG logit contrast and successor committor values. Finally, we show that the resulting boundary organizes stage-specific decoding interventions: localized reopening can redirect recoverable pre-commitment trajectories, whereas post-commitment decoding supports guidance handoff and increased parallelism with little additional loss in terminal success.

Our contributions are to define CFG necessity through paired continuations from the same partial output, introduce a prompt-specific commitment horizon for removing CFG within a prespecified tolerance, derive a local covariance characterization of CFG's first-order effect, and show how this framework supports CFG guidance handoff, later parallel decoding, and recovery through reopening.

\section{Related Work}

% \textbf{Masked diffusion language models.}
% Diffusion language models generate text by iterative denoising rather than left-to-right decoding, through either continuous latent representations \citep{li2022diffusionlm,gong2023diffuseq,han2023ssdlm,
% gulrajani2023likelihood,lovelace2023latent} or discrete token corruption
% \citep{austin2021d3pm,meng2022concrete,campbell2022continuous,lou2024sedd}. Masked or absorbing-state formulations \citep{sahoo2024mdlm,shi2024simplified,zheng2023reparameterized}, together with related discrete-flow models \citep{gat2024discrete,campbell2024generative}, have recently scaled to
% instruction-following dLLMs \citep{nie2025llada,gong2024scaling,arriola2025block}. We abstract their
% reverse process as a Markov chain over partially decoded states and study the terminal effect of switching between guided and base continuation policies, independently of architecture or noise schedule.

\textbf{Masked diffusion language models.}
Diffusion language models generate text through iterative denoising, using either continuous representations \citep{li2022diffusionlm,gong2023diffuseq} or discrete token corruption
\citep{austin2021d3pm,lou2024sedd}. Masked and absorbing-state formulations have recently scaled to instruction-following language models \citep{sahoo2024mdlm,shi2024simplified,nie2025llada}. We view their reverse process as a sequence of partially decoded states and study the effect of switching future decoding from CFG to the base model.

\textbf{Guidance in diffusion models.}
Classifier and classifier-free guidance modify reverse transitions toward a conditioning signal \citep{dhariwal2021classifierguidance,ho2022cfg}. Prior work improves guidance through geometric corrections, annealing, and adaptive or interval-restricted schedules \citep{bradley2024predictorcorrector,karras2024autoguidance,
chung2024cfgpp,hong2023improving,zheng2023characteristic,
sadat2024cads,sadat2024apg,he2025guidance,rojas2025improving,
kynkaanniemi2024interval,wang2024schedulers,li2025acfg,
zhou2026guidance}, including extensions to discrete diffusion \citep{schiff2024simple,nisonoff2024unlocking}. These methods mainly choose guidance strength or timing at the task or schedule level. We instead ask, for a particular prompt and partial state, whether future CFG still changes terminal constraint success.

\textbf{Committors and commitment.}
The committor is the probability of reaching a target set under specified dynamics and is central to transition-path theory and conditioned processes \citep{evanden2006towards,evanden2010tpt}. Related ideas have been used in generative modeling through learned $h$-transforms, constrained sampling, and bridge constructions
\citep{didi2023framework,denker2024deft,guo2026hardconstraint,
wang2025committor,vargas2023denoising,zhang2021path,
debortoli2021diffusion}. We use
committors only to measure terminal success from a partial state. The
difference between guided and base-continuation committors gives the remaining value of CFG. Unlike work on speciation or coarse-to-fine transitions in continuous diffusion
\citep{raya2023symmetry,biroli2024dynamical,achilli2026theory,
ambrogioni2023statistical,sclocchi2024phase,wang2023painters,
falck2025fourier,handke2026speciation}, our commitment horizon is defined directly by the effect of switching future decoding to the base model.

\textbf{Decoding and constrained generation.}
Recent dLLM decoders increase parallelism or adapt late-stage schedules
\citep{chen2025dparallel,shu2026dcd}. We study whether such changes remain compatible with terminal constraint satisfaction after CFG becomes unnecessary, without treating the commitment horizon as a general certificate of parallel safety. Our evaluation also draws on constrained-generation benchmarks covering lexical, structural, and instruction-following constraints \citep{lin2020commongen,yao2024collie,zhou2023ifeval}. We use them as controlled settings for measuring guidance necessity across prompts and decoding stages, rather than primarily for ranking generation systems.

\section{Setup: Decoding and Guidance Necessity}
\label{sec:setup}

\subsection{Masked Diffusion Decoding and Terminal Success}
\label{sec:setup:markov}
\label{sec:setup:success}

Fix a prompt $p$ and suppress this dependence below. Let
$X_t \in (V \cup \{\texttt{M}\})^L$ denote the partially decoded sequence at
step $t$, where $V$ is the vocabulary, $\texttt{M}$ is the mask symbol, and
$L$ is the canvas length. Decoding starts from the fully masked state $X_0$
and ends with a complete output $X_T=Y\in V^L$.

\begin{assumption}[Markov generation]
\label{as:markov}
Conditioned on the prompt, the reverse process is a first-order Markov chain
over partially decoded states:
\begin{equation}
\begin{aligned}
\Pr(X_{t+1}=x' \mid X_{0:t}=x_{0:t})
&=
\Pr(X_{t+1}=x' \mid X_t=x_t) \\
&=
\Kbase_t(x' \mid x_t).
\end{aligned}
\label{eq:markov-kernel}
\end{equation}
Its natural filtration is
$\mathcal F_t=\sigma(X_0,\ldots,X_t)$.
\end{assumption}

This representation is standard for masked-diffusion decoding. If a decoder
uses additional history, that history can be included in the state.

A deterministic criterion $S:V^L\to\{0,1\}$ indicates whether the terminal
output satisfies the target constraint, with success set
$\mathcal A=\{y\in V^L:S(y)=1\}$. For a decoding kernel $K$, the
prompt-conditioned success rate is
\begin{equation}
\mathrm{SR}(K)=\Pr_K\!\left(S(Y)=1\right).
\label{eq:success-rate}
\end{equation}
Dataset-level success rates average this quantity over prompts.

\subsection{Base and Guided Committors}
\label{sec:setup:committor}

A committor is the probability of terminal success under a specified future
decoding policy. The \textbf{base committor} is
\begin{equation}
\qbase_t(x)
=
\Pr_{\Kbase}\!\left(S(Y)=1 \mid X_t=x\right),
\label{eq:setup-committor}
\end{equation}
the probability of satisfying the target constraint when all remaining steps
use the pure-conditional base decoder, with the prompt retained.

To define the guided policy used below, let
$\ell_t^c(\cdot\mid x)$ and $\ell_t^u(\cdot\mid x)$ denote the conditional and
unconditional logits. CFG with weight $w$ uses
\[
\ell_t^c(\cdot\mid x)+w\,\delta_t(\cdot\mid x),
\qquad
\delta_t(\cdot\mid x)
=
\ell_t^c(\cdot\mid x)-\ell_t^u(\cdot\mid x),
\]
where $\delta_t$ is the guidance direction. Let $\Kguid_t$ denote the
resulting transition kernel. The \textbf{guided committor} is
\begin{equation}
\qguid_t(x)
=
\Pr_{\Kguid}\!\left(S(Y)=1 \mid X_t=x\right),
\label{eq:setup-guided-committor}
\end{equation}
the probability of terminal success when CFG remains active for all remaining
steps.

\subsection{Remaining Guidance Value and the Commitment Horizon}
\label{sec:setup:astar}

The \textbf{remaining guidance value}
\begin{equation}
\Vval_t(x)
=
\qguid_t(x)-\qbase_t(x)
\label{eq:setup-value}
\end{equation}
measures the gain in terminal success from keeping CFG active rather than
switching immediately to the base decoder. For a switch point $t$, let $\Ssw(t)$ denote the terminal success probability
of using CFG before $t$ and the base decoder from $t$ onward. Thus,
$\Ssw(0)=\mathrm{SR}(\Kbase)$ and
$\Ssw(T)=\mathrm{SR}(\Kguid)$.

\begin{definition}[Commitment horizon]
\label{def:astar}
For tolerance $\varepsilon>0$, define
\begin{equation}
\astar_{\varepsilon}
=
\inf\left\{
t:
\Ssw(s)\ge \Ssw(T)-\varepsilon
\ \text{for all }s\ge t
\right\}.
\label{eq:setup-astar}
\end{equation}
\end{definition}

The commitment horizon is the earliest persistent switch point after which
removing CFG costs at most $\varepsilon$ in terminal success. If no
preterminal switch point satisfies this condition, then
$\astar_{\varepsilon}=T$ and we call the prompt \emph{right-censored}.
Well-definedness and monotonicity in
$\varepsilon$ are discussed in Appendix~\ref{app:A-variance}.

Let $X_t^g$ denote the state reached at time $t$ under fully guided decoding.
By the Markov property,
\[
\Ssw(t)
=
\Ebb^g\!\left[\qbase_t(X_t^g)\right],
\qquad
\Ssw(T)
=
\Ebb^g\!\left[\qguid_t(X_t^g)\right].
\]
Therefore,
\begin{equation}
\Ssw(T)-\Ssw(t)
=
\Ebb^g\!\left[\Vval_t(X_t^g)\right].
\label{eq:setup-vlink}
\end{equation}
Thus, the cost of switching off CFG at time $t$ is the average remaining
guidance value over states produced by the guided prefix. The horizon is
therefore prompt-specific and is defined by changing the future decoding
policy.

\paragraph{Estimation.}
We estimate committors and the switch curve $\Ssw$ using repeated continuation
rollouts. The estimate $\astar$ is the earliest evaluated switch point
that satisfies the tolerance condition at every later grid point. Horizon
selection and post-handoff evaluation use disjoint rollout halves through
cross-fitting. Rollout counts, switch grids, and tolerance settings are given
in Section~\ref{sec:results} and Appendix~\ref{app:census}.

\section{When Is Guidance Useful?}
\label{sec:theory}
\label{sec:transport}

\subsection{How Guidance Creates Local Lift: Martingales and Covariance}
\label{sec:theory:transport}
\label{sec:theory:root}
\label{sec:theory:cfg}
\label{sec:theory:cov}

Under Assumption~\ref{as:markov}, the base committor satisfies a backward
harmonic equation.

\begin{theorem}[Martingale root]
\label{thm:martingale}
Under Assumption~\ref{as:markov},
\begin{equation}
\qbase_t(x)
=
\sum_{x'}
\Kbase_t(x'\mid x)\,
\qbase_{t+1}(x'),
\label{eq:th-harmonic}
\end{equation}
with terminal condition $\qbase_T(Y)=S(Y)$. Equivalently,
\begin{equation}
\qbase_t(X_t)
=
\Ebb^{0}\!\left[
\qbase_{t+1}(X_{t+1})
\mid \mathcal F_t
\right],
\label{eq:th-martingale}
\end{equation}
so $\{\qbase_t(X_t)\}_{t=0}^{T}$ is a martingale under the base kernel.
\end{theorem}

This is the backward equation for terminal success
(Appendix~\ref{app:A-martingale}). Along a base trajectory,
$\qbase_t(X_t)$ may fluctuate and eventually resolve to $0$ or $1$, but its
conditional expected increment is zero. The base process therefore provides a
zero-drift reference for measuring the effect of guidance.

To measure that effect, let $\Kguid_{t,w}$ denote the guided kernel at weight
$w$, obtained from the CFG logit tilt defined in
Section~\ref{sec:setup:committor}. The one-step \textbf{local lift} is
\begin{equation}
\Delta q_t^{g,w}(x)
=
\Ebb_{\Kguid_{t,w}}\!\left[
\qbase_{t+1}(X_{t+1})
\mid X_t=x
\right]
-
\qbase_t(x).
\label{eq:th-local-lift}
\end{equation}
It measures how one guided step changes expected success under future base
continuation. It is the local counterpart of the remaining guidance value
$\Vval_t(x)=\qguid_t(x)-\qbase_t(x)$ defined in
Section~\ref{sec:setup:astar}.

Because the guided next-step law is an exponential tilt of the base law, its
derivative at $w=0$ satisfies the following linear-response identity
(Appendix~\ref{app:A-cov}):
\begin{equation}
\left.
\frac{\partial}{\partial w}
\Ebb_{\Kguid_{t,w}}\!\left[
\qbase_{t+1}(X_{t+1})
\mid X_t=x
\right]
\right|_{w=0}
=
\mathrm{Cov}_{\Kbase_t(\cdot\mid x)}
\!\left(
\delta_t,\qbase_{t+1}
\right).
\label{eq:th-cov}
\end{equation}

Equation~\ref{eq:th-cov} gives the local mechanism by which CFG can help.
Guidance increases expected continuation success when its logit direction
assigns more probability to successor states with larger base committors. If
the guidance direction is unrelated or opposed to successor success, the
first-order effect is zero or negative.

For finite but small $w$,
\begin{equation}
\Delta q_t^{g,w}(x)
=
w\,
\mathrm{Cov}_{\Kbase_t(\cdot\mid x)}
\!\left(
\delta_t,\qbase_{t+1}
\right)
+
O(w^2).
\label{eq:th-cov-first-order}
\end{equation}
Thus, Eq.~\ref{eq:th-cov} is exact as a derivative at $w=0$, whereas
Eq.~\ref{eq:th-cov-first-order} is a first-order approximation. Only the
component of $\delta_t$ aligned with the successor committor contributes at
first order (Appendix~\ref{app:A-align}).

For comparison, ideal conditioning through the Doob transform has one-step
gain
\begin{equation}
\frac{
\Var^{0}\!\left(
\qbase_{t+1}(X_{t+1})
\mid X_t=x
\right)
}{
\qbase_t(x)
},
\label{eq:th-doob-gain}
\end{equation}
when $\qbase_t(x)>0$
(Corollary~\ref{cor:utility}; Appendix~\ref{app:A-doob}).
Both results identify variation in successor committor values as the local
opportunity for steering. The ideal transform exploits this variation
directly, while CFG helps only when its logit tilt aligns with it.

In short, CFG can help by shifting next-step probability toward states with
higher future success probability. This is a local, first-order statement; it
does not determine how long guidance remains useful or where the commitment
horizon occurs.

\subsection{When Does Guidance Transport? Failure Modes of the First-Order Law}
\label{sec:theory:useful}

The covariance in Eq.~\ref{eq:th-cov} separates into alignment and two
sources of local variation. Let all moments below be taken under
$\Kbase_t(\cdot\mid x)$, and define
\begin{equation}
\rho_t(x)
=
\frac{
\mathrm{Cov}\!\left(\delta_t,\qbase_{t+1}\right)
}{
\sigma(\delta_t)\,
\sigma(\qbase_{t+1})
},
\label{eq:th-rho}
\end{equation}
whenever both standard deviations are nonzero. For positive and sufficiently
small $w$,
\begin{equation}
\Delta q_t^{g,w}(x)
=
w\,\rho_t(x)\,
\sigma(\delta_t)\,
\sigma(\qbase_{t+1})
+
O(w^2).
\label{eq:th-alignment}
\end{equation}

The first-order effect can fail for three reasons. First, the
\textbf{guidance tilt can collapse}:
$\sigma(\delta_t)=0$. In this case, $\delta_t$ is constant over reachable
successors, so normalization removes the tilt and
$\Kguid_{t,w}=\Kbase_t$ for every $w$. The local lift is therefore exactly
zero, and $\rho_t$ is undefined rather than zero.

Second, the \textbf{success opportunity can collapse}:
$\sigma(\qbase_{t+1})=0$. All reachable successors then have the same
base-continuation success probability, so redistributing probability among
them cannot change the expected committor. If this condition persists over
the remaining continuation, future guidance value vanishes and the
commitment horizon is reached.

Third, the guidance direction can be
\textbf{misaligned} with continuation success. When $\rho_t=0$, the
first-order effect vanishes; when $\rho_t<0$, guidance moves probability
toward successors with lower base-continuation success and the first-order
effect is negative.

Thus, for $w>0$, the first-order term is positive exactly when
\[
\sigma(\delta_t)>0,
\qquad
\sigma(\qbase_{t+1})>0,
\qquad
\rho_t>0.
\]
This is a local statement about one decoding step. It does not require
$\delta_t$ to approximate the committor globally, and it does not determine
the commitment horizon by itself. How often positive local transport occurs
is empirical; Appendix~\ref{app:transport-modes} finds that it is uncommon
and concentrated early in decoding.

The factorization is independent of the particular constraint, but its
factors are not. In particular, when successor committor variation disappears
depends on the prompt and on the structure of the success set $\mathcal A$.
Section~\ref{sec:results} measures this variation across constraint families.

\subsection{For How Long? The Commitment Horizon
\texorpdfstring{$\astar$}{a*}}
\label{sec:theory:astar}
\label{sec:theory:existence}

For readability, we suppress the fixed guidance weight $w$ and write
$\Delta q_t^g$ for the local lift defined in
Eq.~\ref{eq:th-local-lift}.

\paragraph{The horizon depends on future transport.}
A single local lift does not determine the commitment horizon. By
Definition~\ref{def:astar}, $\astar_\varepsilon$ is defined by the effect of
switching all remaining decoding to the base kernel, not by a threshold on a
one-step statistic.

The local and global quantities are nevertheless linked exactly.

\begin{proposition}[Accumulated local transport]
\label{prop:transport-sum}
For any state $x$ reached at step $t$,
\begin{equation}
\Vval_t(x)
=
\Ebb^g\!\left[
\sum_{s=t}^{T-1}
\Delta q_s^g(X_s)
\;\middle|\;
X_t=x
\right].
\label{eq:th-transport-sum}
\end{equation}
\end{proposition}

Combining this identity with Eq.~\ref{eq:setup-vlink} gives
\begin{equation}
\Ssw(T)-\Ssw(t)
=
\Ebb^g\!\left[
\sum_{s=t}^{T-1}
\Delta q_s^g(X_s^g)
\right].
\label{eq:th-freeze-transport}
\end{equation}
Thus, the cost of switching CFG off at time $t$ is exactly the expected future
accumulation of local guided transport. A small lift at one step is not enough
to certify the horizon, because later lifts may still contribute.

\paragraph{Why future transport can vanish.}
By Cauchy--Schwarz and Eq.~\ref{eq:th-cov},
\begin{equation}
\left|\Delta q_t^g(x)\right|
\leq
w\,\sigma(\delta_t)\,
\sigma(\qbase_{t+1})
+
O(w^2),
\label{eq:th-cs-bound}
\end{equation}
with moments under $\Kbase_t(\cdot\mid x)$. Hence, guidance has less room to
help when reachable successors have similar base-continuation values. Under
the base kernel this variation obeys an exact budget that is exhausted as the
committor resolves (Corollary~\ref{cor:budget};
Appendix~\ref{app:A-variance}), and the Cauchy--Schwarz cap above is the only
bridge we assert between that base-measure budget and guided transport. If
this contraction persists, the future sum in
Eq.~\ref{eq:th-freeze-transport} eventually falls within
$\varepsilon$.

\paragraph{What the theory guarantees.}
The weak result $\astar_\varepsilon\leq T$ always holds, since switching at the
terminal point has zero cost. The theory does not imply that
$\astar_\varepsilon$ is early or sharp; those are empirical properties of the
decoder and constraint, since the martingale may deposit its variance on any
schedule consistent with the budget
(Appendix~\ref{app:A-weakstrong}; a worked absorbing-committor example in
which both regimes are visible is in Appendix~\ref{app:toy}). It also does
not guarantee that wider parallel decoding is safe after $\astar$, because
changing parallelism changes the base
kernel (Appendix~\ref{app:parallel-stability} bounds the resulting error). We
test both questions separately in Section~\ref{sec:results}.

\section{Results}
\label{sec:results}

\subsection{Guidance Dependence Is Prompt-Specific}
\label{sec:results:taxonomy}

We first ask whether guidance dependence is shared by all prompts in a task.
We evaluate LLaDA-Instruct~\citep{nie2025llada} on 13 subtasks drawn from
CommonGen~\citep{lin2020commongen}, IFEval~\citep{zhou2023ifeval},
COLLIE~\citep{yao2024collie}, and our newly constructed
Constraint-Controlled Generation (CCG) suite. Each subtask contains $n{=}200$
prompts, and every prompt is decoded at every grid point, with no sampling and
no filtering; we call this exhaustive sweep a \emph{census}. We use $w{=}2$,
64 generated tokens, and 64 denoising steps as the center configuration, and
vary one factor at a time (OFAT) around this setting. Success criteria and the
full experimental design are given in Appendix~\ref{app:constraints} and
Appendix~\ref{app:census}.

\begin{figure}[t]
\centering
\includegraphics[width=\linewidth]{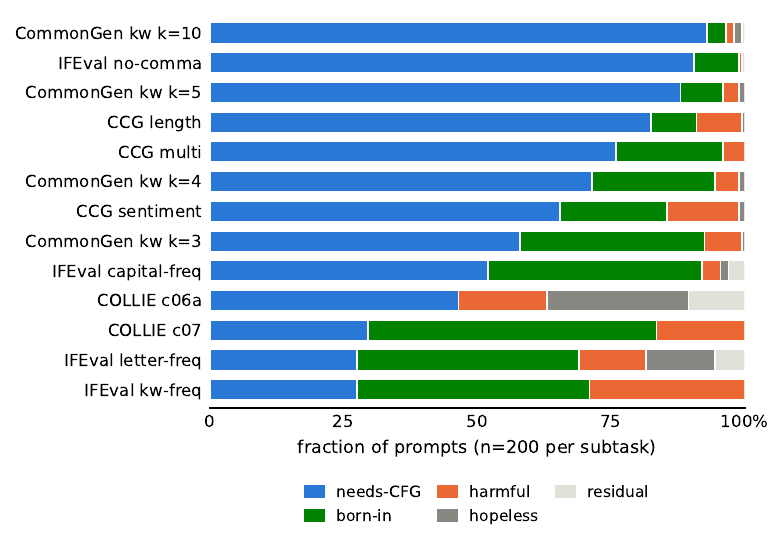}
\caption{Prompt-level guidance fates across the 13-subtask census
($n{=}200$ prompts each, center configuration). Every subtask mixes prompts
that need guidance, succeed without it, are harmed by it, or never succeed;
COLLIE c07 and c06a anchor the born-in and failure-dominated ends.}
\label{fig:typology}
\end{figure}

Guidance dependence is not uniform within any subtask. Every subtask contains
a mixture of prompts that succeed without guidance (born-in), benefit from guidance (needs-CFG),
are harmed by guidance (harmful), or fail under both policies (hopeless)
(Figure~\ref{fig:typology}). 

These results show that task-level success rates hide qualitatively different
prompt-level behaviors. We therefore report the full fate mixture and define a
nontrivial commitment horizon only for prompts whose success depends on
guidance over part of the trajectory. Fluency is evaluated separately:
moderate guidance can improve perplexity relative to the unguided model,
whereas stronger guidance can degrade it. Full census, OFAT, and fluency
results are reported in Appendix~\ref{app:census}.

% Auto-generated by analysis_utilities/injection_sharpening/census_noninferiority.py
% pattern {task}_cfg2_gen64_c*.json -- do not edit by hand.
\begin{table*}[t]
\centering
\small
\setlength{\tabcolsep}{4pt}
\begin{tabular}{l r c c c c c c r}
\toprule
Constraint & $n$ & Full CFG & Freeze at $a^{*}$ & Abs.\ gap & 95\% CI & Retention & Med.\ $ a^{*}/T$ & $ a^{*}{=}T$ (\%) \\
\midrule
CommonGen kw $k{=}3$ & 200 & 0.828 & 0.819 & -0.009 & $[-0.021,+0.007]$ & 0.99 & 0.22 & 27 \\
CommonGen kw $k{=}4$ & 200 & 0.761 & 0.769 & +0.008 & $[-0.007,+0.027]$ & 1.01 & 0.30 & 33 \\
CommonGen kw $k{=}5$ & 200 & 0.644 & 0.636 & -0.008 & $[-0.016,+0.003]$ & 0.99 & 0.26 & 49 \\
CommonGen kw $k{=}10$ & 200 & 0.762 & 0.750 & -0.012 & $[-0.020,+0.001]$ & 0.98 & 0.39 & 42 \\
\addlinespace
CCG length & 200 & 0.821 & 0.837 & +0.016 & $[-0.002,+0.036]$ & 1.02 & 0.16 & 27 \\
CCG sentiment & 200 & 0.666 & 0.708 & +0.042 & $[+0.022,+0.065]$ & 1.06 & 0.30 & 46 \\
CCG multi & 200 & 0.890 & 0.867 & -0.024 & $[-0.030,-0.017]$ & 0.97 & 0.30 & 18 \\
\addlinespace
IFEval kw-freq & 200 & 0.935 & 0.964 & +0.029 & $[+0.014,+0.046]$ & 1.03 & 0.22 & 4 \\
IFEval letter-freq & 200 & 0.696 & 0.684 & -0.012 & $[-0.017,-0.008]$ & 0.98 & 0.22 & 33 \\
IFEval capital-freq & 200 & 0.737 & 0.739 & +0.002 & $[-0.004,+0.011]$ & 1.00 & 0.42 & 35 \\
IFEval no-comma & 200 & 0.865 & 0.847 & -0.017 & $[-0.024,-0.010]$ & 0.98 & 0.24 & 24 \\
\addlinespace
COLLIE c07 & 200 & 0.957 & 0.962 & +0.005 & $[-0.000,+0.012]$ & 1.01 & 0.30 & 10 \\
COLLIE c06a & 200 & 0.064 & 0.064 & +0.000 & $[+0.000,+0.000]$ & 1.00 & -- & 100 \\
\bottomrule
\end{tabular}
\caption{Freeze-at-$ a^{*}$ versus full CFG on the full $n{=}200$ census
per family, evaluated with cross-fitting and right-censoring at
$ a^{*}{=}T$. All rows meet the prespecified noninferiority margin
$\varepsilon{=}0.03$; see Appendix~C.5.}
\label{tab:noninferiority}
\end{table*}

\begin{table}[t]
\centering

\small
\setlength{\tabcolsep}{3.8pt}
\renewcommand{\arraystretch}{1.08}

\begin{tabular}{
  l
  c
  S[table-format=1.2]
  S[table-format=-1.2]
  S[table-format=3.0]
  S[table-format=3.0]
}
\toprule
{Policy}
& {$K$}
& {Fresh SR $\uparrow$}
& {$\Delta_K$}
& {PPL $\downarrow$}
& {Fwd. $\downarrow$} \\
\midrule

\multirow{4}{*}{\textit{Full CFG}}
& 1  & 0.83 & {--}    & 123 & 128 \\
& 4  & 0.81 & -0.12 & 301 &  32 \\
& 8  & 0.77 & -0.16 & 272 &  16 \\
& 16 & 0.68 & -0.25 & 238 &   8 \\
\midrule
\addlinespace[3pt]

\multirow{4}{*}{\shortstack[l]{\textit{Handoff at}\\
\textit{recorded $\astar$}}}
& 1  & 0.82 & {--}   &  85 & 84 \\
& 4  & 0.82 &  0.00 & 107 & 51 \\
& 8  & 0.80 & -0.02 & 153 & 46 \\
& 16 & 0.76 & -0.07 & 264 & 43 \\

\bottomrule
\end{tabular}

\caption{
Fresh-trajectory deployment on 905 handoff survivors from the
$w{=}2$ census. Higher Fresh SR (success rate) is better; lower PPL (perplexity) and Fwd.(forward evaluation) are better.
}
\label{tab:policy-quant}
\end{table}
% Auto-generated by analysis_utilities/injection_sharpening/
% make_remask_recovery_table.py
% from results/committor_remask/replay_*.json -- do not edit by hand.

\begin{table*}[t]
\centering
\small
\setlength{\tabcolsep}{5pt}
\renewcommand{\arraystretch}{1.05}

\begin{tabular}{l c c c c c c c}
\toprule
&
&
\multicolumn{3}{c}{Pre-anchor}
&
\multicolumn{3}{c}{Post-peak anchor} \\
\cmidrule(lr){3-5}
\cmidrule(lr){6-8}

Selection signal
& {Cost $\downarrow$}
& {SR $\uparrow$}
& {Gain $\uparrow$}
& {PPL $\downarrow$}
& {SR $\uparrow$}
& {Gain $\uparrow$}
& {PPL $\downarrow$} \\
\midrule

No reopen (control)
& --
& 0.23
& --
& 195
& 0.20
& --
& 185 \\

\addlinespace

Full committor $\Delta q^{0}$
& $\sim100{\times}$
& 0.39
& +0.16
& 138
& 0.34
& +0.14
& 168 \\

Truncated committor ($k{=}8$)
& $\sim13{\times}$
& 0.41
& +0.18
& 132
& 0.31
& +0.11
& 165 \\

Scorer leave-one-out (no rollout)
& $\approx 0$
& 0.40
& +0.17
& 147
& 0.30
& +0.10
& 250 \\

Guidance disagreement (no rollout)
& $\approx 0$
& 0.37
& +0.14
& 174
& 0.26
& +0.07
& 193 \\

Commit confidence
& free
& 0.38
& +0.15
& 163
& 0.35
& +0.16
& 187 \\

\bottomrule
\end{tabular}

% CAPTION SHORTENED 2026-07-28 when this table moved into the body
% (Sec. 5.3): the K=1 overlap sentence (14--36%), the anchor-gradient
% recovery CIs, and the census composition now live in app:remask,
% which states them verbatim.

\caption{
Reopen recovery over the full intervention census
(396 prompts, 13 subtasks; $K{=}3$ reopened positions; guided continuation;
two seeds).}
\label{tab:remask-recovery}
\end{table*}
% Auto-generated by analysis_utilities/injection_sharpening/
% make_cheap_gate_table.py -- do not edit by hand.

\begin{table*}[t]
\centering
\footnotesize
\setlength{\tabcolsep}{3pt}
\renewcommand{\arraystretch}{1.05}

\begin{tabular}{l c c c l r r}
\toprule
&
\multicolumn{3}{c}{Success $\uparrow$}
&
{\shortstack{$\Delta$ vs.\ oracle $\uparrow$}}
&
{\shortstack{PPL $\downarrow$}}
&
{\shortstack{Fwd. $\downarrow$}} \\
\cmidrule(lr){2-4}

Gate source
& {$K{=}4$}
& {$K{=}8$}
& {$K{=}16$}
& {at $K{=}8$ [95\% CI]}
& {$K{=}8$}
& {$K{=}8$} \\
\midrule

Oracle $\astar$ (rollouts)
& .773
& .757
& .710
& ---
& 157
& 43.1 \\

\addlinespace[2pt]

\multicolumn{7}{l}{\emph{Rollout-free gates}} \\

\quad raw $\hat a^{*}$
& .782
& .772
& .733
& $+.015$\,$[-.001,+.031]$
& 150
& 46.5 \\

\quad policy
& .857
& .849
& .829
& $+.092$\,$[+.075,+.108]$
& 142
& 64.6 \\

Per-subtask constant
& .880
& .873
& .855
& $+.116$\,$[+.100,+.131]$
& 141
& 75.3 \\

\bottomrule
\end{tabular}

\caption{
Comparison of rollout-based and rollout-free gate sources on the same
fresh gate-by-$K$ arms over 905 prompts with recorded horizons.
}
\label{tab:cheap-gate}
\end{table*}

\subsection{Commitment Is Prompt-Specific and Guidance Value Is Front-Loaded}
\label{sec:results:horizon}
\label{sec:results:astar}

Commitment happens at different times for different prompts. We estimate each
prompt's horizon $\astar$ from paired freeze interventions at seven normalized
generation fractions between $0.05$ and $0.60$. At each fraction, the prefix is
generated with CFG and the continuation with the base kernel. We define
$\astar$ as the earliest tested point from which switching remains safe at all
later grid points. Prompts that already succeed without guidance are excluded,
because they do not have a meaningful handoff point. Prompts that still require
guidance at the final tested fraction are kept as unresolved cases rather than
assigned $\astar=0.60$. Estimator details, sample counts, and the treatment of
these unresolved cases are given in Appendix~\ref{app:astar-ofat}.

\begin{figure}[t]
\centering
\includegraphics[width=\linewidth]{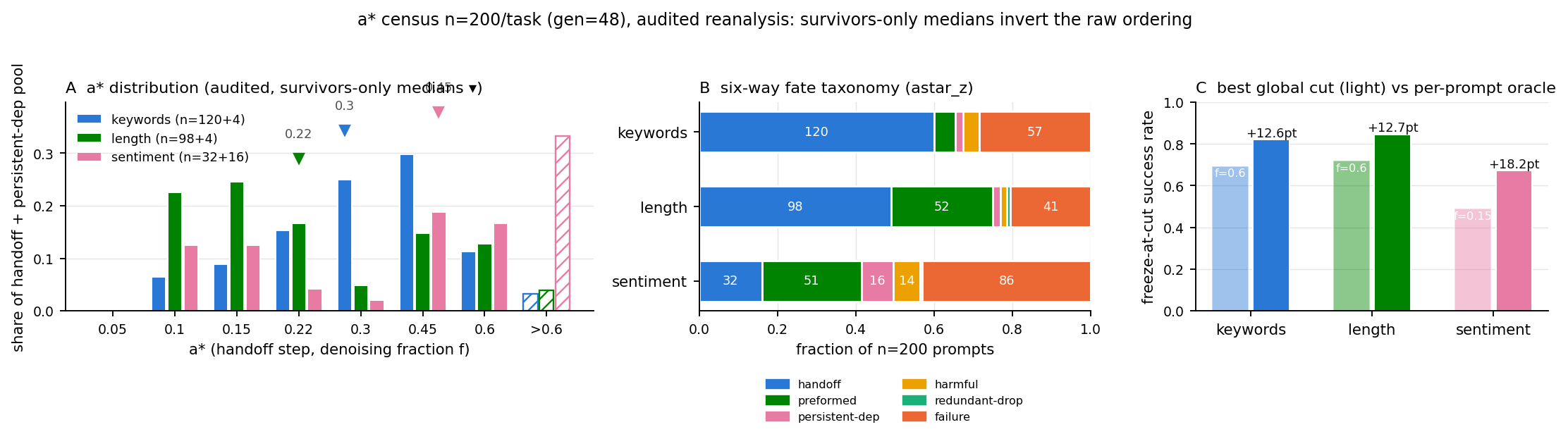}
\caption{Prompt-level $\astar$ distributions and the cross-fitted advantage of
prompt-specific horizons over a single global cut. The hatched ${>}0.6$ bucket
contains prompts for which no safe switch was observed before the final tested
fraction.}
\label{fig:astar-dist}
\end{figure}

The estimated horizons vary widely across prompts
(Figure~\ref{fig:astar-dist}). Changing the constraint shifts the horizon
distribution, so $\astar$ is not determined by the noise schedule alone.
However, most variation occurs between prompts within the same constraint
family: constraint family accounts for only about $10\%$ of the total
variation in $\astar$ ($\eta^{2}{=}0.097$). This variation matters in
practice. Using each prompt's cross-fitted $\astar$ outperforms the best single
global cut by $9.5\,[6.1,13.2]$, $11.9\,[7.2,17.1]$, and
$13.4\,[9.9,16.9]$ success points on keywords, length, and sentiment,
respectively. We therefore treat $\astar$ as a prompt-level handoff boundary,
not as a task-level constant or a predictor of eventual success.

Most of guidance's benefit is gained before $\astar$. We switch guidance off
at the estimated horizon and let the base model finish. On the full
13-subtask census, the handoff policy is noninferior to full CFG on all
13 subtasks at the prespecified margin $\varepsilon{=}0.03$
(Table~\ref{tab:noninferiority}). By
Eq.~\ref{eq:th-freeze-transport}, the switch-off cost is the expected sum of
future local transport. A switch curve that remains flat after $\astar$
therefore indicates that this remaining guidance value has fallen below the
tolerance.

At the switch, most of the sequence is still masked---$70\%$, $78\%$, and
$55\%$ of positions at the median horizon on keywords, length, and sentiment,
respectively. Thus, constraint success is effectively committed before most
tokens are revealed. The freeze curves are reported in
Appendix~\ref{app:astar-ofat}. This plateau concerns terminal constraint
success only. Measured separately, freezing does not reduce fluency and
improves it in the audited wave (Appendix~\ref{app:census}). The result shows
that guidance adds little after $\astar$; it does not mean that guidance was
unnecessary before $\astar$. On CCG sentiment, handoff is strictly better than
full CFG, indicating that late guidance can even be harmful.
Appendix~\ref{app:transport-modes} provides an auxiliary analysis of why
transport weakens after the boundary, but the main result does not depend on
that explanation.

\subsection{Interventions on a Committed State}
\label{sec:results:realization}
\label{sec:results:repair}

After locating the commitment horizon, we study two post-commitment actions.
For trajectories with a successful preterminal handoff, we test whether
parallel decoding can finish the remaining tokens at low cost. For collapse
and hopeless failures, we test whether remasking already filled positions can
repair earlier mistakes, since absorbing unmasking cannot revise committed
positions (Appendix~\ref{app:macrostate}).

\textbf{Parallel realization.}
Handoff at $\astar$ does not guarantee parallel safety
(Proposition~\ref{prop:twohop}), so we test the continuation kernel separately
with fixed-prefix comparisons and matched gate sweeps. Widening the decoding
block before the horizon is costly: moving the gate from one grid step before
$\astar$ to $\astar$ gives the largest gain in the sweep
($+0.178\pm0.012$ in constraint success), while each later step adds at most
$0.02$. The parallel-specific $K{=}8$ versus $K{=}1$ gap likewise falls from
$7.8\pm2.5$ points near $\astar$ to $0.3\pm1.2$ one grid step later; on length
constraints it remains below $2$ points throughout. Thus, $\astar$ marks the
start of a transition to a low-cost parallel regime, typically reached one to
two gate intervals later. Table~\ref{tab:policy-quant} confirms the pattern on fresh rollouts. Increasing
$K$ from $1$ to $16$ reduces success by $0.25$ under full CFG but only $0.07$
after handoff. However, wider decoding worsens GPT-2 perplexity relative to
$K{=}1$ under both policies. Parallelism after handoff is therefore less
costly in constraint success, but not necessarily in fluency. Full results are
in Appendix~\ref{app:gate-sweep}.

\textbf{Repair by reopening.} Both failure modes of
Figure~\ref{fig:failure-modes} respond to reopening. Across $396$ prompts from
$13$ subtasks, reopening written positions and resuming guided decoding
improves recovery in all eight (anchor $\times$ type $\times$ $K$) cells and in
every subtask---$+0.176$ on collapse and $+0.152$ on hopeless at the earlier
anchor---so the gain does not require that the trajectory ever reached a
constraint-satisfying configuration (Table~\ref{tab:remask-recovery}). Nor does
it require identifying a particular position: at $K{=}1$ the
counterfactual-committor and commit-confidence orderings select the same
position only $14$--$36\%$ of the time and recover the same amount. What
carries the effect is that a committed position is reopened at all. All outcome
arms were rerun after we corrected a length-matching artifact; complete design,
per-cell results and implementation audits appear in
Appendix~\ref{app:remask}.

\subsection{A Cheap $\hat \astar$ Is Sufficient}
\label{sec:results:predicted}

Measuring $\astar$ requires paired rollouts per prompt, but deployment needs
only to localize it well enough to choose a gate. We therefore train a
rollout-free random forest on confidence, entropy and commit-order features of
a prompt's own run, within subtask. At $K{=}8$ the induced gate matches the
rollout-measured oracle in success---$+0.015$ $[-0.001, +0.031]$ over $905$
prompts with a recorded horizon---while using $46.5$ against $43.1$ forward
evaluations per sequence (Table~\ref{tab:cheap-gate}).

This near-parity despite coarse horizon estimates reflects the success plateau
around $\astar$: modest gate errors usually cost little terminal success. Two
qualifications travel with it. The features do not transfer across constraint
families---trained leave-one-task-out the pipeline falls to chance---so a
deployed gate must stay family-calibrated. And read against the other
baseline the comparison runs the other way: a later per-subtask constant cut
scores above gating at $\astar$ ($+0.116$ $[+0.100, +0.131]$ pooled), but it
gates later and buys the difference with guided steps, so that contrast states
how much guidance a deployment is willing to spend rather than where the
boundary lies. CCG sentiment is the main exception to the plateau---the family
in which guidance past the horizon is harmful, so the switch curve turns down
instead of flattening and the per-prompt location is what a policy must get
right. Prediction accuracy, the policy rule, the constant-cut and
intention-to-treat comparisons and the matched-budget analysis are in
Appendix~\ref{app:cheap-gate}. Compute is reported throughout as model forward
evaluations, descriptively: we make no wall-clock claim and none of a net gain
over existing schedulers.

\section{Conclusion}

Our results support a prompt-specific view of CFG in masked discrete
diffusion. Prompts differ in whether they need guidance at all, and, for
prompts with an observed preterminal commitment horizon $\astar$, switching
decoding to the base model causes little loss in constraint success. In the
settings we study, this horizon is often reached while many positions remain
masked, showing that constraint commitment can precede token-level
realization.

The theory explains the mechanism but does not locate the horizon. Under the
base kernel, the committor is a martingale. CFG changes its predictable drift,
with a first-order rate determined by the covariance between the guidance
direction and the successor committor. These local lifts accumulate into the
remaining value of guidance. However, neither an early horizon nor a later
low-cost parallel regime follows from the theory. Both are empirical and
depend on the decoder and constraint, and $\astar$ concerns terminal
constraint success rather than all aspects of generation quality.

We therefore view $\astar$ as a prompt-specific diagnostic for when CFG can be
removed within a tolerance, not as a universal certificate for later decoding
changes. Wider parallel decoding must be tested separately. For failed
trajectories, reopening written positions improves recovery in both collapse
and hopeless failure, with the main benefit coming from restoring revision
capacity rather than from a consistent ordering among position selectors.
Limitations include the discrete freeze grid, family-specific horizon
prediction, and evaluation on a limited set of models and constraints.
Overall, CFG is better viewed as an early trajectory corrector than as a
mechanism that must remain active until the final decoding step.

\bibliography{aaai2026}

\clearpage

\appendix

\section{Transport Theory: Derivations and Technical Qualifications}
\label{app:transport}
\label{app:backbone}

The body of Section~\ref{sec:theory} states claims and intuition; this appendix
carries the full derivations, the degenerate cases, and the precise scope of each
claim. It mirrors the three questions of Section~\ref{sec:theory}: A.1--A.2 ground
question 4.1 (what moves the committor), A.3--A.4 make its answer exact and handle
degeneracy (feeding 4.2), and A.5--A.6 carry the variance accounting and the
weak/strong boundary of 4.3.

\subsection{The Martingale Root}
\label{app:A-martingale}

Throughout, $\{\mathcal{F}_t\}$ is the filtration of Assumption~\ref{as:markov} and
$C = \{Y \in \mathcal{A}\}$ is the terminal success event of
Section~\ref{sec:setup:success}, so $\qbase_t(x) = \Ebb^{0}[\mathbf{1}_C \mid
X_t = x]$ and the terminal condition $\qbase_T(Y) = S(Y)$ of
Theorem~\ref{thm:martingale} reads $\qbase_T = \mathbf{1}_C$.

\begin{proof}[Proof of Theorem~\ref{thm:martingale}]
The indicator $\mathbf{1}_C$ is $\mathcal{F}_T$-measurable. By the tower property and
the Markov property (Assumption~\ref{as:markov}),
\[
\begin{aligned}
\Ebb^{0}\big[\qbase_{t+1}(X_{t+1}) \mid X_t = x\big]
&= \Ebb^{0}\big[\,\Ebb^{0}[\mathbf{1}_C \mid X_{t+1}] \mid X_t = x\big] \\
&= \Ebb^{0}[\mathbf{1}_C \mid X_t = x] = \qbase_t(x),
\end{aligned}
\]
which written as a sum over next states is the backward equation
$\qbase_t(x) = \sum_{x'} \Kbase_t(x' \mid x)\,\qbase_{t+1}(x')$. The terminal condition
$\qbase_T = \mathbf{1}_C$ is immediate. Hence $\Ebb^{0}[\qbase_{t+1}(X_{t+1}) \mid
\mathcal{F}_t] = \qbase_t(X_t)$ and $\{\qbase_t(X_t)\}_t$ is a $\Kbase$-martingale.
\end{proof}

\emph{Scope.} The theorem is a statement about the predictable drift of the base
committor under the base next-step law---nothing more. Along individual trajectories
$\qbase$ fluctuates and, on successful ones, ends at $1$; the theorem does not say
base trajectories cannot succeed, only that the base process cannot
\emph{manufacture} success in expectation.

\begin{corollary}[optional stopping]
\label{cor:stopping}
For any stopping time $\tau \le T$---in particular any adapted implementation of the
handoff rule, capped at the horizon---$\Ebb^{0}[\qbase_{\tau}(X_{\tau})] = q_0$.
\end{corollary}
\begin{proof}
$\{\qbase_t(X_t)\}$ is a martingale bounded in $[0,1]$ and $\tau \le T$ is bounded, so
the optional stopping theorem gives $\Ebb^{0}[\qbase_{\tau}(X_{\tau})] =
\Ebb^{0}[\qbase_0(X_0)] = q_0$. The base process cannot, in expectation, manufacture
success it did not already have.
\end{proof}

\begin{corollary}[recovery identity]
\label{cor:recovery}
Under the base continuation, the probability of eventual success after an intervention
$\mathrm{do}(X_t = x')$ is exactly $\qbase_t(x')$.
\end{corollary}
\begin{proof}
Immediate from the definition $\qbase_t(x') = \Pr_{\Kbase}(Y \in \mathcal{A} \mid X_t = x')$:
setting the state to $x'$ and continuing under $\Kbase$ succeeds with probability
$\qbase_t(x')$. This is the interface through which the repairability results of
Section~\ref{sec:results} are read.
\end{proof}

\subsection{CFG and the Doob Transform}
\label{app:A-doob}

\paragraph{Validity of the Doob kernel.} With $h = \qbase$, the kernel
$\Kdoob_t(x' \mid x) = \Kbase_t(x' \mid x)\, \qbase_{t+1}(x')/\qbase_t(x)$ sums to one
over $x'$ precisely by the backward equation of Theorem~\ref{thm:martingale}, so it is
a bona fide kernel wherever $\qbase_t(x) > 0$.

\begin{corollary}[one-step utility of ideal guidance]
\label{cor:utility}
For any state $x$ with $\qbase_t(x) > 0$,
$\Ebb^{C}[\qbase_{t+1} \mid x] - \qbase_t(x) = \Var^{0}(\qbase_{t+1} \mid x)/\qbase_t(x)$.
\end{corollary}
\begin{proof}
The terminal-conditioned (Doob) kernel gives
\[
\begin{aligned}
\Ebb^{C}[\qbase_{t+1} \mid x]
&= \sum_{x'} \Kdoob_t(x' \mid x)\,\qbase_{t+1}(x') \\
&= \frac{1}{\qbase_t(x)}
   \sum_{x'} \Kbase_t(x' \mid x)\,\qbase_{t+1}(x')^2 \\
&= \frac{\Ebb^{0}\!\left[(\qbase_{t+1})^2 \mid x\right]}{\qbase_t(x)}.
\end{aligned}
\]
By Theorem~\ref{thm:martingale}, $\qbase_t(x) = \Ebb^{0}[\qbase_{t+1} \mid x]$, hence
\[
\begin{aligned}
\Ebb^{C}[\qbase_{t+1} \mid x] - \qbase_t(x)
&= \frac{\Ebb^{0}\!\left[(\qbase_{t+1})^2 \mid x\right]
        - \Ebb^{0}[\qbase_{t+1} \mid x]^2}{\qbase_t(x)} \\
&= \frac{\Var^{0}(\qbase_{t+1} \mid x)}{\qbase_t(x)}. \qedhere
\end{aligned}
\]
\end{proof}

\paragraph{Centering is free (gauge).} The guided kernel $\Kguid_t(x' \mid x) \propto
\Kbase_t(x' \mid x)\, e^{w \delta_t(x')}$ is invariant under $\delta_t \mapsto \delta_t
+ c$: the factor $e^{wc}$ cancels between the numerator and the normalizer. Only the
centered direction $\tilde\delta_t = \delta_t - \Ebb[\delta_t]$ is physical, so every
covariance statement below is automatically centered.

\paragraph{Support caveat (no tilt escape).} The exponential tilt reweights within the
support of $\Kbase_t(\cdot \mid x)$ and assigns zero mass outside it, for every $w$.
Guidance can therefore only redistribute probability among next states the base kernel
can already reach; it cannot create transitions. Every transport statement in
Section~\ref{sec:theory} is a statement about reweighting on the base support.

\subsection{Infinitesimal Transport as a Covariance}
\label{app:A-cov}

\paragraph{Linear-response derivation.} The vocabulary $V$, the canvas length $L$,
and the horizon $T$ are finite, so every expectation here is a finite sum, $D(w)$
below is finite and strictly positive for all $w$, and differentiation under the sum
is immediate. We write $\Kguid_t$ for the guided kernel at the fixed weight $w$ under
discussion, dropping the weight index $\Kguid_{t,w}$ of Section~\ref{sec:theory:cov}
except where the $w$-dependence is the object of study; likewise $\Delta q^{g}_t$ for
$\Delta q^{g,w}_t$. Fix a state $x$ and write $q' = \qbase_{t+1}$,
$\delta = \delta_t$, all sums over next states $x'$ weighted by
$\Kbase_t(x' \mid x)$. Let $F(w) = N(w)/D(w)$ with
$N(w) = \sum_{x'} \Kbase_t(x' \mid x)\, e^{w\delta(x')}\, q'(x')$ and
$D(w) = \sum_{x'} \Kbase_t(x' \mid x)\, e^{w\delta(x')}$, so that
$F(w) = \Ebb_{\Kguid_t(\cdot \mid x)}[q']$. Here $\Kguid_t(\cdot \mid x)$ is the
exponential-family tilt of $\Kbase_t(\cdot \mid x)$ with natural parameter $w$,
sufficient statistic $\delta$, and normalizer $D(w)$; thus $F(w)$ is a tilted
expectation, and the identity below is the standard \emph{linear-response} fact that the
derivative of a tilted mean at $w = 0$ is its covariance with the tilt statistic. Then
$F'(w) = [N'(w)D(w) - N(w)D'(w)]/D(w)^2$. At $w = 0$: $D(0) = 1$,
$N(0) = \Ebb[q']$, $N'(0) = \Ebb[\delta q']$, $D'(0) = \Ebb[\delta]$, hence
\begin{equation}
  F'(0) = \Ebb[\delta q'] - \Ebb[q']\,\Ebb[\delta] = \mathrm{Cov}(\delta, q').
\end{equation}
This derivative identity is \textbf{exact}---it is Eq.~\ref{eq:th-cov} of
Section~\ref{sec:theory:cov}. The same quotient-rule computation at general $w$
gives $F'(w) = \mathrm{Cov}_{\Kguid_t(\cdot \mid x)}(\delta, q')$: the transport
rate is a covariance at \emph{every} strength, under the current tilted law;
Eq.~\ref{eq:th-cov} is its $w = 0$ evaluation, and
$\Delta q^{g}_t = \int_0^w F'(s)\,ds$ is the exact integral that the body's
first-order law truncates at $s = 0$.

\paragraph{Finite-$w$ qualification.} Taylor expansion gives $\Delta q^{g}_t(x) =
F(w) - F(0) = w\,\mathrm{Cov}(\delta_t, \qbase_{t+1}) + \tfrac{1}{2}w^2 F''(0) +
\dots$, where $F''(0)$ is the third joint cumulant of $\delta$ against $q'$ under
$\Kbase$ (tilted-mean derivatives generate the cumulant hierarchy). The body's
finite-$w$ statements use only the first-order term and say so; no claim in
Section~\ref{sec:theory} requires the higher-order terms to vanish. The
covariance statement is exact for the rate $F'(0)$ (Eq.~\ref{eq:th-cov}) and
first-order for $\Delta q^{g}_t$ at finite $w$
(Eq.~\ref{eq:th-cov-first-order}), and we never conflate the two.

\subsection{Alignment, Degeneracy, and Failure Modes}
\label{app:A-align}

\paragraph{Committor-aligned/residual decomposition.} The claim of
Section~\ref{sec:theory:cov}---that only the committor-aligned part of $\delta_t$
transports---is an \emph{orthogonal projection} in $L^2(\Kbase_t(\cdot \mid x))$, the
space of functions of the next state under the base-weighted inner product
$\langle f, g \rangle = \Ebb_{\Kbase_t}[fg]$, which on mean-zero functions equals
$\mathrm{Cov}(f, g)$. Equivalently, it is the $\Kbase_t$-weighted least-squares
regression of $\delta_t$ on $q'$. Centering being free (A.2), assume $\Var(q') > 0$,
write $\tilde q' = q' - \Ebb[q']$, and set the regression coefficient
$\beta_t = \langle \tilde\delta_t, \tilde q' \rangle /
\langle \tilde q', \tilde q' \rangle = \mathrm{Cov}(\delta_t, q')/\Var(q')$. Split
\[
  \tilde\delta_t = \underbrace{\beta_t\,\tilde q'}_{\delta^{q}_t}
    + \underbrace{\big(\tilde\delta_t - \beta_t\,\tilde q'\big)}_{\delta^{r}_t}.
\]
The residual is orthogonal to the committor,
\[
\begin{aligned}
  \langle \delta^{r}_t, \tilde q' \rangle
  &= \Ebb\big[(\tilde\delta_t - \beta_t \tilde q')\,\tilde q'\big] \\
  &= \mathrm{Cov}(\delta_t, q') - \beta_t \Var(q') = 0,
\end{aligned}
\]
by the definition of $\beta_t$---the defining property of the least-squares residual.
By the covariance identity of A.3, $\mathrm{Cov}(\delta_t, q') =
\langle \tilde\delta_t, \tilde q' \rangle = \beta_t \Var(q')$, so $\Delta q^{g}_t(x) =
w\,\beta_t\,\Var(q') + O(w^2)$: replacing $\delta_t$ by its residual $\delta^{r}_t$
alone would give \emph{exactly} zero first-order transport. The transport identity
sees $\delta_t$ only through the scalar $\mathrm{Cov}(\delta_t, q')$, so this rank-one
projection onto $q'$ is not a simplification but precisely the information
first-order transport depends on. (The split is taken against the committor $q'$, not
against the conditional direction used by projected-guidance variants, so we
deliberately avoid the parallel/perpendicular labels of that construction.)

\paragraph{Alignment score.} In the same $\Kbase_t(\cdot \mid x)$ geometry, when both
spreads are positive, $\rho_t = \mathrm{Cov}(\delta_t, q')/(\sigma(\delta_t)\,
\sigma(q'))$ is exactly the cosine of the angle between $\delta_t$ and $q'$ under the
covariance inner product, so $\rho_t \in [-1, 1]$ by Cauchy--Schwarz, and the
factorized law of Section~\ref{sec:theory:useful} follows by normalizing the
covariance identity of A.3 by the two standard deviations.

\paragraph{Degenerate cases.} The factorized form requires both spreads positive; the
two collapse modes must be handled separately, and both are \emph{stronger} than
first-order statements:
\begin{itemize}
\item \emph{Tilt collapse, $\sigma(\delta_t) = 0$.} Then $\delta_t$ is constant on the
  support of $\Kbase_t(\cdot \mid x)$, the tilt factor $e^{w\delta_t}$ cancels in the
  normalizer, and $\Kguid_t = \Kbase_t$ \textbf{identically, for every $w$}. Transport
  is exactly zero at all orders, and $\rho_t$ is undefined ($0/0$). This mode is
  \emph{not} misalignment---there is no direction to be misaligned---and any empirical
  estimator of $\rho_t$ must report it as a separate category rather than coerce it to
  $0$.
\item \emph{Outcome collapse, $\sigma(\qbase_{t+1}) = 0$.} Then $q'$ is constant on
  the support, every reweighting has the same mean, and $\Delta q^{g}_t = 0$
  \textbf{exactly, for every $w$} (by Theorem~\ref{thm:martingale} the constant is
  $\qbase_t$). Nothing is left to steer.
\end{itemize}

\paragraph{Exhaustiveness.} At first order, transport vanishes iff
$\mathrm{Cov}(\delta_t, q') = 0$, and the factorization $\mathrm{Cov} =
\rho\,\sigma(\delta)\,\sigma(q')$ (valid whenever both spreads are positive, with the
two collapse modes exhausting the remaining cases) shows this happens iff
$\sigma(\delta_t) = 0$, $\sigma(\qbase_{t+1}) = 0$, or $\rho_t = 0$; transport is
negative iff both spreads are positive and $\rho_t < 0$. This is the complete case
analysis behind the iff criterion of Section~\ref{sec:theory:useful}.

\subsection{Variance Accounting and the Commitment Horizon}
\label{app:A-variance}

\begin{corollary}[Doob decomposition and orthogonality]
\label{cor:doob}
Write the increment $\Delta q_t = \qbase_{t+1}(X_{t+1}) - \qbase_t(X_t) = \Delta A_t +
\Delta M_t$ with predictable part $\Delta A_t = \Ebb^{0}[\Delta q_t \mid \mathcal{F}_t]$
and martingale increment $\Delta M_t = \Delta q_t - \Delta A_t$. Under $\Kbase$,
$\Delta A_t \equiv 0$, and the increments are $L^2$-orthogonal:
$\Ebb^{0}[\Delta M_s \Delta M_t] = 0$ for $s \ne t$.
\end{corollary}
\begin{proof}
By Theorem~\ref{thm:martingale}, $\Ebb^{0}[\Delta q_t \mid \mathcal{F}_t] = 0$, so
$\Delta A_t \equiv 0$ and $\Delta q_t = \Delta M_t$. For $s < t$,
$\Ebb^{0}[\Delta M_s \Delta M_t] = \Ebb^{0}[\Delta M_s\,\Ebb^{0}[\Delta M_t \mid
\mathcal{F}_t]] = 0$ since $\Delta M_s$ is $\mathcal{F}_t$-measurable and
$\Ebb^{0}[\Delta M_t \mid \mathcal{F}_t] = 0$.
\end{proof}

\emph{Division of labor.} Under a guided measure $\Delta A_t$ need not vanish, and
this is the predictable-drift sense in which transport is guidance-exclusive: the
base process contributes only the martingale part, so any predictable drift acquired
by $\qbase_t(X_t)$ is attributable to the guidance intervention, while the variance
channel that resolves the committor to $0$ or $1$ is present under $\Kbase$ already.
Guidance supplies the drift (transport); the model's own martingale supplies the
variance resolution (realization)---with the honest qualifier that the base
martingale resolves to success only once $\qbase$ is near one, so before commitment
it resolves mostly to failure. Under the guided kernel the drift is exactly the
local lift $\Delta q^{g}_t$ of Eq.~\ref{eq:th-local-lift}, which is what the next
proof accumulates.

Summing that guided-measure predictable drift is exact and gives the accumulation
identity of the main text.
\begin{proof}[Proof of Proposition~\ref{prop:transport-sum}]
Under the guided continuation law the next transition from $X_s$ is
$\Kguid_s(\cdot \mid X_s)$, so the predictable part of the base-committor increment
is exactly the local lift:
$\Ebb^{g}[\qbase_{s+1}(X_{s+1}) - \qbase_s(X_s) \mid X_s] = \Delta q^{g}_s(X_s)$.
Taking $\Ebb^{g}[\,\cdot \mid X_t = x]$ and summing from $s = t$ to $T - 1$
telescopes:
\begin{align}
  \Ebb^{g}\!\left[\sum_{s=t}^{T-1} \Delta q^{g}_s(X_s) \,\middle|\, X_t = x\right]
  &= \Ebb^{g}\big[\qbase_T(X_T) \,\big|\, X_t = x\big] - \qbase_t(x) \nonumber \\
  &= \qguid_t(x) - \qbase_t(x) = \Vval_t(x),
\end{align}
where the last line uses the terminal condition $\qbase_T = \mathbf{1}_C$, whose expectation under the
fully guided continuation from $x$ is by definition the guided committor
$\qguid_t(x)$. The identity is exact at every $w$; no first-order approximation is
involved. It lives entirely under the guided law and asserts no relation between
the base-measure budget (Corollary~\ref{cor:budget}) and guided transport beyond
the Cauchy--Schwarz cap below.
\end{proof}

\begin{corollary}[variance budget]
\label{cor:budget}
$\sum_t \Ebb\big[\Var^{0}(\qbase_{t+1} \mid X_t)\big] = q_0(1 - q_0)$.
\end{corollary}
\begin{proof}
By orthogonality (Corollary~\ref{cor:doob}) and $\Delta q_t = \Delta M_t$,
$\Var(\qbase_T) = \Var(\qbase_0) + \sum_t \Ebb[(\Delta M_t)^2]$. Here $\qbase_0 = q_0$
is deterministic so $\Var(\qbase_0) = 0$, and $\qbase_T = \mathbf{1}_C$ is Bernoulli
with mean $q_0$ so $\Var(\qbase_T) = q_0(1 - q_0)$. Since $\Delta M_t = \qbase_{t+1} -
\Ebb^{0}[\qbase_{t+1} \mid X_t]$ has conditional mean zero,
$\Ebb[(\Delta M_t)^2] = \Ebb[\Ebb^{0}[(\Delta M_t)^2 \mid \mathcal{F}_t]] =
\Ebb[\Var^{0}(\qbase_{t+1} \mid X_t)]$. Summing gives the budget.
\end{proof}

\paragraph{Optionality closed form.} Combining Corollaries~\ref{cor:doob}
and~\ref{cor:budget}, the total budget $q_0(1 - q_0)$ decomposes into per-step spends
$\Ebb[\Var^{0}(\qbase_{t+1} \mid X_t)]$; the remaining option value at $X_t$ is the
Bernoulli-terminal form $\qbase_t(1 - \qbase_t)$, which vanishes as $\qbase_t \to 0$ or
$1$. This is the closed form behind the contraction argument of
Section~\ref{sec:theory:astar}: what the horizon exhausts is an option value that
vanishes once the constraint is decided either way.

\paragraph{The Cauchy--Schwarz cap, and what the budget does not say.} Applying
Cauchy--Schwarz to the covariance identity of A.3,
\begin{equation}
  \big|\Delta q^{g}_t(x)\big| \le w\,\sigma(\delta_t)\,\sigma(\qbase_{t+1}) + O(w^2):
\end{equation}
local base committor variance upper-bounds first-order steering value. This inequality
is the \emph{only} bridge we assert between the budget and guided transport. The
budget is an identity for the base martingale under the base measure; $\Delta q^{g}_t$
is defined under the guided tilt; they are not entries in one additive account, and we
make no claim that guided transport ``spends'' base variance step-for-step. What is
true, and what Section~\ref{sec:theory:astar} uses, is that the cap and the switch-off
cost both collapse when the base committor resolves locally
($\sigma(\qbase_{t+1}) \to 0$).

\paragraph{Well-definedness of the interventional horizon.} With $\Ssw(t)$ the terminal
success probability of the process that follows $\Kguid$ before $t$ and $\Kbase$ from
$t$ on (Definition~\ref{def:astar}), $\astar_{\varepsilon} = \inf\{t : \Ssw(s) \ge \Ssw(T) - \varepsilon \ \forall s \ge
t\}$ is well-defined for every $\varepsilon > 0$ (the set contains $t = T$), monotone
in $\varepsilon$ ($\varepsilon' \ge \varepsilon \Rightarrow \astar_{\varepsilon'} \le
\astar_{\varepsilon}$; enlarging the band enlarges the persistent set), and
scale-free (it references only terminal success probabilities, never the raw
magnitude of a single-step lift). The persistence quantifier matters when $\Ssw$ is
non-monotone: a bare first crossing can precede a later excursion below the band, and
every ``past $\astar$'' statement in the paper quantifies over all later switch
points; the estimator of Section~\ref{sec:results:astar} implements exactly this
form (the safety condition is required at every later grid point). Prompts with
$\qbase_0$ near $1$---and, symmetrically, prompts guidance cannot help, for which
$\Ssw$ is flat at a low level---get $\astar_{\varepsilon} = 0$ because guidance
carries no measurable remaining value, not by threshold artifact; prompts for which $\Ssw(t) < \Ssw(T) - \varepsilon$
for all $t < T$ are right-censored at $T$ (Section~\ref{sec:setup:astar}). This is
the population-level object;
Section~\ref{sec:results} estimates it per prompt by the freeze sweep, with the
survivor taxonomy supplying the censoring accounting.

\paragraph{Why no local statistic can replace it.} These three properties are also
why Section~\ref{sec:theory:astar} declines to define the horizon by a threshold on
$\Delta q^{g}_t$. A fixed threshold on the raw lift is not scale-free; it ignores
persistence, so it can fire before a later excursion below the band; and it is
satisfied vacuously at $t = 0$ by any prompt born with $\qbase_0$ near $1$, for which
no step transports anything because there is nothing left to steer. The
interventional definition is immune to all three by construction, since it references
only terminal success probabilities of complete continuation policies.

\paragraph{Ensemble versus state.} Eq.~\ref{eq:setup-vlink} averages the remaining
guidance value over the states the guided prefix induces, so $\astar_{\varepsilon}$
is an ensemble statement at the prompt level. Individual realized states reached at
the same step may still carry very different $\Vval_t(x)$; nothing in the definition
or in the freeze estimator claims a per-trajectory horizon, and no result in
Section~\ref{sec:results} is read at that resolution.

\subsection{Weak and Strong Claims, and the CFG--Doob Bridge}
\label{app:A-weakstrong}

\paragraph{Weak existence (derived).} $\astar_{\varepsilon} \le T$ always, since
switching off at $T$ costs nothing. The nontrivial derived content is the mechanism,
not the bare existence: by A.5, the forgone transport after $t$ is capped by the
remaining outcome-variance profile, so wherever the base committor resolves before
$T$, the switch-off cost---and hence $\astar_{\varepsilon}$---arrives strictly before
the schedule ends. Bare existence is nearly trivial and we claim no more for it.

\paragraph{Strong existence (empirical).} That $\astar_{\varepsilon}$ is early and
sharp is not derivable from the martingale, which may deposit its variance on any
schedule consistent with the budget (the toy model of Appendix~B makes this failure of
derivability concrete: front-loaded and flat hazards obey the same budget with
opposite horizons). The schedule is pinned by the decoder---absorbing unmasking spends
committor variance at the steps where constraint-relevant tokens commit, and deployed
samplers front-load commitment---which is a structural property of the kernel,
measured in Section~\ref{sec:results}, not a consequence of
Assumption~\ref{as:markov}.

\paragraph{The CFG--Doob bridge.} CFG reweights in logit space by
$\exp(w\,\delta_t)$; the ideal transform reweights in committor space by
$\qbase_{t+1}$. The two kernels coincide at state $x$ iff
$w\,\delta_t(x' \mid x) = \log \qbase_{t+1}(x') + c_t(x)$ for all $x'$ in the support
of $\Kbase_t(\cdot \mid x)$, with $c_t(x)$ free of $x'$---the two exponents must match
token by token, up to the additive gauge the normalizer absorbs (A.2). The weaker
reading, that $w\,\delta_t$ merely be affine in $\log \qbase_{t+1}$, leaves the slope
unpinned and is not sufficient: a slope other than one rescales the tilt and gives a
different kernel. Nothing in Section~\ref{sec:theory} assumes
this; the transport identity treats $\delta_t$ as an arbitrary tilt statistic. Whether
$\delta_t$ is in fact a monotone surrogate for the successor committor is an empirical
question---Section~\ref{sec:results} measures it in \emph{both} the base and the
guided gauge and finds per-step alignment confined to a sparse, early subset of
cells---and by the support caveat of A.2, even a perfectly aligned tilt can only
redistribute mass the base kernel already reaches. Whether those sparse early
transport events suffice on their own to account for guidance's end-to-end success
gains is a further question that only targeted intervention can settle; the freeze
results of Section~\ref{sec:results} answer its prefix form (guidance restricted to
the pre-$\astar$ prefix recovers full-guidance success), and the cell-level form is
answered in the negative by the closure test of Section~\ref{sec:results:astar}:
guidance applied only at detected transport steps---or on the event-covering early
window---does not recover full-guidance success on most guidance-dependent prompts,
and the majority of such prompts show no detectable event at all, in shares matching
the tilt-collapse fractions. The detected events are real but insufficient; how the
terminal benefit arises from per-step action that is largely committor-orthogonal
remains open, and we do not close it by assumption.

\subsection{Post-Commitment Stability under Parallel Kernel Substitution}
\label{app:parallel-stability}

This subsection states the formal counterpart of
Section~\ref{sec:results:realization}: handoff safety is definitional, parallel
safety is a conditional theorem whose premise is empirical.

% Moved out of Sec 5.3 on 2026-07-28 (4th pass). The paragraphs below already
% said "the counterexample of Sec 5.3" and cited Sec 5.3 for the churn data;
% the counterexample and the churn numbers now live HERE, so those two
% references are internal. Do not re-point them at the body.
\paragraph{Why the second license is not a corollary of the first.} A
freeze-safe state need not be parallel-safe. Take the constraint ``the output
contains exactly one of $\{X, Y\}$''. Sequential base decoding fills one
position with $X$ and thereafter, conditioned on $X$, never emits $Y$: the
constraint is committed, $\astar$ is reached, and handoff is safe. Parallel
filling draws the two positions from their marginals and can place $X$ and $Y$
together. The committor is flat along every sequentially reachable
continuation yet not flat over the product-support states that only
parallelism reaches, so whatever safety parallelism enjoys past $\astar$ is a
fact about the model's committor field, not a consequence of the horizon's
definition. This is why the horizon caps $\varepsilon_g$ only, and the
$\varepsilon_s$ of the bound below must be measured.

\paragraph{Setup.} Group decoding into macro-steps. At macro-step $s$ the parallel
kernel $K^{P,s}$ reveals a block of $k_s$ positions in a single draw from the
product of conditional marginals; $\bar K^{0,s}$ denotes the time-scale-matched
comparison, the $k_s$-fold composition of the sequential base kernel. Define hybrid
policies $\pi^{(s)}$ that apply parallel macro-steps before $s$ and sequential
macro-steps from $s$ on, so $\pi^{(t)}$ is all-sequential and $\pi^{(T)}$
all-parallel, and adjacent hybrids differ in exactly one macro-step. Write
$q^{\pi}_s(x)$ for the terminal success probability of continuation policy $\pi$
from state $x$, and $\rho_s$ for the state distribution at macro-step $s$ under the
parallel prefix.

\begin{proposition}[Two-hop parallel-safety bound]
\label{prop:twohop}
Fix a state $x$ at macro-step $t$. For each $s \ge t$ let
\begin{equation*}
\varepsilon_s
=
\Ebb_{z \sim \rho_s}\!\Big[
\operatorname{osc}_{\mathcal{R}_s(z)} q^{\pi^0}_{s+1}
\Big],
\end{equation*}
% [2026-07-28] split off R_s(z): the single display ran 78pt past the column.
with the reachable set
\begin{equation*}
\mathcal{R}_s(z)
=
\operatorname{supp} K^{P,s}(\cdot \mid z)
\,\cup\,
\operatorname{supp} \bar K^{0,s}(\cdot \mid z),
\end{equation*}
where $\operatorname{osc}_{\mathcal{R}} f = \sup_{\mathcal{R}} f -
\inf_{\mathcal{R}} f$ and $q^{\pi^0}_{s+1}$ is the value of the all-sequential
continuation. Then
\begin{equation*}
\big| q^{\pi^P}_t(x) - q^{\pi^0}_t(x) \big|
\;\le\;
\sum_{s = t}^{T-1} \varepsilon_s,
\end{equation*}
and if in addition the state satisfies the handoff condition
$\qguid_t(x) - q^{\pi^0}_t(x) \le \varepsilon_g$ (the gap that defines
$\astar_{\varepsilon}$), then
\begin{equation*}
q^{\pi^P}_t(x)
\;\ge\;
\qguid_t(x) - \Big( \varepsilon_g + \sum_{s=t}^{T-1} \varepsilon_s \Big).
\end{equation*}
\end{proposition}

\emph{Proof (hybrid telescoping).} $q^{\pi^P}_t - q^{\pi^0}_t = \sum_{s=t}^{T-1}
\big(q^{\pi^{(s+1)}}_t - q^{\pi^{(s)}}_t\big)$. Adjacent hybrids share the parallel
prefix---hence the same visited distribution $\rho_s$---and the same sequential
tail; they differ only in the kernel applied at macro-step $s$. Conditional on the
visited state $z$, each policy's value is the one-step expectation of the common
tail value $q^{\pi^0}_{s+1}$ under its own kernel; both expectations lie between
the infimum and supremum of that value over $\mathcal{R}_s(z)$, so their difference
is at most $\operatorname{osc}_{\mathcal{R}_s(z)} q^{\pi^0}_{s+1}$. Averaging over
$\rho_s$ bounds the $s$-th telescope term by $\varepsilon_s$; summing gives the
first bound, and adding the handoff gap gives the second. \qed

\paragraph{Supports and matching.} $\mathcal{R}_s(z)$ must contain the \emph{union}
of both supports: flatness on their intersection, or on the sequentially reachable
set alone, does not suffice---the counterexample of
Section~\ref{sec:results:realization} lives exactly on product-support states the
sequential kernel never visits, which is why $\astar$ alone cannot deliver the
conclusion. Position matching ($\bar K^{0,s}$ revealing the same block as
$K^{P,s}$) is not required for validity---the oscillation bound holds for any two
kernels supported in $\mathcal{R}_s(z)$---but it makes $\varepsilon_s$
interpretable, separating the joint-versus-product error from the
position-selection error; practical parallel decoders change both at once, and a
measured $\varepsilon_s$ bundles them.

\paragraph{Why flatness rather than total variation.} Since $q \in [0,1]$, one also
has $|\Ebb_{K^{P}}[q] - \Ebb_{\bar K^{0}}[q]| \le \|K^{P} - \bar
K^{0}\|_{\mathrm{TV}}$, and by Pinsker $\|\cdot\|_{\mathrm{TV}} \le
\sqrt{\mathrm{KL}/2}$ with $\mathrm{KL}$ the residual inter-token coupling
$\mathrm{KL}(\prod_i \text{marginals} \,\|\, \text{joint})$. This route is
sufficient but strictly stronger: it demands that parallel and sequential agree as
\emph{distributions over realizations}. The data refuse it while satisfying
flatness: under post-$\astar$ freezes token identities keep churning far beyond
the horizon---identity agreement is still changing up to $a \approx
0.5$--$0.85$, so the kernels stay far apart in total variation---while success
and fluency are flat past $\astar \approx 0.22$, i.e., the committor is flat
where both kernels land. Their differences are confined to
outcome-equivalent states, which is the mathematical content of \emph{semantic}
commitment: what stabilizes at $\astar$ is the outcome level of the field, not
the realization distribution---the remaining uncertainty is \emph{which
successful realization}, no longer \emph{success or failure}. Outcome fidelity, not
distributional fidelity, is the operative invariance; this is the formal sense in
which the commitment is \emph{semantic}---the flatness hypothesis permits the
realization distribution to move arbitrarily, so long as it moves within a
near-outcome-equivalent region.

\paragraph{Estimation.} The proposition is stated at a single state; the empirical
program estimates its terms in expectation. The freeze sweep over survivors
(Section~\ref{sec:results:astar}) caps $\varepsilon_g$; the two-arm $K$-sweep
(Table~\ref{tab:policy-quant}) tests the \emph{conclusion} directly; and the
gate-position sweep of Appendix~\ref{app:gate-sweep} measures the
\emph{premise}: the paired same-gate difference between $K{=}8$ and $K{=}1$ arms
reads out the accumulated $\varepsilon_s$ interventionally, finding it measurably
nonzero at $\astar$ ($+7.8 \pm 2.5$ points on keywords) and indistinguishable from
zero one grid point later. Premise and conclusion are thus verified independently,
closing the two-hop argument.

\section{Toy Model: an Absorbing-Committor Illustration}
\label{app:toy}

This appendix isolates the minimal ingredients the backbone requires---a bounded
committor martingale, an absorbing commit event, and a variance budget---in a chain
simple enough to inspect by hand. It is a diagnostic \emph{illustration} of the
martingale picture (Section~\ref{sec:theory}), not a generative model of the
Transformer trajectory, and makes no claim about CFG dynamics; its role is to show
that front-loading is a \emph{schedule} property, not a consequence of the martingale
(the weak/strong distinction of Section~\ref{sec:theory:existence}), and that a
downstream coordinate's horizon is set by an upstream commit (feeding
Section~\ref{sec:results}).

\paragraph{A single-coordinate dual-hazard chain.} A constraint coordinate carries
a committor $q_t \in [0,1]$ observed at steps $t = 0, \dots, T$. At each step, with
\emph{commit hazard} $\lambda_t \in [0,1]$ the coordinate absorbs---jumping to $1$ with
probability $q_t$ and to $0$ with probability $1 - q_t$---and otherwise holds at $q_t$:
\begin{equation}
  q_{t+1} =
  \begin{cases}
    \mathrm{Bernoulli}(q_t) & \text{w.p. } \lambda_t \ \text{(absorb)},\\
    q_t & \text{w.p. } 1 - \lambda_t.
  \end{cases}
\end{equation}
By construction $\Ebb[q_{t+1} \mid q_t] = \lambda_t q_t + (1 - \lambda_t) q_t = q_t$, so
$\{q_t\}$ is a martingale (a toy instance of Theorem~\ref{thm:martingale}) with per-step
spent variance $\Var(q_{t+1} \mid q_t) = \lambda_t\, q_t(1 - q_t)$. Summing recovers the
budget of Corollary~\ref{cor:budget}, $\sum_t \Ebb[\lambda_t q_t(1 - q_t)] = q_0(1 -
q_0)$, and the remaining option value $q_t(1 - q_t)$ falls to zero exactly as mass
absorbs. The handoff $\astar$---the step the option value is spent---is therefore set by
the \emph{shape} of $\lambda_t$: a front-loaded hazard (large $\lambda_t$ early) yields
an early, sharp $\astar$; a flat hazard yields a late, diffuse one. Both respect the
same budget, which is precisely why weak existence is derivable but the early-and-sharp
location is not (Section~\ref{sec:theory:existence}).

\paragraph{Two-position coupling.} Let coordinate~1 run the chain above, absorbing
at time $T_1$. Coordinate~2's committor is gated by coordinate~1's committed value: its
initial propensity is high only if coordinate~1 absorbed to success, a one-edge
toy-ization of context coupling. Then coordinate~2's horizon satisfies $\astar(2) =
T_1 + \epsilon$: withdrawing guidance before $T_1$ reverts the upstream commit and drags
coordinate~2 to failure, whereas after $T_1$ coordinate~2 realizes on its own. Thus a
coordinate can be pointwise on-track at its own commit yet have its horizon set by an
\emph{upstream} binding time---the mechanism behind the collapse-versus-hopeless
repairability of Section~\ref{sec:results}.

\section{Census Estimation: Protocol, Selection Audit, and Cross-Fitted Values}
\label{app:census}
% Moved out of Sec 5.1 on 2026-07-28 when that block was condensed for the
% 7-page main text.  tab_taxonomy is load-bearing beyond its own caption: the
% captions of tab_sr_vs_w and tab_itt_bytask both \ref it.
\textbf{The census at the center configuration.}
Table~\ref{tab:taxonomy} gives the per-subtask fate composition summarized by
Figure~\ref{fig:typology}, together with the subtask-level success rate at no
guidance, at the best weight, and at $w{=}4$. Two labellings appear in this
paper and are not interchangeable. The \emph{four-fate} shares here classify a
prompt by how its success responds to guidance weight (needs-CFG / born-in /
harmful / hopeless); the \emph{handoff census} used from
Section~\ref{sec:results:horizon} onwards instead labels each prompt by whether
a freeze point exists at all (handoff / persistent-dependent / failure). COLLIE
c06a is the clearest case of the difference: its modal \emph{fate} is needs-CFG
($46.5\%$), while under the handoff labelling it is $90\%$ failure, $10\%$
persistent-dependent and $0\%$ handoff --- guidance measurably moves these
prompts, and they still never reach a state from which it can be switched off.
Table~\ref{tab:dream-fates} repeats the fate classification on Dream-7B
(fate-level only: that sweep carries no freeze branches, so no $\astar$ is
estimated for Dream), and Table~\ref{tab:ppl-vs-w} reports the fluency sweep.
% Extracted from sec_results.tex 2026-07-28 when Sec 5.1 was condensed for the
% 7-page main text; it now lives in Appendix app:census.  Referenced from the
% captions of tab_sr_vs_w and tab_itt_bytask, so it must stay somewhere.
\begin{table*}[t]
\centering
\small
\begin{tabular}{l cccc l ccccc}
\toprule
 & \multicolumn{4}{c}{Success rate} & & \multicolumn{5}{c}{Per-prompt fate share (\%)} \\
\cmidrule(lr){2-5} \cmidrule(lr){7-11}
Subtask & $w{=}0$ & $w_{\text{best}}$ & at $w_{\text{best}}$ & $w{=}4$ &
Modal fate & needs & born-in & harmful & hopeless & resid. \\
\midrule
CommonGen $k{=}3$ & 0.69 & 0.5 & 0.86 & 0.77 & needs-CFG & 58 & 34.5 & 7 & 0.5 & -- \\
CommonGen $k{=}4$ & 0.53 & 0.5 & 0.78 & 0.72 & needs-CFG & 71.5 & 23 & 4.5 & 1 & -- \\
CommonGen $k{=}5$ & 0.32 & 1 & 0.66 & 0.60 & needs-CFG & 88 & 8 & 3 & 1 & -- \\
CommonGen $k{=}10$ & 0.38 & 2 & 0.73 & 0.63 & needs-CFG & 93 & 3.5 & 1.5 & 1.5 & 0.5 \\
\addlinespace
IFEval kw-freq & 0.84 & 1 & 0.98 & 0.84 & born-in & 27.5 & 43.5 & 29 & -- & -- \\
IFEval letter-freq & 0.61 & 2.5 & 0.70 & 0.67 & born-in & 27.5 & 41.5 & 12.5 & 13 & 5.5 \\
IFEval capital-freq & 0.51 & 2 & 0.73 & 0.69 & needs-CFG & 52 & 40 & 3.5 & 1.5 & 3 \\
IFEval no-comma & 0.49 & 3.5 & 0.89 & 0.88 & needs-CFG & 90.5 & 8.5 & 0.5 & -- & 0.5 \\
\addlinespace
CCG length & 0.33 & 0.5 & 0.88 & 0.56 & needs-CFG & 82.5 & 8.5 & 8.5 & 0.5 & -- \\
CCG multi & 0.60 & 1.5 & 0.87 & 0.79 & needs-CFG & 76 & 20 & 4 & -- & -- \\
CCG sentiment & 0.59 & 0.5 & 0.67 & 0.58 & needs-CFG & 65.5 & 20 & 13.5 & 1 & -- \\
\addlinespace
COLLIE c07 & 0.89 & 1 & 0.90 & 0.72 & born-in & 29.5 & 54 & 16.5 & -- & -- \\
COLLIE c06a & 0.08 & 1 & 0.14 & 0.01 & needs-CFG & 46.5 & -- & 16.5 & 26.5 & 10.5 \\
\bottomrule
\end{tabular}
\caption{Cross-benchmark census at the center configuration (64 generated
tokens, 64 steps): all 13 subtasks from the four benchmark families
(CommonGen, IFEval, our CCG suite, COLLIE), $n{=}200$ prompts per subtask,
$N{=}24$ fresh rollouts per (prompt, $w$) cell over
$w \in \{0, 0.5, \dots, 4\}$. Left block: subtask-level success rate at no
guidance, at the best guidance weight, and at $w{=}4$. Right block:
per-prompt fate shares under the four-fate classification
(gap threshold $\tau{=}0.1$, high-base threshold $0.8$); \emph{modal fate}
is the largest share. Residual = prompts fitting no fate template. Every
subtask mixes fates internally (banding), so subtask-level curves are
mixtures; COLLIE c07 and c06a anchor the born-in and hopeless ends of the
spectrum.}
\label{tab:taxonomy}
\end{table*}

% Auto-generated by analysis_utilities/injection_sharpening/make_dream_fate_table.py
% from results/ofat/TAXONOMY.json and results/dream/ofat/ -- do not edit by hand.
\begin{table*}[t]
\centering
\scriptsize
\setlength{\tabcolsep}{3.5pt}
\begin{tabular}{l rrrrr rrrrr}
\toprule
 & \multicolumn{5}{c}{LLaDA-8B (\%)} & \multicolumn{5}{c}{Dream-7B (\%)} \\
\cmidrule(lr){2-6} \cmidrule(lr){7-11}
Subtask & need & born & harm & hope & res. & need & born & harm & hope & res. \\
\midrule
CommonGen kw $k{=}3$ & 58 & 34 & 7 & 0 & 0 & 71 & 16 & 8 & 2 & 4 \\
CommonGen kw $k{=}4$ & 72 & 23 & 4 & 1 & 0 & 77 & 9 & 9 & 2 & 3 \\
CommonGen kw $k{=}5$ & 88 & 8 & 3 & 1 & 0 & 78 & 6 & 8 & 6 & 0 \\
CommonGen kw $k{=}10$ & 93 & 4 & 2 & 2 & 0 & 96 & 4 & 0 & 0 & 0 \\
CCG length & 82 & 8 & 8 & 0 & 0 & 92 & 2 & 6 & 0 & 0 \\
CCG sentiment & 66 & 20 & 14 & 1 & 0 & 64 & 18 & 12 & 4 & 1 \\
CCG multi & 76 & 20 & 4 & 0 & 0 & 90 & 8 & 3 & 0 & 0 \\
IFEval kw-freq & 28 & 44 & 29 & 0 & 0 & 35 & 35 & 30 & 0 & 0 \\
IFEval letter-freq & 28 & 42 & 12 & 13 & 6 & 52 & 24 & 12 & 6 & 5 \\
IFEval capital-freq & 52 & 40 & 4 & 2 & 3 & 55 & 25 & 10 & 8 & 2 \\
IFEval no-comma & 90 & 8 & 0 & 0 & 0 & 34 & 66 & 0 & 0 & 0 \\
COLLIE c07 & 30 & 54 & 16 & 0 & 0 & 28 & 62 & 10 & 0 & 0 \\
COLLIE c06a & 46 & 0 & 16 & 26 & 10 & 48 & 0 & 24 & 6 & 22 \\
\midrule
All subtasks & 62 & 23 & 9 & 4 & 2 & 63 & 21 & 10 & 3 & 3 \\
\bottomrule
\end{tabular}
\caption{Four-fate shares (\% of $n{=}200$ prompts per subtask and model) under the guidance-weight sweep at the center configuration ($w \in \{0,\dots,4\}$, nine levels; 64 generated tokens, 64 steps, $N{=}24$ rollouts per cell), classified per prompt by the pre-registered rule of the LLaDA census (needs-CFG: best positive-$w$ gain $> 0.1$; born-in: base $\ge 0.8$ and flat; harmful: loss $< -0.1$ at $w{=}4$ without a compensating gain; hopeless: at most one success at every $w$; residual: remainder). LLaDA aggregate row spans all 13 subtasks; the Dream aggregate spans the subtasks with landed data. }
\label{tab:dream-fates}
\end{table*}

% Auto-generated by analysis_utilities/injection_sharpening/make_ppl_w_table.py
% from results/fluency_ppl_w/*.json -- do not edit by hand.
\begin{table*}[t]
\centering
\small
\begin{tabular}{l rrrrrrrrr}
\toprule
Subtask & $w{=}0$ & $0.5$ & $1$ & $1.5$ & $2$ & $2.5$ & $3$ & $3.5$ & $4$ \\
\midrule
CommonGen kw $k{=}3$ & 43 & 32 & 37 & 41 & 45 & 59 & 63 & 94 & 124 \\
CommonGen kw $k{=}4$ & 43 & 35 & 42 & 44 & 52 & 67 & 75 & 91 & 128 \\
CommonGen kw $k{=}5$ & 38 & 30 & 36 & 43 & 50 & 62 & 77 & 96 & 110 \\
CommonGen kw $k{=}10$ & 40 & 32 & 37 & 45 & 51 & 61 & 81 & 94 & 128 \\
\addlinespace
CCG length & 255 & 121 & 125 & 159 & 257 & 403 & 675 & 1278 & 1364 \\
CCG sentiment & 84 & 31 & 36 & 46 & 53 & 64 & 70 & 92 & 132 \\
CCG multi & 75 & 35 & 40 & 47 & 55 & 67 & 79 & 89 & 130 \\
\bottomrule
\end{tabular}
\caption{Fluency versus guidance weight: corpus-level GPT-2 perplexity of
fresh generations at the census center configuration (64 generated tokens, 64
steps; $n{=}100$ prompts $\times$ 4 rollouts per cell, seeds disjoint from
the census). The $w{=}0$ column is the base (unguided) arm. Two structural
facts: fluency is \emph{not} monotone in guidance---on 7 of
7 subtasks the minimum sits at $w{=}0.5$, and on the CCG families
\emph{weak guidance is more fluent than no guidance} (sentiment: $84 \to
31$; length: $255 \to 121$), so the base arm is not the fluency ceiling;
past $w \approx 1$ perplexity rises, reaching $3.6$--$11.2\times$
the minimum at $w{=}4$ (length degrades most, consistent with its harmful
overshoot in Table~\ref{tab:sr-vs-w}). The same runs replicate the census
success curves out of sample: the maximum per-cell deviation from the
$n{=}200 \times 24$ census values is $\le 0.06$ across all
subtasks and weights. PPL is a fluency proxy on a separate axis from
constraint success.}
\label{tab:ppl-vs-w}
\end{table*}

% (duplicate \input{tab_ofat_fates} removed 2026-07-28 -- the table is
%  input once, in "Per-Configuration Fate Tables" below.)
% Auto-generated by analysis_utilities/injection_sharpening/make_policy_quant_tables.py
% from results/ofat/summary/*.json -- do not edit by hand.
\begin{table*}[t]
\centering
\small
\begin{tabular}{l rrrrrrrrr}
\toprule
Subtask & $w{=}0$ & $0.5$ & $1$ & $1.5$ & $2$ & $2.5$ & $3$ & $3.5$ & $4$ \\
\midrule
CommonGen kw $k{=}3$ & 0.69 & 0.86 & 0.77 & 0.81 & 0.84 & 0.78 & 0.74 & 0.79 & 0.77 \\
CommonGen kw $k{=}4$ & 0.53 & 0.78 & 0.70 & 0.73 & 0.74 & 0.72 & 0.67 & 0.69 & 0.72 \\
CommonGen kw $k{=}5$ & 0.32 & 0.65 & 0.66 & 0.66 & 0.62 & 0.61 & 0.60 & 0.62 & 0.60 \\
CommonGen kw $k{=}10$ & 0.38 & 0.64 & 0.69 & 0.71 & 0.73 & 0.72 & 0.68 & 0.65 & 0.63 \\
\addlinespace
CCG length & 0.33 & 0.88 & 0.79 & 0.77 & 0.81 & 0.76 & 0.69 & 0.65 & 0.57 \\
CCG sentiment & 0.59 & 0.67 & 0.65 & 0.64 & 0.64 & 0.62 & 0.62 & 0.60 & 0.58 \\
CCG multi & 0.60 & 0.80 & 0.84 & 0.87 & 0.87 & 0.84 & 0.80 & 0.81 & 0.79 \\
\addlinespace
IFEval kw-freq & 0.84 & 0.97 & 0.98 & 0.95 & 0.93 & 0.92 & 0.89 & 0.84 & 0.84 \\
IFEval letter-freq & 0.61 & 0.67 & 0.69 & 0.70 & 0.70 & 0.70 & 0.69 & 0.69 & 0.67 \\
IFEval capital-freq & 0.51 & 0.67 & 0.71 & 0.72 & 0.73 & 0.71 & 0.71 & 0.69 & 0.69 \\
IFEval no-comma & 0.49 & 0.69 & 0.79 & 0.81 & 0.80 & 0.82 & 0.88 & 0.89 & 0.88 \\
\addlinespace
COLLIE c07 & 0.89 & 0.88 & 0.90 & 0.90 & 0.83 & 0.83 & 0.84 & 0.79 & 0.72 \\
COLLIE c06a & 0.08 & 0.12 & 0.14 & 0.03 & 0.04 & 0.00 & 0.00 & 0.00 & 0.00 \\
\bottomrule
\end{tabular}
\caption{Subtask-level success rate at every guidance weight of the census
grid (center configuration: 64 generated tokens, 64 steps; $n{=}200$
prompts, $N{=}24$ rollouts per cell). Each row is a mixture over the fates
of Table~\ref{tab:taxonomy}, so subtask-level curves understate within-family
structure; the high-$w$ decline visible on most rows is the harmful fate
(overshoot) gaining share, not a uniform degradation.}
\label{tab:sr-vs-w}
\end{table*}

% Auto-generated by analysis_utilities/injection_sharpening/make_policy_quant_tables.py
% from results/parallel_k/PARALLEL_K_HARVEST.json + ASTAR_OFAT_HARVEST.json -- do not edit by hand.
\begin{table*}[t]
\centering
\small
\setlength{\tabcolsep}{4pt}
\begin{tabular}{l r c rr rr rr rr rr}
\toprule
 & & & \multicolumn{2}{c}{Full CFG} &
\multicolumn{8}{c}{handoff at $a^{*}$, then parallel-$K$} \\
\cmidrule(lr){4-5} \cmidrule(lr){6-13}
 & & & & & \multicolumn{2}{c}{$K{=}1$} & \multicolumn{2}{c}{$K{=}4$} &
\multicolumn{2}{c}{$K{=}8$} & \multicolumn{2}{c}{$K{=}16$} \\
\cmidrule(lr){6-7} \cmidrule(lr){8-9} \cmidrule(lr){10-11} \cmidrule(lr){12-13}
Subtask & $n$ (handoff) & $a^{*}$ med.\ [IQR] & SR & PPL & SR & PPL & SR & PPL & SR & PPL & SR & PPL \\
\midrule
CommonGen kw $k{=}3$ & 78 & 0.30 [0.15, 0.45] & 0.95 & 56 & 0.79 & 50 & 0.80 & 76 & 0.77 & 125 & 0.74 & 261 \\
CommonGen kw $k{=}4$ & 80 & 0.30 [0.22, 0.45] & 0.91 & 64 & 0.80 & 64 & 0.78 & 72 & 0.76 & 115 & 0.70 & 226 \\
CommonGen kw $k{=}5$ & 87 & 0.22 [0.15, 0.45] & 0.93 & 54 & 0.75 & 36 & 0.75 & 53 & 0.73 & 83 & 0.69 & 154 \\
CommonGen kw $k{=}10$ & 108 & 0.45 [0.30, 0.45] & 0.92 & 55 & 0.82 & 40 & 0.79 & 53 & 0.75 & 76 & 0.67 & 135 \\
\addlinespace
CCG length & 84 & 0.15 [0.14, 0.30] & 0.96 & 579 & 0.88 & 433 & 0.90 & 466 & 0.89 & 610 & 0.88 & 807 \\
CCG sentiment & 49 & 0.22 [0.10, 0.45] & 0.89 & 180 & 0.82 & 101 & 0.79 & 158 & 0.77 & 204 & 0.72 & 372 \\
CCG multi & 129 & 0.30 [0.22, 0.45] & 0.95 & 63 & 0.83 & 53 & 0.81 & 76 & 0.77 & 119 & 0.71 & 223 \\
\addlinespace
IFEval kw-freq & 58 & 0.22 [0.15, 0.45] & 0.99 & 41 & 0.88 & 25 & 0.90 & 37 & 0.92 & 59 & 0.89 & 121 \\
IFEval letter-freq & 42 & 0.22 [0.15, 0.30] & 0.99 & 165 & 0.94 & 38 & 0.94 & 56 & 0.94 & 82 & 0.91 & 190 \\
IFEval capital-freq & 35 & 0.30 [0.15, 0.60] & 0.94 & 66 & 0.76 & 40 & 0.73 & 62 & 0.70 & 97 & 0.67 & 192 \\
IFEval no-comma & 147 & 0.22 [0.15, 0.45] & 0.89 & 86 & 0.80 & 43 & 0.82 & 68 & 0.84 & 101 & 0.79 & 211 \\
\addlinespace
COLLIE c07 & 8 & 0.30 [0.30, 0.30] & 0.99 & 206 & 0.98 & 162 & 0.98 & 197 & 0.97 & 340 & 0.94 & 628 \\
\midrule
pooled & 905 & -- & 0.93 & 123 & 0.82 & 85 & 0.82 & 107 & 0.80 & 153 & 0.76 & 264 \\
\bottomrule
\end{tabular}
\caption{Post-commitment parallel realization by subtask: guidance runs until
each prompt's $a^{*}$, then decoding hands off to the base kernel committing
$K$ positions per step ($K{=}1$ is pure sequential handoff). Evaluated on
fresh trajectories over the handoff survivors of the $w{=}2$ census ($n$ per
subtask as shown; pooled $n{=}905$). $a^{*}$ is the survivors-only median
[IQR] at the center configuration. PPL is GPT-2 perplexity of the generated
text, a fluency proxy on a separate axis from success: success degrades
gracefully in $K$ while fluency pays the parallelism tax
(Section~6.2). The Full CFG columns are the sequential full-guidance arm on
the same prompts; its gap to the $K{=}1$ handoff
column is dominated by survivor-selection regression on fresh trajectories,
not by guidance removal, and the same parallelism applied \emph{without}
handoff costs far more success (e.g.\ length $K{=}8$: $0.40$ versus $0.89$
with handoff; Table~\ref{tab:policy-quant} and Section~6.2).}
\label{tab:parallel-k-bytask}
\end{table*}

% tab_itt_bytask was \ref'd from Sec 5 but never \input anywhere (pre-existing
% broken reference, found 2026-07-28); it belongs with the per-subtask tables.
% Auto-generated by analysis_utilities/injection_sharpening/make_itt_policy_table.py
% from results/parallel_k_census/ITT_PARALLEL_K_HARVEST.json -- do not edit by hand.
\begin{table*}[t]
\centering
\small
\setlength{\tabcolsep}{5pt}
\begin{tabular}{l c rrrr rrrr}
\toprule
 & & \multicolumn{4}{c}{Full CFG} &
\multicolumn{4}{c}{Handoff at $\hat a^{*}$ (ITT)} \\
\cmidrule(lr){3-6} \cmidrule(lr){7-10}
Subtask & Coverage & $K{=}1$ & $4$ & $8$ & $16$ & $K{=}1$ & $4$ & $8$ & $16$ \\
\midrule
CommonGen kw $k{=}3$ & 0.39 & 0.84 & 0.86 & 0.84 & 0.77 & 0.78 & 0.78 & 0.77 & 0.76 \\
CommonGen kw $k{=}4$ & 0.40 & 0.74 & 0.80 & 0.75 & 0.72 & 0.69 & 0.69 & 0.68 & 0.66 \\
CommonGen kw $k{=}5$ & 0.43 & 0.62 & 0.65 & 0.63 & 0.57 & 0.54 & 0.55 & 0.53 & 0.52 \\
CommonGen kw $k{=}10$ & 0.54 & 0.74 & 0.80 & 0.78 & 0.47 & 0.68 & 0.67 & 0.64 & 0.60 \\
\addlinespace
CCG length & 0.42 & 0.81 & 0.56 & 0.38 & 0.26 & 0.78 & 0.79 & 0.79 & 0.78 \\
CCG sentiment & 0.24 & 0.64 & 0.52 & 0.47 & 0.43 & 0.62 & 0.61 & 0.61 & 0.60 \\
CCG multi & 0.65 & 0.87 & 0.78 & 0.73 & 0.57 & 0.80 & 0.78 & 0.76 & 0.72 \\
\addlinespace
IFEval kw-freq & 0.29 & 0.93 & 0.74 & 0.64 & 0.67 & 0.90 & 0.90 & 0.91 & 0.90 \\
IFEval letter-freq & 0.21 & 0.70 & 0.66 & 0.69 & 0.72 & 0.69 & 0.69 & 0.69 & 0.68 \\
IFEval capital-freq & 0.17 & 0.73 & 0.69 & 0.75 & 0.77 & 0.70 & 0.69 & 0.69 & 0.68 \\
IFEval no-comma & 0.73 & 0.80 & 0.97 & 0.99 & 0.99 & 0.74 & 0.75 & 0.76 & 0.73 \\
\addlinespace
COLLIE c07 & 0.04 & 0.96 & 0.96 & 0.98 & 1.00 & 0.96 & 0.96 & 0.96 & 0.96 \\
COLLIE c06a & 0.00 & 0.04 & 0.12 & 0.11 & 0.06 & 0.04 & 0.04 & 0.04 & 0.04 \\
\midrule
pooled & 0.35 & 0.72 & 0.70 & 0.67 & 0.62 & 0.69 & 0.68 & 0.68 & 0.66 \\
\bottomrule
\end{tabular}
\caption{Per-subtask intention-to-treat success (all $n{=}200$ census
prompts per subtask, $N{=}24$ fresh rollouts each; pooled row over
$n{=}2600$). Coverage is the share of prompts carrying a recorded
handoff horizon, i.e.\ the share on which the handoff policy actually switches;
on the remainder it decodes the serial full-CFG fallback, so its $K$-columns
flatten towards the fallback as coverage falls. COLLIE c06a has zero
coverage---no prompt in that family ever reaches a persistent safe point---so
the two policies coincide there by construction, and its row is a null control
rather than a result. Coverage varies by an order of magnitude across
subtasks, which is the deployment-side reading of the dependence typology of
Table~\ref{tab:taxonomy}: it is the fraction of the prompt distribution on
which guidance becomes removable at all. \textbf{Read the full-CFG columns
with the fluency axis.} On letter-freq, capital-freq, no-comma, c07, c06a the full-CFG arm scores \emph{higher}
at $K{=}16$ than at $K{=}1$ while its perplexity simultaneously worsens:
these are avoidance-type or already-satisfied constraints that degraded text
meets trivially, so their rise is a scorer artifact rather than a benefit of
parallelism. Success alone is not quality-preserving under parallel decoding
on such families, and the pooled $\Delta_K$ of
Table~\ref{tab:policy-itt} is conservative for exactly this reason.}
\label{tab:itt-bytask}
\end{table*}

\begin{figure*}[t]
\centering
\includegraphics[width=0.32\textwidth]{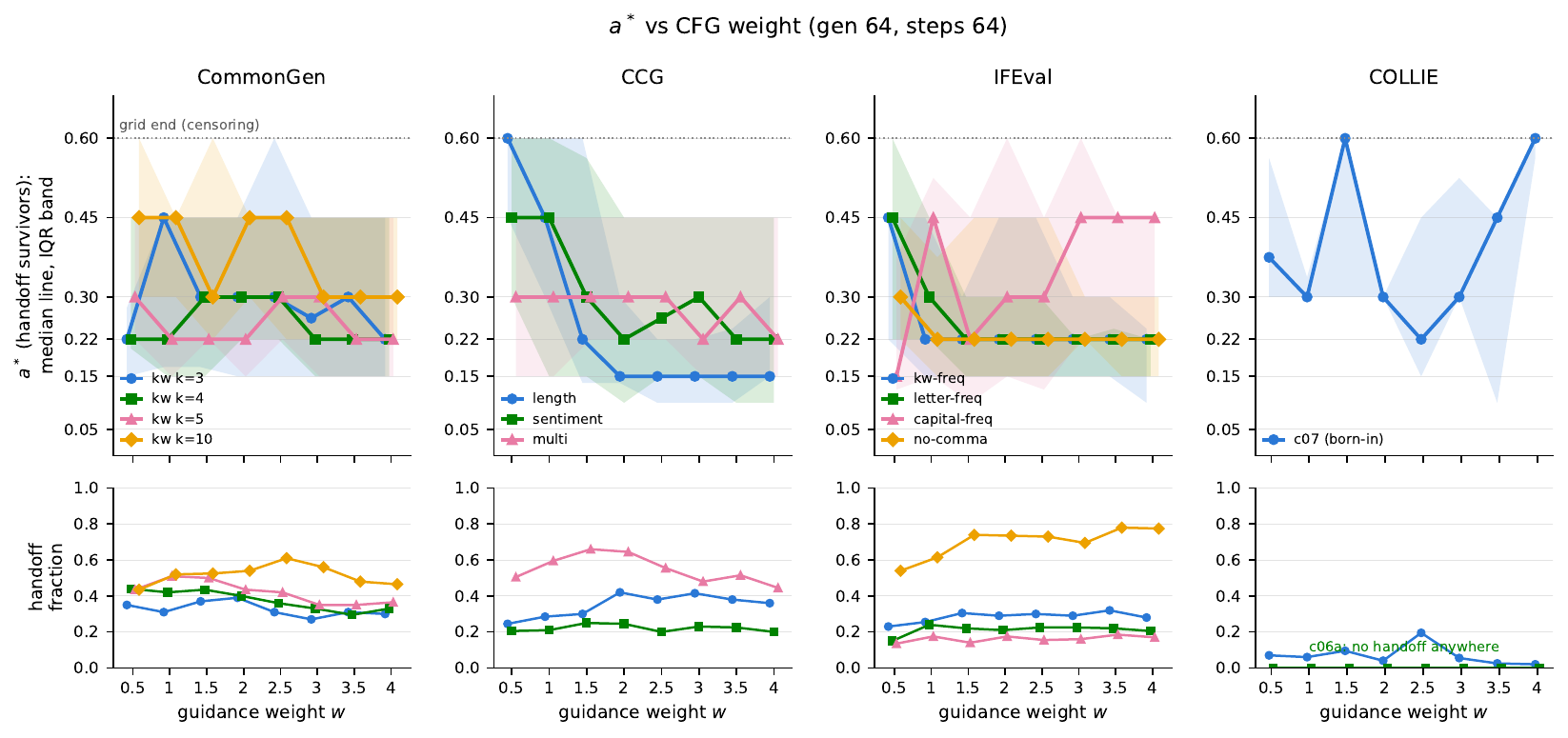}\hfill
\includegraphics[width=0.32\textwidth]{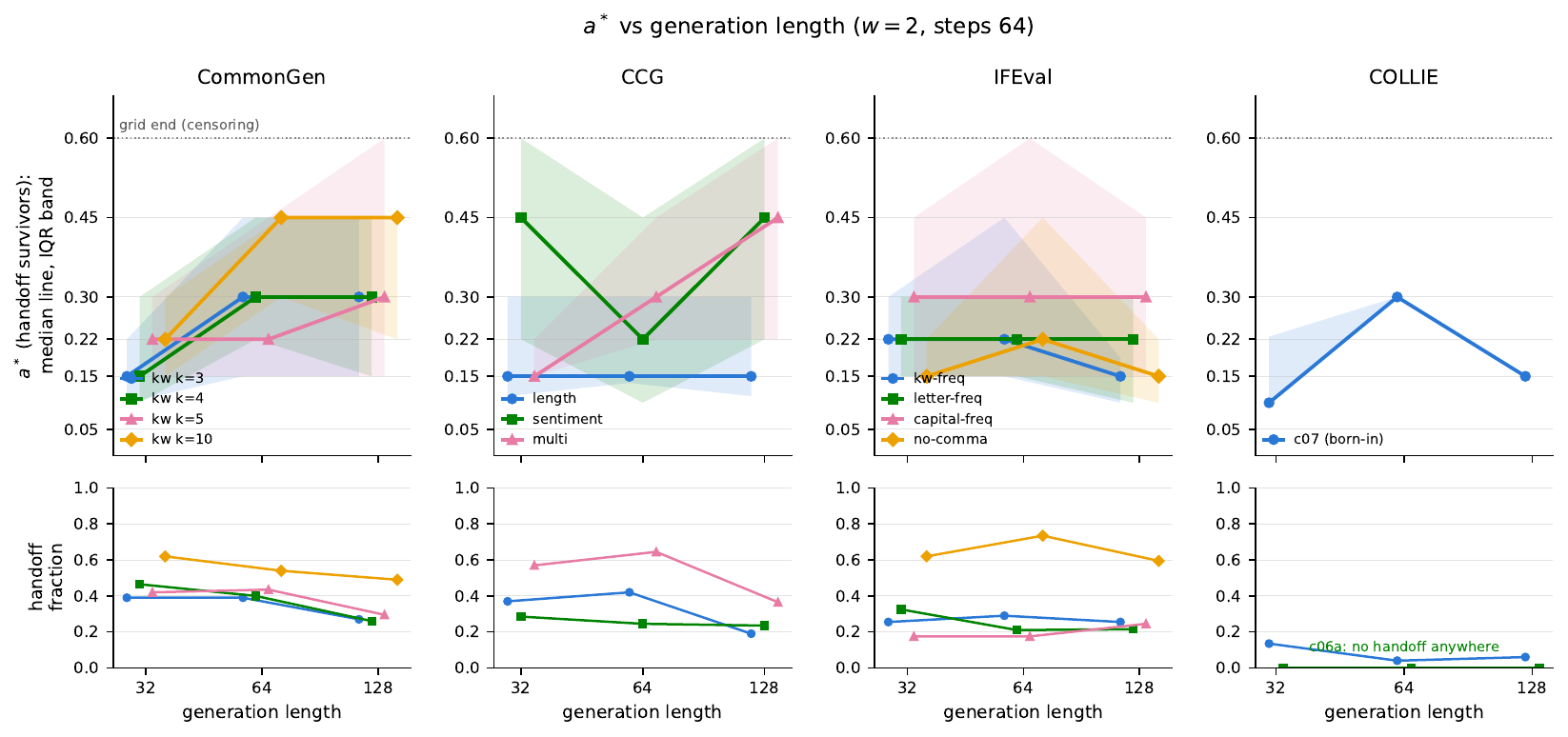}\hfill
\includegraphics[width=0.32\textwidth]{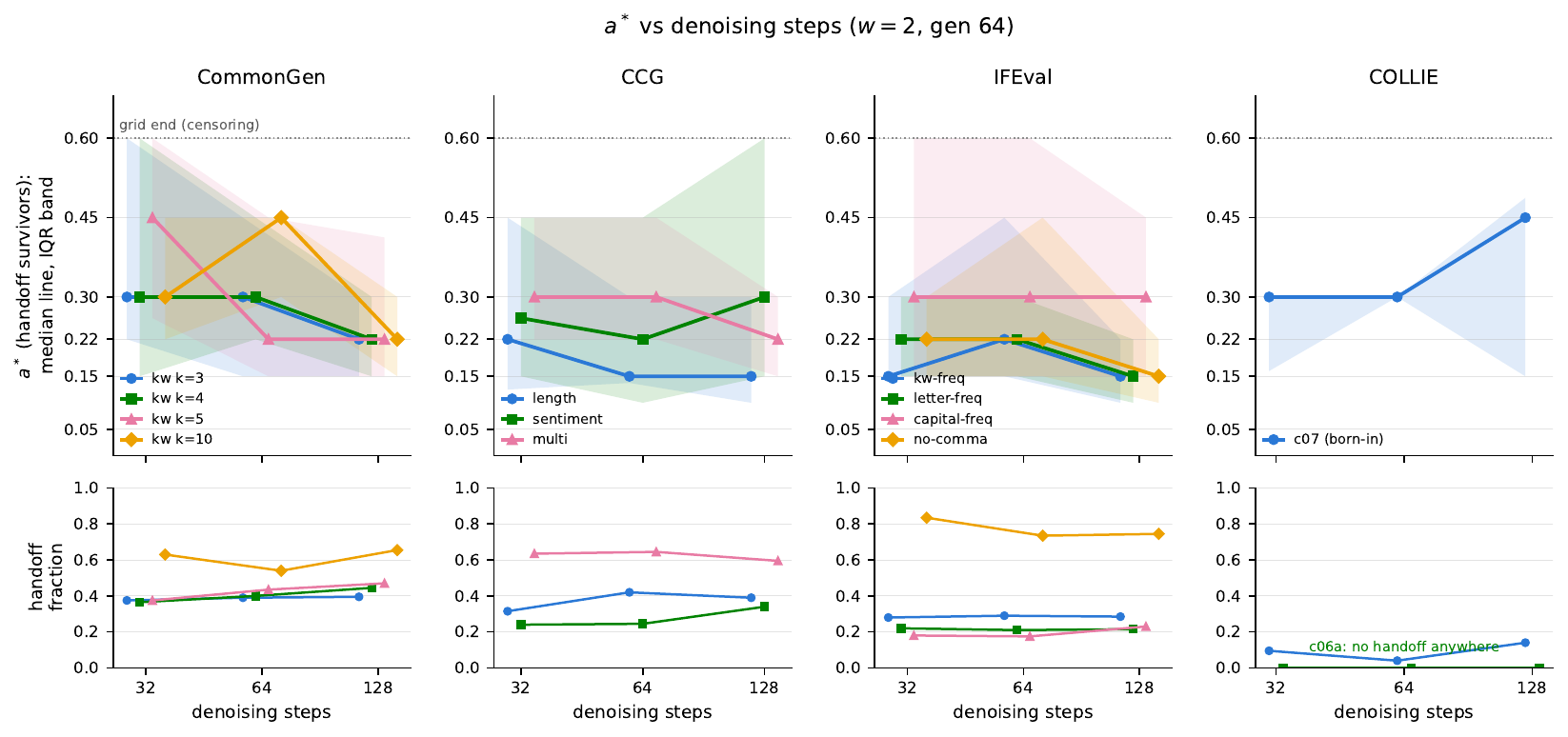}
\caption{One-factor-at-a-time census, three axes: survivors-only median
$a^{*}$ with IQR band and handoff share versus guidance weight $w$ (left),
generation length (middle), and denoising steps (right), for the 13 subtasks
grouped by benchmark. The horizon does not track $w$ for $w \ge 1.5$; it
scales with the realization budget where the carrier is lexical; the steps
axis moves no subtask by more than one grid point.}
\label{fig:astar-ofat}
\end{figure*}

\subsection{The Freeze Estimator}

Per prompt, one guided trajectory is recorded and its state reconstructed at each
grid fraction $f \in \{0.05, 0.1, 0.15, 0.22, 0.3, 0.45, 0.6\}$. From each state we
run $N = 24$ guidance-free continuations (their terminal success fraction estimates
$\qbase(f)$, the freeze-at-$f$ outcome) and $N = 24$ guided continuations
(estimating $\qguid(f)$), with independent seeds per cell. The horizon estimate is
the earliest grid point from which $\qbase \ge 0.9$ and $\qguid - \qbase \le 0.1$
hold at \emph{every} later grid point (the persistent form of A.5); the handoff
class additionally requires guidance to have helped earlier ($\qguid - \qbase >
0.1$ at some prior grid point). Because all freeze branches share the recorded
trajectory, the per-prompt $\astar$ is a trajectory-level estimate of the
ensemble-averaged horizon defined in Section~\ref{sec:setup:astar}, and carries
that definition's caveat: individual realized states can hold very different
$\Vval_t(x)$.

\subsection{Post-Selection Optimism and Cross-Fitting}

Adding error bars does not correct the selection problem, because the issue is not
merely sampling variance at a pre-specified freeze point. The same noisy rollout
estimates are used to select the earliest passing point (and, for the oracle of
Section~\ref{sec:results:astar}, each prompt's best-looking point) and to evaluate
its apparent performance, inducing post-selection optimism; naive intervals
computed after selection do not account for it and can substantially undercover.
Averaging over more prompts reduces random variation around the optimistic estimate
but does not remove the optimism while the per-candidate rollout budget stays
fixed. We therefore separate selection from evaluation by cross-fitting and report
uncertainty only on held-out evaluation outcomes.

Rollouts within a cell are i.i.d.\ with independent seeds across cells, so
conditional on a cell's success count $k$ of $N$, a random half-split is exactly
hypergeometric: no per-rollout storage is needed to simulate it. Per replicate
($R = 200$), every cell is split into a selection half and an evaluation half; the
unchanged taxonomy and selection rules run on the selection half; values are read
on the evaluation half; both the per-prompt oracle and the global-cut comparator
undergo the same isolation. The selection-versus-evaluation difference at selected
points is reported as \emph{observed post-selection optimism} (an empirical
estimate, not the exact bias: both halves carry sampling noise, and 12-rollout
selection is noisier than the deployed 24-rollout rule, so the cross-fitted policy
value is mildly conservative).

Audited census (48-token budget; plug-in $\to$ cross-fitted [2.5, 97.5
percentile]): oracle-vs-global gap $12.6 \to 9.5\ [8.4, 10.6]$ (keywords), $12.7
\to 11.9\ [11.1, 12.6]$ (length), $18.2 \to 13.4\ [11.9, 15.1]$ (sentiment);
oracle-level optimism $3.6 / 0.9 / 6.2$ points; freeze-at-$\astar$ value on
evaluation halves $0.968 / 0.993 / 0.967$ (optimism $1.7 / 0.4 / 1.9$ points). The
64-token wave repeats the pattern: gaps $17.6 \to 14.1$, $8.0 \to 6.2$, $19.1 \to
15.0$; freeze optimism $0.7$--$2.1$ points. Cross-fitting corrects selected
\emph{values}, not selected \emph{locations}; under half-rollout selection the
survivors-only medians move by at most one grid point (e.g., keywords-64: $0.45$
plug-in vs $0.38\ [0.30, 0.45]$), which bounds the location sensitivity we do not
correct.

% Moved out of Sec 5.3 on 2026-07-28 when that block was condensed for the
% 7-page main text: the 48-token paired contrast and the per-subtask survivors
% subgroup.  UPDATE 2026-07-28 (2nd cut): the body no longer mentions either --
% it states only the 13/13 result at eps=0.03 and lets Table 1's caption carry
% "primary estimand = full census; subgroup and replication in this appendix".
% So this paragraph is now the ONLY place the subgroup appears. Keep it
% self-contained; do not shorten it on the assumption the body sets it up.
\textbf{Prespecification of the noninferiority test.} Table~\ref{tab:noninferiority}
reports a test whose margin and estimand were fixed before the contrast was
read, and the caption calls the margin prespecified; the protocol is collected
here so that claim is checkable in one place. (i) Margin: $\varepsilon = 0.03$
in absolute success rate, with $0.05$ retained as the sensitivity margin for
subgroup analyses. (ii) Primary estimand: the paired handoff-minus-full-CFG
difference over the \emph{full} $n{=}200$ prompt census of each subtask, with
right-censored prompts ($a^{*} = T$) kept in the denominator at a paired
difference of zero, which is their value by construction. (iii) Selection and
evaluation are separated: the switch point is chosen on one half of each
cell's rollouts and both arms are read on the held-out half, so no value is
read at the point that selected it. (iv) Uncertainty: percentile bootstrap
over prompts as the resampling unit, prompts being the level at which the
horizon is defined. (v) Secondary, reported alongside and never in place of
the primary: the survivors-only subgroup and the 48-token replication wave,
both immediately below. A subtask counts as noninferior when the lower
confidence bound exceeds $-\varepsilon$; this is a per-subtask statement and we
make no multiplicity correction across the 13, so ``13 of 13'' should be read
as a description of the table rather than as a single family-wise test.

\textbf{The noninferiority receipt across waves and subgroups.} On the audited
48-token wave the paired handoff-minus-full difference over the full census is
$+0.000\ [-0.015, +0.018]$ (keywords), $+0.007\ [-0.005, +0.023]$ (length) and
$+0.023\ [+0.009, +0.041]$ (sentiment), so the receipt of
Section~\ref{sec:results:astar} is stable across generation budgets. Restricted
to the survivors subgroup---prompts whose selected switch point is strictly
earlier than $T$---five of the twelve estimable 64-token subtasks have a lower
confidence bound below $-0.03$ and therefore miss the primary margin while
clearing the $0.05$ sensitivity margin: CommonGen kw $k{=}3$ $-0.024\ [-0.037,
-0.008]$, kw $k{=}5$ $-0.025\ [-0.034, -0.017]$, kw $k{=}10$ $-0.016\ [-0.031,
+0.005]$, CCG multi $-0.029\ [-0.035, -0.023]$ and IFEval letter-freq $-0.033\
[-0.046, -0.020]$; COLLIE c06a has no survivors, so the subgroup estimand is
undefined there. The two estimands answer different questions and we report
both. The primary one is the contrast over the population a switch policy
actually meets; the subgroup drops exactly the prompts on which the two
policies are identical by construction, so the distance between the two grows
with a row's censoring share (CommonGen kw $k{=}5$, censored $49\%$, moves from
$-0.008$ to $-0.025$; CCG multi, censored $18\%$, from $-0.024$ to $-0.029$).

\textbf{The handoff arm is not less fluent.} The census shards store scalar
outcomes and no generated text, so fluency has to be measured by re-decoding.
We do that on two waves. \emph{(i)} On the audited 48-token wave we re-decoded
every survivor twice from the same seed---once under full CFG, once freezing at
$a^{*}$---and scored GPT-2 perplexity on the paired outputs
(Table~\ref{tab:freeze-ppl}): freezing \emph{lowers} perplexity on all three
families, by $0.47$ to $0.65$ in mean per-token NLL, every interval excluding
zero. \emph{(ii)} On the 64-token census of Table~\ref{tab:noninferiority}
itself we rebuilt each handoff survivor's prefix at $a^{*}$ from that prompt's
stored commit schedule---a replay, so it costs no forward passes---and re-ran
both arms from there under the census seeds ($N{=}24$, temperature $1.0$), then
decoded (Table~\ref{tab:census-ppl}). The replay reproduces the recorded
success of \emph{both} arms exactly on $1787$ of $1787$ survivors, so the
perplexities are read on the same rollouts whose success the census reports.
The paired per-prompt difference in mean NLL per token favours freezing on $11$
of the $12$ subtasks that have a survivor, from $-0.14$ (CCG multi) to $-1.37$
(IFEval letter-freq), every interval excluding zero; COLLIE c07 is the single
row in the other direction ($+0.24\ [+0.15, +0.32]$), and we leave it as an
exception rather than read a mechanism off one row. Censored prompts are
excluded throughout: they run the same policy in both arms, so their fluency
difference is zero by construction. This is the fluency axis of
Section~\ref{sec:results:taxonomy} read at the horizon: sustained guidance past
$\astar$ costs fluency without buying success, so the handoff is not a
quality-for-success trade.
% Auto-generated by analysis_utilities/injection_sharpening/make_freeze_ppl_table.py -- do not edit by hand.
\begin{table}[t]
\centering
\footnotesize
\setlength{\tabcolsep}{4pt}
\begin{tabular}{l r r r l}
\toprule
Family & $n$ & Full CFG & Freeze $a^{*}$ & $\Delta \log$PPL [95\% CI] \\
\midrule
keywords & 148 & 60 & 38 & $-0.47$\,$[-0.53,-0.42]$ \\
length & 155 & 225 & 111 & $-0.65$\,$[-0.80,-0.51]$ \\
sentiment & 116 & 89 & 48 & $-0.53$\,$[-0.66,-0.40]$ \\
\bottomrule
\end{tabular}
\caption{Fluency of the handoff arm (GPT-2 perplexity; lower is more fluent by this proxy). Each survivor is re-decoded greedily under full CFG and under freeze-at-$a^{*}$ from the same seed, so the two arms are paired per prompt. Perplexity columns pool tokens across prompts; the last column is the per-prompt difference in mean NLL per token, bootstrapped over prompts. Freezing lowers perplexity on all three families, so the noninferiority of Table~\ref{tab:noninferiority} is not bought with fluency. \emph{Scope:} this is the audited 48-token three-family wave under greedy decoding, not the 64-token 13-subtask census of Table~\ref{tab:noninferiority}, whose shards store scalar success and no text; the two are separate estimands and we do not merge them.}
\label{tab:freeze-ppl}
\end{table}

% Auto-generated by analysis_utilities/injection_sharpening/census_noninferiority.py -- do not edit by hand.
\begin{table*}[t]
\centering
\footnotesize
\setlength{\tabcolsep}{3pt}
\begin{tabular}{l r r r l}
\toprule
Subtask & $n$ & Full CFG & Freeze & $\Delta \log$PPL [95\% CI] \\
\midrule
CommonGen kw $k{=}3$ & 151 & 46 & 34 & $-0.19$\,$[-0.27,-0.12]$ \\
CommonGen kw $k{=}4$ & 140 & 50 & 33 & $-0.36$\,$[-0.47,-0.26]$ \\
CommonGen kw $k{=}5$ & 107 & 49 & 32 & $-0.37$\,$[-0.46,-0.29]$ \\
CommonGen kw $k{=}10$ & 125 & 53 & 33 & $-0.43$\,$[-0.49,-0.36]$ \\
CCG length & 147 & 228 & 122 & $-0.66$\,$[-0.80,-0.51]$ \\
CCG sentiment & 125 & 55 & 35 & $-0.25$\,$[-0.39,-0.11]$ \\
CCG multi & 170 & 55 & 40 & $-0.14$\,$[-0.22,-0.06]$ \\
IFEval kw-freq & 191 & 37 & 20 & $-0.59$\,$[-0.63,-0.55]$ \\
IFEval letter-freq & 140 & 89 & 23 & $-1.37$\,$[-1.47,-1.26]$ \\
IFEval capital-freq & 140 & 48 & 24 & $-0.68$\,$[-0.73,-0.63]$ \\
IFEval no-comma & 164 & 78 & 33 & $-0.86$\,$[-0.93,-0.79]$ \\
COLLIE c07 & 187 & 50 & 64 & $+0.24$\,$[+0.15,+0.33]$ \\
\bottomrule
\end{tabular}
\caption{Fluency of the two arms of Table~\ref{tab:noninferiority}, on that table's own census. For every handoff survivor the arms at $\astar$ were replayed from the prompt's stored commit schedule under the census seeds and decoded, so the perplexity is measured on the rollouts whose success the census recorded; the replay reproduces both recorded success rates exactly on $1787$ of $1787$ prompts. Perplexity columns pool tokens across prompts (exponential of the token-weighted mean NLL); the last column is the per-prompt difference in mean NLL per token, bootstrapped over prompts, so it is the contrast that respects the pairing. Negative favours freezing. Censored prompts are excluded: the two arms are the same policy there. COLLIE c06a has no survivor and no row.}
\label{tab:census-ppl}
\end{table*}

The smallness of the freeze-level optimism ($\le 2$ points) is itself informative:
it rules out estimator selection noise as the main source of the $\approx 11$-point
fresh-rollout deficit of Appendix~\ref{app:transfer-gap}, whose measured
decomposition (reference optimism from survivor classification, shared by the
$K{=}1$ control at every gate) is given there.

\subsection{The Horizon under the OFAT Grid}
\label{app:astar-ofat}

% Moved out of Sec 5.2 on 2026-07-28 when that block was condensed for the
% 7-page main text.  Sec 5.2 keeps only the one-sentence budget caveat and
% points here; the numbers and the estimator audit live below.
The 13-subtask census of Section~\ref{sec:results:taxonomy} also carries
survivors-only $\astar$ at every configuration, which tests whether the three
legs of Section~\ref{sec:results:horizon} are artifacts of one operating point
(Figure~\ref{fig:astar-ofat}). \emph{(i) $\astar$ does not track the guidance
weight.} For $w \ge 1.5$ the survivors-only median moves by at most one grid
point in 11 of 12 subtasks with survivors (the exception, IFEval
capital-frequency, has ${\approx}30$ survivors per cell); only starved guidance
($w \le 1$) completes transport later (length: median $0.60$ at $w{=}0.5$
versus $0.15$ at $w \ge 2$, with correspondingly heavier censoring). The
horizon is a property of the prompt--constraint pair, not of the control knob.
\emph{(ii) $\astar$ scales with the realization budget where the carrier is
lexical.} Quadrupling generation length (32 to 128) moves the keyword-family
medians later ($k{=}3/4$: $0.15 \to 0.30$; $k{=}10$: $0.22 \to 0.45$;
multi-constraint: $0.15 \to 0.45$) while length control stays at $0.15$ and
every IFEval family moves by at most one grid point; the steps axis moves no
subtask by more than one grid point in either direction.

\textbf{Estimator scope, and why the family ordering is not quoted as a
constant.} All freeze branches of a prompt share a single recorded guided
trajectory, so the per-prompt $\astar$ is a \emph{trajectory-level} estimate;
its transfer to fresh rollouts mixes estimation noise with genuine
trajectory-to-trajectory variability of the horizon, which
Appendix~\ref{app:transfer-gap} audits directly. The quoted medians are
plug-in locations: under half-rollout selection they move by at most one grid
point, and the cross-fitting of Section~\ref{sec:results:horizon} corrects the
\emph{values} read at selected points, not the selected locations themselves.
The audited three-family census runs a single generation budget (48 tokens),
and its survivors-only medians are $\astar \approx 0.45$ for sentiment
($n{=}32$; 16 further prompts remain guidance-dependent past the grid end and
are right-censored), $\approx 0.30$ for keywords ($n{=}120$) and
$\approx 0.22$ for length ($n{=}98$). The naive medians ($0.05/0.10/0.22$)
reverse this order, and the reason is algebraic rather than empirical: since
$\Vval = \qguid - \qbase \le 1 - \qbase$, any prompt born with
$\qbase \ge 1-\eta$ satisfies a first-crossing rule at the first grid point by
construction, so born-in prompts---$48\%$ of sentiment's saturated set versus
$17\%$ of keywords'---are forced to $\astar \approx 0$ and the naive
per-family median becomes a headcount of prompts that never needed guidance
rather than a timing. This is why Section~\ref{sec:results:horizon} classifies
fates before reading $\astar$, and reports it over handoff survivors only. At the 64-token budget the ordering reverses
again: the keyword family commits latest ($0.45$) and sentiment earlier
($0.22$). We therefore do not read the between-family ordering as a stable
constant of the constraint type. The plug-in oracle-versus-global gaps are
$12.6/12.7/18.2$ points against the cross-fitted $9.5/11.9/13.4$; selecting
each prompt's best-looking grid point inflates the oracle by $0.9$--$6.2$
points, which is the post-selection optimism the cross-fitting removes.

\begin{figure}[t]
\centering
\includegraphics[width=\linewidth]{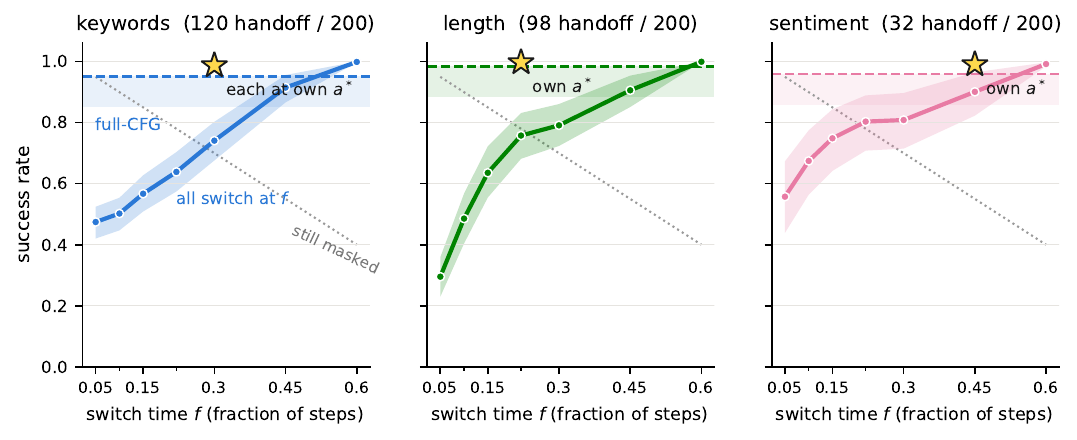}
\caption{Freeze-time sweep on handoff survivors, per constraint family
(audited 48-token census). Solid: mean success when every survivor switches at
the same normalized time $f$, with a bootstrap CI over prompts; star: mean
success when each prompt switches at its own $\astar$, plotted at the
survivors-median horizon; dashed line and shaded band: the full-CFG reference
and the handoff tolerance beneath it; dotted: the remaining-mask fraction
$1-f$, on the same fraction axis.}
\label{fig:freeze-plateau}
\end{figure}

% Moved out of Sec 5.2 on 2026-07-28 (4th pass, float budget): the body keeps
% ONE figure and ONE table in that block -- fig:astar-dist and Table 1. The
% only quantities the body still reads off this figure are the three
% remaining-mask fractions (70/78/55%), which are deterministic (1 - f at the
% survivors-median a* of 0.30/0.22/0.45) and are quoted there as numbers.
% It sits in app:astar-ofat because this is the same audited 48-token
% three-family wave the rest of the subsection reports.
\textbf{The freeze curves behind the plateau.}
Figure~\ref{fig:freeze-plateau} draws the switch sweep the noninferiority
receipt of Section~\ref{sec:results:astar} summarizes: terminal success against
the freeze time, on handoff survivors, one panel per family. The solid curve is
the best-single-global-cut policy family---every survivor switching at the same
normalized time---while the star marks the policy the body reports, each prompt
switching at its own $\astar$; the vertical distance between them at the
survivors-median horizon is the oracle-versus-global gap in visual form. Note
that the common-$f$ curve enters the tolerance band only at $f \approx
0.45$--$0.6$, later than any family's median $\astar$: a pooled switch curve
smears a distribution of per-prompt horizons and is not a location estimate,
which is the same discipline Section~\ref{sec:results:realization} enforces on
the gate sweep.

\begin{figure*}[t]
\centering
\includegraphics[width=\linewidth]{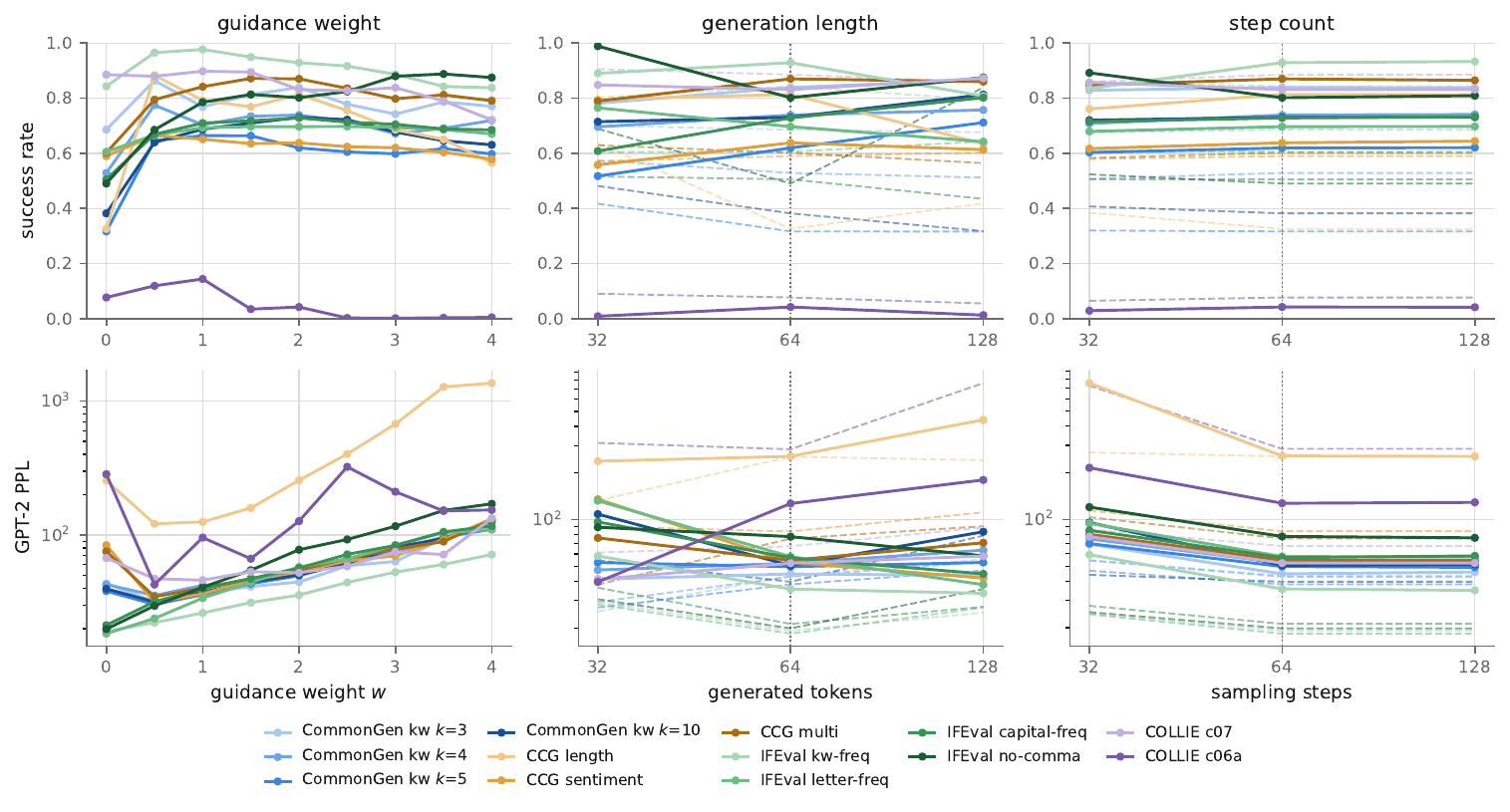}
\caption{Success (top) and fluency (bottom) along the three axes of the OFAT
census, one factor at a time off the center configuration ($w{=}2$, $64$
generated tokens, $64$ steps): guidance weight, generation length, step count.
Solid lines are the guided arm ($w{=}2$ on the two budget axes); dashed lines
on those axes are the unguided arm ($w{=}0$), i.e.\ the reference the guided
line is bought against, and the dotted vertical rule marks the center. Colour
is by benchmark, lightness by subtask within it. The two rows come from two
waves, because the census shards store scalar outcomes and no text: success is
the census/OFAT grid ($n{=}200$ prompts $\times$ $24$ rollouts per cell, the
numbers of Table~\ref{tab:sr-vs-w}), perplexity is the fluency wave that stores
generations ($n{=}100 \times 4$ per cell, the numbers of
Table~\ref{tab:ppl-vs-w}); the two agree on success to $\le 0.065$ per cell on
12 of 13 subtasks (Table~\ref{tab:ppl-vs-w} caption). Perplexity is on a log
scale and is a fluency proxy on a separate axis from constraint success---the
two are never read as one quantity.}
\label{fig:ofat-axes}
\end{figure*}

\textbf{The same axes, read on success and on fluency.}
Figure~\ref{fig:ofat-axes} completes the OFAT picture by reading the three axes
on the two outcome quantities rather than on $\astar$. Two things it makes
visible and the $\astar$ panels cannot. First, the guidance weight buys success
only over its first half-unit or so---success is flat to slowly declining past
$w \approx 1$ in every subtask---while perplexity keeps climbing over the whole
sweep from its minimum at $w{=}0.5$ (Table~\ref{tab:ppl-vs-w}), so the two
quantities stop moving together well before the end of the axis. Second, on the
two budget axes the guided and unguided arms stay separated at every point,
i.e.\ none of the configurations we ran is one where guidance has stopped
paying on success---which is what makes the horizon a within-trajectory
question rather than a choice of operating point. The two rows come from
different waves and are never read as one quantity; see the caption for the
per-cell agreement between them on success.

% Generated by analysis_utilities/injection_sharpening/fig_ofat_axes.py
% (figs/fig_ofat_axes.pdf).  Appendix companion to fig_astar_ofat_{w,gen,steps}:
% the same three OFAT axes, read on success and on fluency instead of on a*.

\subsection{OFAT-Grid Stability}

The OFAT census (Section~\ref{sec:results:taxonomy} breadth) quotes only fate
shares and survivors-only $a^{*}$ medians---no value is read at a selected
point---so its exposure to selection noise is a \emph{stability} question, not an
optimism decomposition. Applying the same hypergeometric half-split machinery
($N{=}24$, $R{=}200$) to every cell of the grid (13 subtasks $\times$ the
$w$/generation-length/steps axes; 168 cells with a defined survivor median): fate
shares move by at most $\pm 5.5$ points of $n$ (median interval half-width $2.5$
points), and the survivors-only $a^{*}$ median stays within one grid point of the
plug-in value in $164/168$ cells ($97.6\%$). The four exceptions are all
small-survivor cells ($n_{\text{handoff}} = 4$--$35$; three are the born-in COLLIE
family, whose handoff class has $n \le 19$), where a median is fragile for the
usual reason. The banding ordering quoted in
Section~\ref{sec:results:taxonomy} is unaffected.

\subsection{Per-Configuration Fate Tables}

Tables~\ref{tab:ofat-fates-w-a}--\ref{tab:ofat-fates-steps} report, for every
cell of the OFAT census (13 subtasks $\times$ the $w$/generation-length/steps
axes, $n{=}200$ prompts per cell), the six-way fate counts and the
survivors-only $a^{*}$ median and IQR. Two fates carry no preterminal horizon
and are therefore right-censored in the accounting of
Table~\ref{tab:noninferiority}: persistent-dependent prompts, still
guidance-dependent at the last grid point, and failures, which reach the
success threshold under neither policy. A cell's censoring share is their sum.
CommonGen $k{=}3$ at the center configuration contributes $4 + 51$ of $200$,
the $27\%$ of Table~\ref{tab:noninferiority}.

% Auto-generated by analysis_utilities/injection_sharpening/make_ofat_fate_tables.py
% from results/astar_census/ASTAR_OFAT_HARVEST.json -- do not edit by hand.
% App C.4: full per-config fate tables with censoring counts.

\begin{table*}[p]
\centering
\scriptsize
\begin{tabular}{l r rrrrrr c c}
\toprule
 & & \multicolumn{6}{c}{Fate counts ($n{=}200$)} & \multicolumn{2}{c}{$a^{*}$ (survivors)} \\
\cmidrule(lr){3-8} \cmidrule(lr){9-10}
Subtask & $w$ & pre. & hand. & red. & pers. & harm. & fail. & median & IQR \\
\midrule
CommonGen kw $k{=}3$ & 0.5 & 69 & 70 & 5 & 17 & 1 & 38 & 0.22 & [0.15, 0.28] \\
 & 1 & 63 & 62 & 1 & 12 & 6 & 56 & 0.45 & [0.17, 0.45] \\
 & 1.5 & 59 & 74 & 0 & 13 & 6 & 48 & 0.30 & [0.17, 0.45] \\
 & 2 & 62 & 78 & 1 & 4 & 4 & 51 & 0.30 & [0.15, 0.45] \\
 & 2.5 & 68 & 62 & 1 & 9 & 3 & 57 & 0.30 & [0.15, 0.60] \\
 & 3 & 76 & 54 & 1 & 7 & 2 & 60 & 0.26 & [0.15, 0.45] \\
 & 3.5 & 72 & 62 & 1 & 3 & 0 & 62 & 0.30 & [0.15, 0.45] \\
 & 4 & 69 & 60 & 1 & 4 & 2 & 64 & 0.22 & [0.15, 0.45] \\
\addlinespace
CommonGen kw $k{=}4$ & 0.5 & 40 & 88 & 2 & 12 & 6 & 52 & 0.22 & [0.20, 0.45] \\
 & 1 & 37 & 84 & 2 & 10 & 4 & 63 & 0.22 & [0.15, 0.45] \\
 & 1.5 & 37 & 87 & 1 & 7 & 7 & 61 & 0.30 & [0.15, 0.45] \\
 & 2 & 48 & 80 & 1 & 6 & 5 & 60 & 0.30 & [0.22, 0.45] \\
 & 2.5 & 49 & 72 & 0 & 8 & 1 & 70 & 0.30 & [0.22, 0.45] \\
 & 3 & 52 & 66 & 0 & 3 & 6 & 73 & 0.22 & [0.15, 0.45] \\
 & 3.5 & 50 & 59 & 0 & 2 & 5 & 84 & 0.22 & [0.15, 0.45] \\
 & 4 & 54 & 66 & 0 & 3 & 3 & 74 & 0.22 & [0.15, 0.45] \\
\addlinespace
CommonGen kw $k{=}5$ & 0.5 & 13 & 87 & 2 & 13 & 0 & 85 & 0.30 & [0.22, 0.45] \\
 & 1 & 13 & 102 & 1 & 9 & 1 & 74 & 0.22 & [0.15, 0.45] \\
 & 1.5 & 15 & 100 & 1 & 7 & 3 & 74 & 0.22 & [0.22, 0.45] \\
 & 2 & 12 & 87 & 1 & 6 & 2 & 92 & 0.22 & [0.15, 0.45] \\
 & 2.5 & 13 & 84 & 1 & 8 & 5 & 89 & 0.30 & [0.22, 0.45] \\
 & 3 & 19 & 70 & 1 & 7 & 5 & 98 & 0.30 & [0.15, 0.45] \\
 & 3.5 & 19 & 70 & 1 & 3 & 5 & 102 & 0.22 & [0.15, 0.45] \\
 & 4 & 19 & 73 & 1 & 3 & 3 & 101 & 0.22 & [0.15, 0.45] \\
\addlinespace
CommonGen kw $k{=}10$ & 0.5 & 2 & 87 & 1 & 20 & 1 & 89 & 0.45 & [0.30, 0.60] \\
 & 1 & 5 & 104 & 1 & 19 & 2 & 69 & 0.45 & [0.30, 0.45] \\
 & 1.5 & 5 & 105 & 1 & 7 & 4 & 78 & 0.30 & [0.22, 0.60] \\
 & 2 & 4 & 108 & 1 & 9 & 2 & 76 & 0.45 & [0.30, 0.45] \\
 & 2.5 & 3 & 122 & 0 & 10 & 7 & 58 & 0.45 & [0.22, 0.45] \\
 & 3 & 6 & 112 & 1 & 3 & 9 & 69 & 0.30 & [0.22, 0.45] \\
 & 3.5 & 7 & 96 & 2 & 10 & 9 & 76 & 0.30 & [0.22, 0.45] \\
 & 4 & 7 & 93 & 0 & 9 & 11 & 80 & 0.30 & [0.22, 0.45] \\
\midrule
CCG length & 0.5 & 28 & 49 & 2 & 36 & 1 & 84 & 0.60 & [0.45, 0.60] \\
 & 1 & 30 & 57 & 2 & 29 & 7 & 75 & 0.45 & [0.30, 0.60] \\
 & 1.5 & 47 & 60 & 0 & 15 & 4 & 74 & 0.22 & [0.14, 0.60] \\
 & 2 & 53 & 84 & 2 & 9 & 6 & 46 & 0.15 & [0.14, 0.30] \\
 & 2.5 & 43 & 76 & 4 & 6 & 11 & 60 & 0.15 & [0.10, 0.22] \\
 & 3 & 34 & 83 & 3 & 2 & 21 & 57 & 0.15 & [0.10, 0.22] \\
 & 3.5 & 23 & 76 & 2 & 8 & 25 & 66 & 0.15 & [0.10, 0.24] \\
 & 4 & 21 & 72 & 8 & 2 & 23 & 74 & 0.15 & [0.15, 0.30] \\
\addlinespace
CCG sentiment & 0.5 & 36 & 41 & 8 & 55 & 10 & 50 & 0.45 & [0.30, 0.60] \\
 & 1 & 39 & 42 & 2 & 48 & 10 & 59 & 0.45 & [0.15, 0.60] \\
 & 1.5 & 32 & 50 & 2 & 32 & 14 & 70 & 0.30 & [0.15, 0.56] \\
 & 2 & 32 & 49 & 2 & 32 & 22 & 63 & 0.22 & [0.10, 0.45] \\
 & 2.5 & 38 & 40 & 3 & 23 & 25 & 71 & 0.26 & [0.15, 0.45] \\
 & 3 & 28 & 46 & 4 & 21 & 27 & 74 & 0.30 & [0.15, 0.45] \\
 & 3.5 & 28 & 45 & 5 & 31 & 22 & 69 & 0.22 & [0.10, 0.45] \\
 & 4 & 33 & 40 & 2 & 27 & 26 & 72 & 0.22 & [0.10, 0.45] \\
\addlinespace
CCG multi & 0.5 & 21 & 101 & 2 & 16 & 2 & 58 & 0.30 & [0.15, 0.45] \\
 & 1 & 35 & 119 & 3 & 5 & 2 & 36 & 0.30 & [0.15, 0.45] \\
 & 1.5 & 27 & 132 & 1 & 2 & 3 & 35 & 0.30 & [0.22, 0.45] \\
 & 2 & 30 & 129 & 1 & 3 & 3 & 34 & 0.30 & [0.22, 0.45] \\
 & 2.5 & 36 & 111 & 0 & 6 & 7 & 40 & 0.30 & [0.22, 0.45] \\
 & 3 & 46 & 96 & 2 & 8 & 8 & 40 & 0.22 & [0.15, 0.45] \\
 & 3.5 & 42 & 103 & 1 & 4 & 5 & 45 & 0.30 & [0.15, 0.45] \\
 & 4 & 40 & 89 & 0 & 3 & 9 & 59 & 0.22 & [0.15, 0.45] \\
\bottomrule
\end{tabular}
\caption{Six-way fate counts (preformed / handoff / redundant-drop / persistent-dependent / harmful / failure; Section~5.1) and survivors-only $a^{*}$ (median and IQR over handoff prompts, $N{=}24$ rollouts per prompt) for every cell of the OFAT census. Persistent-dependent prompts are exactly the right-censored ones (still guidance-dependent at the last grid point), so their count is the censoring count. The center configuration ($w{=}2$, 64 generated tokens, 64 steps) appears in all three axes. Guidance-weight axis (gen 64, steps 64), CommonGen and CCG subtasks.}
\label{tab:ofat-fates-w-a}
\end{table*}

\begin{table*}[p]
\centering
\scriptsize
\begin{tabular}{l r rrrrrr c c}
\toprule
 & & \multicolumn{6}{c}{Fate counts ($n{=}200$)} & \multicolumn{2}{c}{$a^{*}$ (survivors)} \\
\cmidrule(lr){3-8} \cmidrule(lr){9-10}
Subtask & $w$ & pre. & hand. & red. & pers. & harm. & fail. & median & IQR \\
\midrule
IFEval kw-freq & 0.5 & 138 & 46 & 0 & 3 & 0 & 13 & 0.45 & [0.22, 0.45] \\
 & 1 & 135 & 51 & 0 & 4 & 5 & 5 & 0.22 & [0.15, 0.45] \\
 & 1.5 & 121 & 61 & 0 & 2 & 13 & 3 & 0.22 & [0.15, 0.30] \\
 & 2 & 110 & 58 & 1 & 0 & 22 & 9 & 0.22 & [0.15, 0.45] \\
 & 2.5 & 100 & 60 & 0 & 1 & 31 & 8 & 0.22 & [0.15, 0.45] \\
 & 3 & 92 & 58 & 0 & 2 & 31 & 17 & 0.22 & [0.15, 0.30] \\
 & 3.5 & 78 & 64 & 0 & 0 & 40 & 18 & 0.22 & [0.15, 0.30] \\
 & 4 & 78 & 56 & 1 & 2 & 44 & 19 & 0.22 & [0.10, 0.24] \\
\addlinespace
IFEval letter-freq & 0.5 & 87 & 30 & 2 & 11 & 0 & 70 & 0.45 & [0.22, 0.60] \\
 & 1 & 87 & 48 & 0 & 4 & 1 & 60 & 0.30 & [0.22, 0.45] \\
 & 1.5 & 88 & 44 & 0 & 4 & 1 & 63 & 0.22 & [0.15, 0.30] \\
 & 2 & 90 & 42 & 0 & 10 & 1 & 57 & 0.22 & [0.15, 0.30] \\
 & 2.5 & 91 & 45 & 0 & 4 & 0 & 60 & 0.22 & [0.15, 0.30] \\
 & 3 & 89 & 45 & 0 & 4 & 2 & 60 & 0.22 & [0.15, 0.22] \\
 & 3.5 & 85 & 44 & 1 & 5 & 6 & 59 & 0.22 & [0.15, 0.24] \\
 & 4 & 82 & 41 & 0 & 5 & 8 & 64 & 0.22 & [0.15, 0.22] \\
\addlinespace
IFEval capital-freq & 0.5 & 89 & 27 & 2 & 7 & 0 & 75 & 0.15 & [0.12, 0.38] \\
 & 1 & 88 & 35 & 1 & 17 & 0 & 59 & 0.45 & [0.15, 0.53] \\
 & 1.5 & 89 & 28 & 2 & 14 & 2 & 65 & 0.22 & [0.10, 0.45] \\
 & 2 & 90 & 35 & 2 & 10 & 4 & 59 & 0.30 & [0.15, 0.60] \\
 & 2.5 & 86 & 31 & 1 & 15 & 2 & 65 & 0.30 & [0.12, 0.45] \\
 & 3 & 86 & 32 & 2 & 8 & 1 & 71 & 0.45 & [0.22, 0.60] \\
 & 3.5 & 86 & 37 & 0 & 6 & 0 & 71 & 0.45 & [0.22, 0.45] \\
 & 4 & 84 & 34 & 0 & 14 & 3 & 65 & 0.45 & [0.22, 0.45] \\
\addlinespace
IFEval no-comma & 0.5 & 3 & 108 & 2 & 30 & 0 & 57 & 0.30 & [0.22, 0.45] \\
 & 1 & 2 & 123 & 0 & 27 & 0 & 48 & 0.22 & [0.15, 0.38] \\
 & 1.5 & 2 & 148 & 0 & 11 & 0 & 39 & 0.22 & [0.15, 0.45] \\
 & 2 & 2 & 147 & 2 & 8 & 1 & 40 & 0.22 & [0.15, 0.45] \\
 & 2.5 & 5 & 146 & 3 & 15 & 1 & 30 & 0.22 & [0.15, 0.45] \\
 & 3 & 6 & 139 & 2 & 20 & 5 & 28 & 0.22 & [0.22, 0.30] \\
 & 3.5 & 6 & 156 & 2 & 17 & 1 & 18 & 0.22 & [0.15, 0.30] \\
 & 4 & 5 & 155 & 2 & 17 & 7 & 14 & 0.22 & [0.15, 0.30] \\
\midrule
COLLIE c07 & 0.5 & 172 & 14 & 3 & 10 & 0 & 1 & 0.38 & [0.30, 0.56] \\
 & 1 & 169 & 12 & 3 & 11 & 0 & 5 & 0.30 & [0.30, 0.34] \\
 & 1.5 & 164 & 19 & 1 & 7 & 0 & 9 & 0.60 & [0.30, 0.60] \\
 & 2 & 163 & 8 & 2 & 15 & 5 & 7 & 0.30 & [0.30, 0.30] \\
 & 2.5 & 140 & 39 & 0 & 3 & 15 & 3 & 0.22 & [0.15, 0.45] \\
 & 3 & 178 & 11 & 1 & 0 & 0 & 10 & 0.30 & [0.30, 0.53] \\
 & 3.5 & 181 & 5 & 1 & 5 & 0 & 8 & 0.45 & [0.10, 0.45] \\
 & 4 & 165 & 4 & 0 & 8 & 16 & 7 & 0.60 & [0.56, 0.60] \\
\addlinespace
COLLIE c06a & 0.5 & 0 & 0 & 0 & 0 & 0 & 200 & -- & -- \\
 & 1 & 0 & 0 & 0 & 25 & 0 & 175 & -- & -- \\
 & 1.5 & 0 & 0 & 0 & 25 & 0 & 175 & -- & -- \\
 & 2 & 0 & 0 & 0 & 20 & 0 & 180 & -- & -- \\
 & 2.5 & 0 & 0 & 0 & 0 & 0 & 200 & -- & -- \\
 & 3 & 0 & 0 & 0 & 0 & 0 & 200 & -- & -- \\
 & 3.5 & 0 & 0 & 0 & 0 & 0 & 200 & -- & -- \\
 & 4 & 0 & 0 & 0 & 0 & 0 & 200 & -- & -- \\
\bottomrule
\end{tabular}
\caption{Six-way fate counts (preformed / handoff / redundant-drop / persistent-dependent / harmful / failure; Section~5.1) and survivors-only $a^{*}$ (median and IQR over handoff prompts, $N{=}24$ rollouts per prompt) for every cell of the OFAT census. Persistent-dependent prompts are exactly the right-censored ones (still guidance-dependent at the last grid point), so their count is the censoring count. The center configuration ($w{=}2$, 64 generated tokens, 64 steps) appears in all three axes. Guidance-weight axis (gen 64, steps 64), IFEval and COLLIE subtasks.}
\label{tab:ofat-fates-w-b}
\end{table*}

\begin{table*}[p]
\centering
\scriptsize
\begin{tabular}{l r rrrrrr c c}
\toprule
 & & \multicolumn{6}{c}{Fate counts ($n{=}200$)} & \multicolumn{2}{c}{$a^{*}$ (survivors)} \\
\cmidrule(lr){3-8} \cmidrule(lr){9-10}
Subtask & gen & pre. & hand. & red. & pers. & harm. & fail. & median & IQR \\
\midrule
CommonGen kw $k{=}3$ & 32 & 69 & 78 & 3 & 0 & 4 & 46 & 0.15 & [0.10, 0.22] \\
 & 64 & 62 & 78 & 1 & 4 & 4 & 51 & 0.30 & [0.15, 0.45] \\
 & 128 & 65 & 54 & 0 & 30 & 2 & 49 & 0.30 & [0.15, 0.45] \\
\addlinespace
CommonGen kw $k{=}4$ & 32 & 44 & 93 & 2 & 0 & 4 & 57 & 0.15 & [0.10, 0.30] \\
 & 64 & 48 & 80 & 1 & 6 & 5 & 60 & 0.30 & [0.22, 0.45] \\
 & 128 & 43 & 52 & 2 & 27 & 5 & 71 & 0.30 & [0.15, 0.45] \\
\addlinespace
CommonGen kw $k{=}5$ & 32 & 23 & 84 & 1 & 0 & 4 & 88 & 0.22 & [0.15, 0.30] \\
 & 64 & 12 & 87 & 1 & 6 & 2 & 92 & 0.22 & [0.15, 0.45] \\
 & 128 & 18 & 59 & 3 & 21 & 4 & 95 & 0.30 & [0.15, 0.60] \\
\addlinespace
CommonGen kw $k{=}10$ & 32 & 17 & 124 & 0 & 0 & 3 & 56 & 0.22 & [0.15, 0.30] \\
 & 64 & 4 & 108 & 1 & 9 & 2 & 76 & 0.45 & [0.30, 0.45] \\
 & 128 & 3 & 98 & 0 & 27 & 1 & 71 & 0.45 & [0.22, 0.45] \\
\midrule
CCG length & 32 & 86 & 74 & 0 & 0 & 5 & 35 & 0.15 & [0.11, 0.30] \\
 & 64 & 53 & 84 & 2 & 9 & 6 & 46 & 0.15 & [0.14, 0.30] \\
 & 128 & 54 & 38 & 1 & 10 & 36 & 61 & 0.15 & [0.11, 0.30] \\
\addlinespace
CCG sentiment & 32 & 41 & 57 & 3 & 0 & 11 & 88 & 0.45 & [0.22, 0.60] \\
 & 64 & 32 & 49 & 2 & 32 & 22 & 63 & 0.22 & [0.10, 0.45] \\
 & 128 & 22 & 47 & 3 & 40 & 13 & 75 & 0.45 & [0.22, 0.60] \\
\addlinespace
CCG multi & 32 & 38 & 114 & 0 & 0 & 6 & 42 & 0.15 & [0.15, 0.22] \\
 & 64 & 30 & 129 & 1 & 3 & 3 & 34 & 0.30 & [0.22, 0.45] \\
 & 128 & 19 & 73 & 4 & 42 & 3 & 59 & 0.45 & [0.22, 0.60] \\
\midrule
IFEval kw-freq & 32 & 111 & 51 & 0 & 0 & 15 & 23 & 0.22 & [0.15, 0.30] \\
 & 64 & 110 & 58 & 1 & 0 & 22 & 9 & 0.22 & [0.15, 0.45] \\
 & 128 & 67 & 51 & 1 & 2 & 58 & 21 & 0.15 & [0.10, 0.18] \\
\addlinespace
IFEval letter-freq & 32 & 79 & 65 & 1 & 0 & 7 & 48 & 0.22 & [0.15, 0.30] \\
 & 64 & 90 & 42 & 0 & 10 & 1 & 57 & 0.22 & [0.15, 0.30] \\
 & 128 & 74 & 43 & 0 & 4 & 9 & 70 & 0.22 & [0.10, 0.30] \\
\addlinespace
IFEval capital-freq & 32 & 84 & 35 & 1 & 0 & 3 & 77 & 0.30 & [0.15, 0.45] \\
 & 64 & 90 & 35 & 2 & 10 & 4 & 59 & 0.30 & [0.15, 0.60] \\
 & 128 & 88 & 49 & 0 & 10 & 3 & 50 & 0.30 & [0.15, 0.45] \\
\addlinespace
IFEval no-comma & 32 & 69 & 124 & 2 & 0 & 0 & 5 & 0.15 & [0.15, 0.22] \\
 & 64 & 2 & 147 & 2 & 8 & 1 & 40 & 0.22 & [0.15, 0.45] \\
 & 128 & 40 & 119 & 3 & 4 & 4 & 30 & 0.15 & [0.10, 0.22] \\
\midrule
COLLIE c07 & 32 & 159 & 27 & 1 & 0 & 0 & 13 & 0.10 & [0.10, 0.22] \\
 & 64 & 163 & 8 & 2 & 15 & 5 & 7 & 0.30 & [0.30, 0.30] \\
 & 128 & 166 & 12 & 0 & 13 & 4 & 5 & 0.15 & [0.15, 0.15] \\
\addlinespace
COLLIE c06a & 32 & 0 & 0 & 0 & 0 & 0 & 200 & -- & -- \\
 & 64 & 0 & 0 & 0 & 20 & 0 & 180 & -- & -- \\
 & 128 & 0 & 0 & 0 & 20 & 0 & 180 & -- & -- \\
\bottomrule
\end{tabular}
\caption{Six-way fate counts (preformed / handoff / redundant-drop / persistent-dependent / harmful / failure; Section~5.1) and survivors-only $a^{*}$ (median and IQR over handoff prompts, $N{=}24$ rollouts per prompt) for every cell of the OFAT census. Persistent-dependent prompts are exactly the right-censored ones (still guidance-dependent at the last grid point), so their count is the censoring count. The center configuration ($w{=}2$, 64 generated tokens, 64 steps) appears in all three axes. Generation-length axis ($w{=}2$, steps 64), all 13 subtasks.}
\label{tab:ofat-fates-gen}
\end{table*}

\begin{table*}[p]
\centering
\scriptsize
\begin{tabular}{l r rrrrrr c c}
\toprule
 & & \multicolumn{6}{c}{Fate counts ($n{=}200$)} & \multicolumn{2}{c}{$a^{*}$ (survivors)} \\
\cmidrule(lr){3-8} \cmidrule(lr){9-10}
Subtask & steps & pre. & hand. & red. & pers. & harm. & fail. & median & IQR \\
\midrule
CommonGen kw $k{=}3$ & 32 & 63 & 75 & 0 & 8 & 1 & 53 & 0.30 & [0.22, 0.60] \\
 & 64 & 62 & 78 & 1 & 4 & 4 & 51 & 0.30 & [0.15, 0.45] \\
 & 128 & 80 & 79 & 0 & 0 & 5 & 36 & 0.22 & [0.15, 0.30] \\
\addlinespace
CommonGen kw $k{=}4$ & 32 & 35 & 73 & 2 & 7 & 10 & 73 & 0.30 & [0.15, 0.60] \\
 & 64 & 48 & 80 & 1 & 6 & 5 & 60 & 0.30 & [0.22, 0.45] \\
 & 128 & 55 & 89 & 1 & 0 & 10 & 45 & 0.22 & [0.15, 0.30] \\
\addlinespace
CommonGen kw $k{=}5$ & 32 & 12 & 75 & 2 & 4 & 3 & 104 & 0.45 & [0.26, 0.60] \\
 & 64 & 12 & 87 & 1 & 6 & 2 & 92 & 0.22 & [0.15, 0.45] \\
 & 128 & 29 & 94 & 0 & 0 & 2 & 75 & 0.22 & [0.15, 0.41] \\
\addlinespace
CommonGen kw $k{=}10$ & 32 & 2 & 126 & 2 & 7 & 7 & 56 & 0.30 & [0.22, 0.45] \\
 & 64 & 4 & 108 & 1 & 9 & 2 & 76 & 0.45 & [0.30, 0.45] \\
 & 128 & 10 & 131 & 0 & 0 & 1 & 58 & 0.22 & [0.15, 0.30] \\
\midrule
CCG length & 32 & 51 & 63 & 1 & 8 & 17 & 60 & 0.22 & [0.12, 0.45] \\
 & 64 & 53 & 84 & 2 & 9 & 6 & 46 & 0.15 & [0.14, 0.30] \\
 & 128 & 73 & 78 & 2 & 0 & 4 & 43 & 0.15 & [0.10, 0.30] \\
\addlinespace
CCG sentiment & 32 & 25 & 48 & 8 & 30 & 19 & 70 & 0.26 & [0.15, 0.45] \\
 & 64 & 32 & 49 & 2 & 32 & 22 & 63 & 0.22 & [0.10, 0.45] \\
 & 128 & 43 & 68 & 3 & 0 & 24 & 62 & 0.30 & [0.15, 0.60] \\
\addlinespace
CCG multi & 32 & 24 & 127 & 2 & 2 & 1 & 44 & 0.30 & [0.22, 0.45] \\
 & 64 & 30 & 129 & 1 & 3 & 3 & 34 & 0.30 & [0.22, 0.45] \\
 & 128 & 49 & 119 & 2 & 0 & 4 & 26 & 0.22 & [0.15, 0.30] \\
\midrule
IFEval kw-freq & 32 & 75 & 56 & 1 & 3 & 44 & 21 & 0.15 & [0.15, 0.30] \\
 & 64 & 110 & 58 & 1 & 0 & 22 & 9 & 0.22 & [0.15, 0.45] \\
 & 128 & 111 & 57 & 0 & 0 & 20 & 12 & 0.15 & [0.10, 0.22] \\
\addlinespace
IFEval letter-freq & 32 & 81 & 44 & 0 & 3 & 7 & 65 & 0.22 & [0.15, 0.30] \\
 & 64 & 90 & 42 & 0 & 10 & 1 & 57 & 0.22 & [0.15, 0.30] \\
 & 128 & 94 & 43 & 1 & 0 & 0 & 62 & 0.15 & [0.10, 0.22] \\
\addlinespace
IFEval capital-freq & 32 & 89 & 36 & 1 & 15 & 0 & 59 & 0.30 & [0.15, 0.60] \\
 & 64 & 90 & 35 & 2 & 10 & 4 & 59 & 0.30 & [0.15, 0.60] \\
 & 128 & 95 & 46 & 1 & 0 & 8 & 50 & 0.30 & [0.15, 0.45] \\
\addlinespace
IFEval no-comma & 32 & 8 & 167 & 0 & 9 & 0 & 16 & 0.22 & [0.15, 0.30] \\
 & 64 & 2 & 147 & 2 & 8 & 1 & 40 & 0.22 & [0.15, 0.45] \\
 & 128 & 10 & 149 & 1 & 0 & 2 & 38 & 0.15 & [0.10, 0.22] \\
\midrule
COLLIE c07 & 32 & 167 & 19 & 0 & 4 & 1 & 9 & 0.30 & [0.16, 0.30] \\
 & 64 & 163 & 8 & 2 & 15 & 5 & 7 & 0.30 & [0.30, 0.30] \\
 & 128 & 164 & 28 & 0 & 0 & 0 & 8 & 0.45 & [0.15, 0.49] \\
\addlinespace
COLLIE c06a & 32 & 0 & 0 & 0 & 0 & 0 & 200 & -- & -- \\
 & 64 & 0 & 0 & 0 & 20 & 0 & 180 & -- & -- \\
 & 128 & 0 & 0 & 1 & 0 & 26 & 173 & -- & -- \\
\bottomrule
\end{tabular}
\caption{Six-way fate counts (preformed / handoff / redundant-drop / persistent-dependent / harmful / failure; Section~5.1) and survivors-only $a^{*}$ (median and IQR over handoff prompts, $N{=}24$ rollouts per prompt) for every cell of the OFAT census. Persistent-dependent prompts are exactly the right-censored ones (still guidance-dependent at the last grid point), so their count is the censoring count. The center configuration ($w{=}2$, 64 generated tokens, 64 steps) appears in all three axes. Denoising-steps axis ($w{=}2$, gen 64), all 13 subtasks.}
\label{tab:ofat-fates-steps}
\end{table*}

\section{Where the First-Order Transport Law Goes Quiet}
\label{app:transport-modes}

Section~\ref{sec:results:horizon} shows that guidance stops buying
success past $\astar$ and defers to this appendix the question of why local
transport goes quiet there. The answer below is partial, and we state the
limit explicitly: the diagnostics identify which factor of the transport law
switches off, and they do not identify the channel through which guidance's
terminal benefit arrives. No claim in the body rests on this section.

The first-order transport law of Section~\ref{sec:transport} identifies
\emph{which factor} switches off where guidance stops moving the committor.
This is an account of local transport only; it is not a decomposition of the
terminal success benefit.

\paragraph{Design.}
The dual-gauge census records, at each measured step of a
guidance-dependent cell, whether the step exhibits positive first-order
transport and, when it does not, which factor of
\[
\Delta \qguid_t
\approx
w\,\rho_t\,\sigma(\delta_t)\,\sigma(q^0_{t+1})
\]
switches off. Tilt collapse occurs when
$\sigma(\delta_t)=0$: the tilt is constant over reachable successors, so
normalization removes the token-level tilt and the conditional guided kernel
equals the conditional base kernel. Outcome collapse occurs when
$\sigma(\delta_t)>0$ but $\sigma(q^0_{t+1})=0$: every reachable successor has
the same base-continuation value, so redistributing probability among them
cannot change the expected committor. The common value need not be near $0$
or $1$. Mis- or anti-alignment occurs when both standard deviations are
positive but $\rho_t\leq0$, so the tilt is orthogonal to or opposed to higher
continuation value.

The categories are defined hierarchically because $\rho_t$ is undefined when
either standard deviation vanishes. The implementation first assigns tilt
collapse, then outcome collapse, and then mis- or anti-alignment. Thus, the
doubly degenerate case
$\sigma(\delta_t)=\sigma(q^0_{t+1})=0$ is assigned to tilt collapse. The
implementation uses exact zero tests rather than tuned thresholds. This rule
makes the reported categories exhaustive and mutually exclusive among
non-transporting measured steps, so their shares stack over normalized time.

\begin{figure}[t]
\centering
\includegraphics[width=\linewidth]{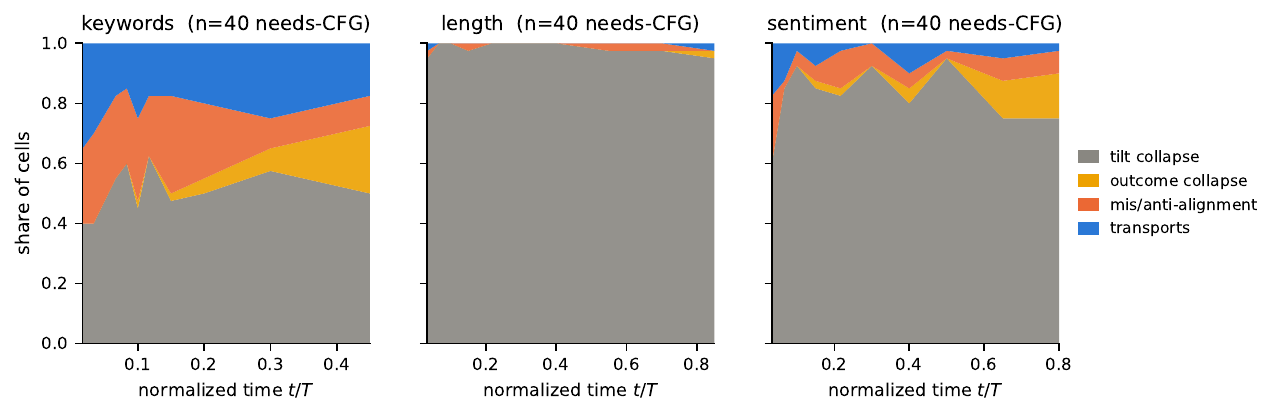}
\caption{Time-resolved failure decomposition of the transport law: stacked
per-step shares of the three failure modes---tilt collapse
($\sigma(\delta_t){=}0$), outcome collapse ($\sigma(q^{0}_{t+1}) \approx 0$),
and mis-/anti-alignment---per family over normalized time, from the
dual-gauge census. The aggregate tilt-collapse shares quoted above
($98/82/51\%$ for length/sentiment/keywords) are this figure's
time-marginal.}
\label{fig:failure-shares}
\end{figure}

\paragraph{Where the law goes quiet.} Stacking the shares shows that the
dominant mode is tilt collapse, and that it dominates precisely on the prompts
where guidance is most valuable (Figure~\ref{fig:failure-shares}): its
time-marginal reaches $98\%$ on length, $82\%$ on sentiment, and $51\%$ on
keywords. The reading is that local transport is \emph{sparse}---most steps of
a guidance-dependent trajectory carry no first-order transport at all, because
guided and base kernels have already converged at that state---not that
guidance is inactive over the trajectory as a whole. These diagnostics
identify which factor suppresses first-order transport at each stage; they do
not license the stronger statement that the three modes account for the
terminal benefit, and the closure test below is the direct test of why not.

\paragraph{Detected events do not suffice: an attribution gap.} The plateau of
Section~\ref{sec:results:astar} establishes that the pre-$\astar$ \emph{prefix}
of guidance is sufficient; a four-arm closure test asks whether the sparse
transport events detected \emph{within} that prefix are. On common seeds we
compare full guidance, no guidance, guidance applied only at detected event
steps, and guidance on the event-covering early window (three carriers,
$N = 16$ rollouts per arm; exploratory, so per-prompt rates are coarse). The
answer is no, twice over. First, on the majority of guidance-dependent prompts
the detector finds no transport event at all---$97\%$ on length, $82\%$ on
sentiment, $52\%$ on keywords---shares that track the tilt-collapse fractions
measured independently by the dual-gauge census ($98/82/51\%$), so exactly
where guidance matters most, event-anchored intervention is impossible by
construction. Second, where events are detected, event-scheduled guidance
recovers only ${\approx}18\%$ of the full-versus-base success gap (keywords),
and the event-anchored window arm is no better under paired comparison. We
report this as an \emph{attribution gap}, not as a mechanism: guidance's
terminal benefit is carried more diffusely than any set of transport events
our measurement grid detects. In particular, we do not attribute the benefit
to the committor-orthogonal remainder of $\delta_t$---higher-order,
multi-step, or support-shifting channels are all consistent with the data,
and distinguishing them would require intervention evidence we do not have.
The honest summary is therefore two-sided: the covariance law tells us
\emph{where} first-order transport is available and which factor removes it,
and it does not tell us through which channel the terminal benefit arrives.

% ============================================================================
% Moved out of Sec 5.5 on 2026-07-28 (7-page condense). NOTE (4th pass): the
% Sec 5 preamble that used to quote the lesion--relocation guard is gone, so
% the guard now lives ONLY here -- the lead-in below no longer claims the body
% states it. The section is still reached from the body twice (multi-basin
% geometry in Sec 5.3, absorbing dynamics in Sec 5.4). The label
% sec:results:macrostate is RETAINED here so existing \ref calls still resolve.
% ============================================================================
\section{Commitment Is Not Carried by a Fixed Token Position}
\label{app:macrostate}
\label{sec:results:macrostate}

This appendix rules out a trivial interpretation of $\astar$, namely
that it only records when a particular token is revealed. It also establishes
the premise used in Section~\ref{sec:results:repair}: under absorbing
unmasking, an already written position cannot be repaired in place without
reopening.

For keyword constraints, commitment is consistent with a semantic set of
valid realizations rather than with one fixed token identity or position. We
test this by destroying an already revealed keyword occurrence and asking
whether the constraint can still be realized elsewhere. This provides
evidence for a distributed macrostate while separating outcome commitment
from the reveal time of one token.

\paragraph{Design.}
We use $100$ prompts, each with one successful full-guidance trajectory. For
each trajectory, we identify one required single-token keyword and the step at
which that occurrence is fully revealed. At a matched later time,
$t/T=0.5$, we apply one of two interventions and run one deterministic
continuation per arm at temperature $0$, using the same schedule and
position-selection rule across arms.

\emph{Lesion} replaces the revealed keyword token with a committed competitor
token. Under absorbing unmasking, that position is never reconsidered, so the
original occurrence cannot recover in place. \emph{Reopen} masks the keyword
position and allows the model to sample it again. Each intervention is
followed by either full guidance or pure-conditional continuation. Matched
controls apply the same operations to committed non-keyword positions selected
by reveal time, confidence, or at random. We measure restoration at the
intervened position, keyword presence anywhere in the final output, and final
keyword count.

\paragraph{Findings.}
The lesioned position never recovers in place under either continuation policy
(span restoration $0.00$), as required by the absorbing kernel. Nevertheless,
the required keyword reappears elsewhere in $0.96$ of guided continuations and
$0.92$ of pure-conditional continuations, with mean final counts of $1.47$ and
$1.46$, respectively. Thus, destroying one revealed occurrence does not
usually destroy the constraint.

This result is not explained by an incomplete lesion. Every target keyword is
a single token, and the injected competitor remains at the intervened position
in every lesion continuation. It is also not explained by a pre-existing
duplicate. Among the $68$ prompts whose baseline output contains exactly one
keyword occurrence, final keyword presence remains $0.941$ under guidance and
$0.882$ under pure-conditional continuation. The re-formed keyword appears a
median of about $5$ words from the intervened position.

Reopening restores the keyword at the original position in $0.71$ of guided
continuations and $0.73$ of pure-conditional continuations. Overall keyword
presence reaches $0.98$ and $0.97$, respectively. Applying the same
interventions to matched non-keyword positions leaves keyword presence at
$1.00$, ruling out a generic effect of replacement or reopening on the
constraint.

The similar relocation rates under guided and pure-conditional continuation
are informative. They show that alternative keyword realization is largely a
property of the conditional decoding dynamics rather than an effect sustained
only by CFG. The constraint is therefore not carried by the identity or
location of one written token. When one occurrence is destroyed, the decoder
can realize the same keyword constraint through another token configuration.

At the same time, the original position cannot repair itself under absorbing
dynamics. Reopening restores revision capacity at that position, which is the
mechanistic premise of the intervention in
Section~\ref{sec:results:repair}.

We use this result as evidence for a multi-basin keyword geometry, meaning that
the success set contains multiple position-equivalent realizations. Length and
sentiment constraints do not admit the same localized intervention, so their
descriptions as a global counter or closure band and a stiff semantic axis are
used only as a descriptive framework in
Section~\ref{sec:results:realization} and
Appendix~\ref{app:gate-sweep}. 
% ============================================================================
% Moved out of Sec 5.6 on 2026-07-28 (7-page condense): the end-to-end gate
% sweep and the width (K) axis. The body keeps ONLY the +7.8 -> +0.3
% premise-level receipt in two sentences plus the two measurement disciplines;
% every table and figure of this leg lives here.
% ============================================================================
\section{The Gate Sweep and the Width Axis}
\label{app:gate-sweep}

This appendix backs the parallel half of
Section~\ref{sec:results:realization} in full: the fixed-prefix design and its
bit-exactness check, the $+7.8 \pm 2.5 \to +0.3 \pm 1.2$ receipt, the width
axis, the conditional and intention-to-treat policy tables, and the two
measurement disciplines that are enforced throughout the paper. The body
quotes only the receipt and the coverage share; everything else below appears
nowhere else, so this section must read self-contained.

% Moved out of Sec 5.3 on 2026-07-28 (4th pass): the policy table, the
% fixed-prefix design paragraph, tab_policy_quant and tab_policy_itt. The body
% keeps no float in that block.
\begin{table}[t]
\centering
\footnotesize
\setlength{\tabcolsep}{3pt}
\begin{tabular}{l ccc l}
\toprule
 & \multicolumn{2}{c}{guidance} & realization & adjacent contrast \\
Policy & pre-$\astar$ & post-$\astar$ & post-$\astar$ & isolates \\
\midrule
Base & -- & -- & sequential & (floor reference) \\
Full CFG & yes & yes & sequential & total guidance value \\
Handoff & yes & -- & sequential & $\varepsilon_g$ (License 1) \\
Handoff${+}$par-$K$ & yes & -- & parallel & $\sum_s \varepsilon_s$ (License 2) \\
\bottomrule
\end{tabular}
\caption{The four decoding policies these sweeps compare. Each
\emph{adjacent} contrast isolates exactly one term of the two-hop bound of
Proposition~\ref{prop:twohop}, so no contrast conflates the two licenses; the
parallel contrast uses the matched serial control described below.}
\label{tab:policies}
\end{table}

% \paragraph{The $K{=}1$ deficit is reference optimism, not a freeze failure.}
\label{app:transfer-gap}
% On fresh rollouts the handoff arm sits $\approx 11$ points below the census
% reference at $K{=}1$ (Table~\ref{tab:policy-quant}), which is easy to misread as
% a cost of freezing. That gap is carried by the $K{=}1$ control at \emph{every}
% gate position---it persists even when guidance stays on for $80\%$ of the
% trajectory---so it is gate-independent and parallelism-independent. Its source
% is selection: handoff prompts are classified partly \emph{because} their census
% rollouts succeeded, so fresh rollouts regress toward the mean (the same
% post-selection mechanism audited in Appendix~\ref{app:census}). The
% decomposition is therefore a common component of $\approx 11$ points at all
% gates plus a parallelism-specific component of $\approx 8$ points confined to
% gates at or before the horizon. The post-$\astar$ arm also pays a serial
% pre-horizon prefix ($\approx 43$ forward evaluations per prompt at $K{=}16$
% against $8$ for the all-parallel arm), so the operational reading is not that
% freezing is cheaper but that the horizon marks where parallelism stops costing
% success.

\paragraph{Fixed-prefix test: changing the kernel at a committed state.} From
each prompt's recorded horizon we branch a single realized prefix
$x_{\astar}$ into a serial ($K{=}1$) and a parallel ($K > 1$) continuation,
under guidance or after handoff. Because both arms of a pair share the prefix
token for token, the contrast isolates the continuation kernel and nothing
else; as a hard check, the $K{=}1$ guided arm must reproduce the recorded
trajectory bit-exactly, which it does on all $3388$ of $3388$ comparisons. The
design is complete: $2600$ prompt-level cells over 13 subtasks, of which
$1695$ carry a preterminal horizon and $905$ are right-censored at $T$ (COLLIE
c06a, the failure-dominated family, contributes zero survivors, exactly as its
census fate predicts). Pooled over survivors, parallelism is cheaper after
handoff than under sustained guidance at moderate $K$ (at $K{=}8$: $-0.060$
from $K{=}1$ after handoff versus $-0.078$ under guidance).
Table~\ref{tab:policy-quant} gives the fresh-deployment version of the same
contrast at scale, with per-subtask detail in
Table~\ref{tab:parallel-k-bytask}; Table~\ref{tab:policy-itt} repeats it as
the intention-to-treat comparison over the full census that
Section~\ref{sec:results:realization} quotes, with per-subtask coverage in
Table~\ref{tab:itt-bytask}. Under that fallback the switch fires on $35\%$ of
prompts, so both effects shrink---the $K{=}1$ policy difference from $0.11$
conditional to $0.04$, and the cost of raising parallelism to $K{=}16$ from
$-0.25$ to $-0.11$ under full CFG and from $-0.07$ to $-0.02$ after
handoff---while the ordering survives dilution. The leading term of the ITT
table is coverage, not effect size.
% tab_policy_quant MOVED to Sec 5.3 (2026-07-28): it is the deployment table of
% the two-table split (validity = tab_noninf_census13 in 5.2), and the body now
% reads its SR and PPL columns directly. Do not re-input here -- duplicate
% label. fig:parallel-k below is the same two-arm sweep in figure form and
% stays here; do not put both in the body.
\begin{figure}[t]
\centering
\includegraphics[width=\linewidth]{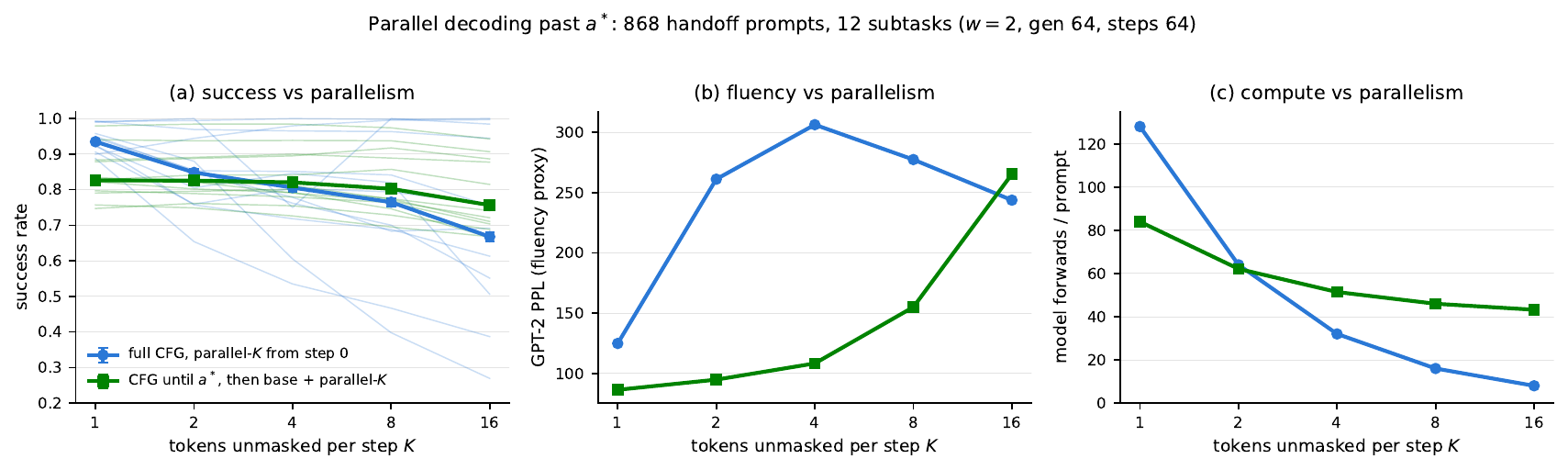}
\caption{Two-arm parallel-$K$ sweep on the 905 handoff prompts: success,
GPT-2 perplexity, and forward count versus $K$ for parallel-from-0 with CFG
throughout versus serial CFG to each prompt's recorded $a^{*}$ followed by
guidance-free parallel filling. The same increase in parallelism costs far
more success before the horizon than after it; perplexity rises with $K$ in
both arms.}
\label{fig:parallel-k}
\end{figure}
% Auto-generated by analysis_utilities/injection_sharpening/make_itt_policy_table.py
% from results/parallel_k_census/ITT_PARALLEL_K_HARVEST.json -- do not edit by hand.
\begin{table*}[t]
\centering
\small
\begin{tabular}{l l r r r r r}
\toprule
Policy & Gate source & $K$ & SR & $\Delta_K$ vs $K{=}1$ & PPL & NFE \\
\midrule
Full CFG & -- & 1 & 0.72 & -- & 145 & 128 \\
 &  & 4 & 0.70 & -0.02 & 337 & 32 \\
 &  & 8 & 0.67 & -0.05 & 338 & 16 \\
 &  & 16 & 0.62 & -0.11 & 347 & 8 \\
\midrule
Handoff at $\hat a^{*}$ & recorded $\hat a^{*}$ & 1 & 0.69 & -- & 131 & 113 \\
 &  & 4 & 0.68 & -0.00 & 139 & 101 \\
 &  & 8 & 0.68 & -0.01 & 155 & 99 \\
 &  & 16 & 0.66 & -0.02 & 193 & 98 \\
\bottomrule
\end{tabular}
\caption{Intention-to-treat deployment on the \emph{full} census: every
prompt of all 13 subtasks enters the denominator
($n{=}2600$ prompts, $N{=}24$ fresh rollouts each), with no survivor
filtering. \emph{Handoff at $\hat a^{*}$} switches guidance off at each
prompt's recorded census horizon and realizes the remainder under the base
kernel committing $K$ positions per step; on a prompt with no recorded horizon
the policy never switches and the prespecified fallback---serial full
CFG---is decoded and charged to the policy, so $K$ is inert there. The switch
therefore fires on 35\% of prompts (per-subtask coverage in
Table~\ref{tab:itt-bytask}), and the ITT contrast is diluted towards the
fallback by construction: it answers what the policy delivers on the arriving
prompt distribution, not what handoff does where it applies. $\Delta_K$ is the
change in success relative to the \emph{same policy} at $K{=}1$: raising
parallelism costs more success under full CFG than after handoff ($K{=}8$:
$-0.05$ versus $-0.01$; $K{=}16$: $-0.11$ versus
$-0.02$). The $K{=}1$ difference between the policies
($-0.04$) is the ITT deployment gap of the recorded horizon; the
\emph{validity} of the horizon is established prompt-by-prompt on the same
census by the cross-fitted noninferiority analysis of
Table~\ref{tab:noninferiority}, and the conditional behaviour among eligible
trajectories is reported in Table~\ref{tab:parallel-k-bytask}. PPL is GPT-2
perplexity (fluency proxy, separate axis); NFE is the mean number of model
forward evaluations per sequence, reported descriptively.}
\label{tab:policy-itt}
\end{table*}

\paragraph{The pooled interaction is a mixture.} The natural summary of the
second license---the handoff-by-$K$ interaction, i.e.\ how much less
parallelism costs once guidance is off---is $+0.018$ $[+0.005, +0.031]$ at
$K{=}8$ pooled over subtasks. That number is uninformative, and reporting it
alone would be a mistake: it is the average of two populations with opposite
signs whose spread is an order of magnitude larger than their mean. Handoff
\emph{protects} against parallelism on IFEval keyword-frequency ($+0.254$
$[+0.204, +0.308]$), CCG length ($+0.150$ $[+0.088, +0.214]$) and CCG
sentiment ($+0.101$ $[+0.037, +0.167]$)---on these subtasks retaining guidance
through parallel filling is actively harmful---and \emph{hurts} on the
CommonGen counting family and its relatives (CommonGen $k{=}4$: $-0.106$
$[-0.147, -0.067]$; CCG multi: $-0.080$ $[-0.110, -0.051]$; CommonGen $k{=}3$:
$-0.061$ $[-0.097, -0.024]$), where guidance is still doing work that parallel
base filling undoes. The sign tracks constraint geometry rather than benchmark
provenance: the subtasks that lose are those whose constraint is a count over
several lexical items, the \emph{multi-basin} geometry of
Appendix~\ref{app:macrostate}, where independent parallel fills can satisfy
each item locally while breaking the joint count. The second license is
therefore reported stratified by subtask throughout, and the pooled value is a
null artifact of mixture cancellation, not evidence of no effect.

The fixed-prefix test of Section~\ref{sec:results:realization} cannot locate
the boundary, because it conditions on the boundary. We therefore sweep the
switch position itself on fresh trajectories, with every arm identical in
form---serial CFG before the gate, guidance-free unmasking after it---and a
matched serial ($K{=}1$) control at every gate, so that any loss the two arms
share is by construction not a cost of parallelism
(Table~\ref{tab:parallel-gate}).
% Auto-generated by analysis_utilities/injection_sharpening/make_parallel_gate_table.py
% from results/parallel_gate/*.json -- do not edit by hand.
\begin{table*}[t]
\centering
\small
\begin{tabular}{l r c c c c c c}
\toprule
\multicolumn{8}{l}{\emph{(a) Gated parallel arms, $K{=}8$: serial CFG to the gate, then guidance-free parallel filling (SR).}}\\
Family & $n$ & $g{=}0$ & $a^{*}/2$ & $a^{*}$ & $a^{*}{+}.15$ & $a^{*}{+}.30$ & Serial full-CFG \\
\midrule
CommonGen kw $k{=}10$ & 108 & 0.18 & 0.51 & 0.75 & 0.83 & 0.86 & 0.99 \\
CCG length & 84 & 0.46 & 0.58 & 0.89 & 0.93 & 0.93 & 0.99 \\
\midrule
\multicolumn{8}{l}{\emph{(b) Parallelism-specific cost: paired $K{1}{-}K{8}$ (points), by grid-step offset from $a^{*}$.}}\\
Family & $n$ & $-2$ & $-1$ & $0$ & $+1$ & $+2$ & \\
\midrule
CommonGen kw $k{=}10$ & 32 & $+0.9 \pm 4.5$ & $+3.8 \pm 3.6$ & $+7.8 \pm 2.5$ & $+0.3 \pm 1.2$ & $+1.8 \pm 1.7$ \\
CCG length & 32 & $-18.8 \pm 7.0$ & $-21.2 \pm 5.4$ & $+1.8 \pm 1.1$ & $+1.2 \pm 0.6$ & $+0.9 \pm 0.6$ \\
\bottomrule
\end{tabular}
\caption{Parallel decoding before versus after the commitment horizon.
\emph{Block (a)}: every arm is serial CFG ($1$ token/step, $w{=}2$) up to the
gate, then guidance-free parallel filling ($K{=}8$ tokens/step) to completion,
with the gate aligned to each prompt's census $a^{*}$; handoff prompts, $24$
rollouts per cell. Success rises steeply while the gate is below the horizon
and is within a few points of its plateau at $g{=}a^{*}$; the serial
full-CFG column is the census reference for the same prompts (its gap to the
plateau is the selection regression quantified in Section~6.2, shared by all
fresh arms). \emph{Block (b)}: on a matched subset the same sweep is run at
$K{=}1$; the paired same-gate difference isolates the cost that is
\emph{specific to parallelism}, shown in horizon-aligned grid-step offsets.
On keywords it is $+7.8 \pm 2.5$ points at the gate nearest $a^{*}$ and
$+0.3 \pm 1.2$ one grid step later; on length it never exceeds $2$ points at
any gate (negative pre-horizon entries: the $K{=}1$ control, which is
guidance-free after the gate, is itself below $K{=}8$ there). Errors are
standard errors over prompts.}
\label{tab:parallel-gate}
\end{table*}

The parallelism-specific cost---the paired same-gate difference between
$K{=}8$ and $K{=}1$---is measurably nonzero at the horizon and
indistinguishable from zero shortly after it: on keywords it is
$+7.8 \pm 2.5$ points at the gate nearest each prompt's $\astar$ and
$+0.3 \pm 1.2$ one grid step later, while on length it never exceeds $2$
points at any gate. This is the premise-level measurement that the two-hop
bound calls for---$\varepsilon_s$ is nonzero at $\astar$ and becomes statistically
indistinguishable from zero shortly after it, so premise and conclusion of the bound are verified independently.

\begin{figure*}[t]
\centering
\includegraphics[width=\linewidth]{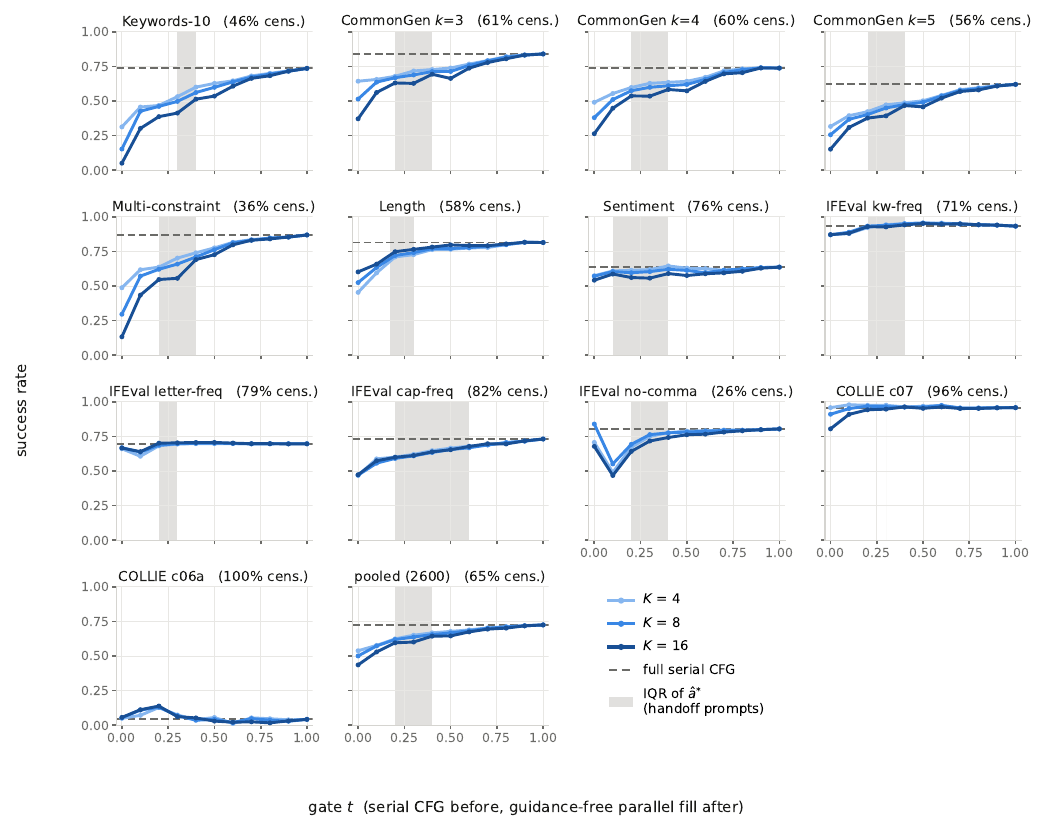}
\caption{The gate $\times$ width grid on the \emph{absolute} gate axis, one panel per
subtask. Every arm is serial CFG ($w{=}2$, 1 token/step) up to gate $t$ and guidance-free
parallel filling at $K$ tokens/step afterwards; all $30$ arms of a prompt branch off one
shared serial prefix. $n{=}200$ census prompts per subtask, $2600$ in total, nothing
excluded --- the gate schedule is prompt-independent, so $\astar$ is not in the design and
the censored prompts ($\astar{=}T$, share printed in each title) contribute complete curves
here even though no aligned analysis can use them. The $t{=}1.0$ endpoint retains no
parallel segment and is therefore $K$-independent: it is full serial CFG, drawn also as the
dashed rule, so the vertical distance from a line to that rule is the cost of handing off at
$t$ and filling $K$ wide. $K$ is an ordered width rather than an identity and is drawn as one
hue, light to dark. The shaded band is the interquartile range of $\astar$ over the
subtask's handoff prompts; it is \emph{not} a located transition, and these curves are
pooled and unaligned, so they are read as levels only --- every plateau and offset claim in
this paper rests on the per-prompt aligned increments of Table~\ref{tab:gate-k-shared}
instead. Two limits on the band: $\astar$ was resolved on the census switch grid
$\{.10,.15,.22,.30,.45,.60\}$, so no band extends right of $0.6$ and the $0.6$ bin is a
pile-up at that cap. Perplexity for the same grid is
Figure~\ref{fig:gate-k-abs-ppl}; the underlying numbers are
Tables~\ref{tab:gate-k-abs-sr} and~\ref{tab:gate-k-abs-ppl}.}
\label{fig:gate-k-abs-sr}
\end{figure*}

\paragraph{What the horizon does and does not locate.} Success along the gate
sweep is typically monotone and saturating, and the horizon marks the \emph{beginning}
of the transition into the post-guidance regime rather than the plateau
itself: on per-prompt horizon-aligned increments the largest single step
occurs immediately at the horizon ($+0.178 \pm 0.012$ from one grid step
before it to the horizon), after which the residual gain decays to at most
$0.02$ per step within the following one to two gate intervals. The plateau
therefore begins one to two grid steps \emph{after} $a^{*}$, not at it,
and at least one subtask (IFEval no-comma) is a genuine counterexample to the
aligned pattern. The two measurement disciplines that
Section~\ref{sec:results:realization} states in short form, and that are
enforced throughout the paper, come from this observation. First, this receipt exists only in per-prompt
horizon-aligned coordinates: the \emph{unaligned} pooled switch curve has no
plateau at all, because averaging over a distribution of horizons smears the
transition, so no plateau, onset or located-transition claim anywhere in this
paper is read off an unaligned aggregate. Unaligned curves appear only where
the quantity being reported is a \emph{level} and the horizon enters as a
distribution rather than as a located point
(Figure~\ref{fig:gate-k-abs-sr}), and their captions say so. Second, compute
is reported descriptively as model forward evaluations; we make no claim
about wall-clock behaviour and none about a net gain over existing
schedulers.

% GATE x K full grid LANDED 2026-07-28 (260/260 elements, shared-prefix
% design, 2600 prompts x 30 arms). NOTE: it does NOT supersede
% tab_parallel_gate -- the +7.8 receipt above is a K=8 vs K=1 same-gate
% difference, and K=1 is outside this grid by design (serial, not parallel
% decoding). The two are complementary: the earlier sweep holds the serial
% control, this grid holds the width axis, PPL, all 13 subtasks and the
% censored prompts. Both stay.

% Gate x K, both readings, moved here from the census table dump 2026-07-28
% so that the tables sit with the prose that reads them. The aligned pair
% carries the offset claims, the absolute pair carries the levels and the
% censored prompts, and neither substitutes for the other.
% Auto-generated by analysis_utilities/injection_sharpening/make_gate_k_shared_tables.py
% from results/parallel_gate/*_shared_s*.json -- do not edit by hand.
\begin{table*}[t]
\centering
\small
\setlength{\tabcolsep}{4pt}
\begin{tabular}{l *{9}{r}}
\toprule
\multicolumn{10}{l}{\emph{(a) Success rate at gate $a^{*}{+}d$, aligned to each
prompt's own $a^{*}$ (offset $d$ in gate-grid units).}}\\
Arm & $-0.2$ & $-0.1$ & $0$ & $+0.1$ & $+0.2$ & $+0.3$ & $+0.4$ & $+0.5$ & $+0.6$ \\
\midrule
$K{=}4$ & 0.554 & 0.630 & 0.808 & 0.871 & 0.889 & 0.904 & 0.922 & 0.929 & 0.933 \\
$K{=}8$ & 0.558 & 0.618 & 0.791 & 0.853 & 0.880 & 0.898 & 0.916 & 0.925 & 0.931 \\
$K{=}16$ & 0.495 & 0.562 & 0.747 & 0.826 & 0.856 & 0.881 & 0.906 & 0.919 & 0.928 \\
\addlinespace
\multicolumn{10}{l}{\emph{(b) Parallelism-specific cost: paired
$K$ vs.\ $K{=}4$ at the same gate and prompt (SR points, mean $\pm$ SE over prompts).}}\\
Arm & $-0.2$ & $-0.1$ & $0$ & $+0.1$ & $+0.2$ & $+0.3$ & $+0.4$ & $+0.5$ & $+0.6$ \\
\midrule
$K{=}8$ & $+0.4\!\pm\!0.6$ & $-1.3\!\pm\!0.5$ & $-1.7\!\pm\!0.3$ & $-1.8\!\pm\!0.2$ & $-0.9\!\pm\!0.2$ & $-0.6\!\pm\!0.2$ & $-0.6\!\pm\!0.1$ & $-0.4\!\pm\!0.1$ & $-0.2\!\pm\!0.1$ \\
$K{=}16$ & $-6.0\!\pm\!0.8$ & $-6.8\!\pm\!0.8$ & $-6.1\!\pm\!0.5$ & $-4.5\!\pm\!0.4$ & $-3.3\!\pm\!0.3$ & $-2.3\!\pm\!0.3$ & $-1.6\!\pm\!0.2$ & $-1.1\!\pm\!0.2$ & $-0.5\!\pm\!0.2$ \\
\addlinespace
\multicolumn{10}{l}{\emph{(c) GPT-2 perplexity (fluency proxy, separate axis),
median over prompts.}}\\
Arm & $-0.2$ & $-0.1$ & $0$ & $+0.1$ & $+0.2$ & $+0.3$ & $+0.4$ & $+0.5$ & $+0.6$ \\
\midrule
$K{=}4$ & 56 & 51 & 53 & 57 & 60 & 61 & 63 & 63 & 64 \\
$K{=}8$ & 107 & 84 & 80 & 87 & 87 & 86 & 86 & 81 & 81 \\
$K{=}16$ & 195 & 180 & 158 & 166 & 156 & 146 & 143 & 119 & 116 \\
\midrule
prompts & 797 & 905 & 905 & 905 & 905 & 905 & 770 & 770 & 587 \\
\bottomrule
\end{tabular}
\caption{How far past the commitment horizon parallel filling stays costly, as a function of
the parallel width $K$. Shared-prefix grid: 13 subtasks, $n{=}200$ prompts
each (2600 total), gates $t\in\{0,0.1,\dots,0.9\}$ crossed with
$K\in\{4, 8, 16\}$; every arm is serial CFG ($w{=}2$) up to the
gate and guidance-free parallel filling at $K$ tokens/step afterwards, and all $30$ arms of a
prompt share one serial prefix, so gates and widths are paired within prompt. Alignment needs
a switch point, so the 905 handoff prompts carry the analysis and the
1695 censored prompts ($\hat a^{*}{=}T$, no switch) are excluded and
counted, never silently dropped. The pooled unaligned curve has no plateau (prompts have
different $\hat a^{*}$); all entries are therefore offset-aligned, and $d$ is a distance
from $\hat a^{*}$, not a claim about where $\hat a^{*}$ sits. \emph{Block (b)} is the
receipt: $K{=}1$ is outside this grid by design (serial, not parallel decoding), so the cost
of widening is measured against the $K{=}4$ arm of the same prompt at the same gate.
It is graded in $K$ and decays with offset --- $K{=}8$ never exceeds
$1.8$ points anywhere from $d{=}-0.2$ on, whereas $K{=}16$
still costs $6.1$ points at the horizon itself, $3.3$ two grid
steps later, and first falls within $2$ points at $d{=}+0.4$
(against $d{=}-0.2$ for $K{=}8$). Perplexity behaves differently again: at
$d{=}+0.4$ the $K{=}16$ success cost is down to
$1.6$ points while its fluency proxy is still
143 against 63 at $K{=}4$ --- the
horizon licenses parallel \emph{success}, not parallel \emph{fluency}, and wider steps
separate the two axes further. Free replication: the $K{=}8$ columns and the independently
drawn absolute-gate sweep agree to
$+0.0001$ $[-0.0008,
+0.0010]$ over 26000 paired cells.}
\label{tab:gate-k-shared}
\end{table*}

% Auto-generated by analysis_utilities/injection_sharpening/make_gate_k_shared_tables.py
% from results/parallel_gate/*_shared_s*.json -- do not edit by hand.
\begin{table*}[t]
\centering
\small
\setlength{\tabcolsep}{4pt}
\begin{tabular}{l r cc rrr c}
\toprule
 & & \multicolumn{2}{c}{Cost closes at} & \multicolumn{3}{c}{Residual $R_K$ (pts)} & \\
\cmidrule(lr){3-4} \cmidrule(lr){5-7}
Subtask & $n$ & $K{=}8$ & $K{=}16$ & $K{=}4$ & $K{=}8$ & $K{=}16$ & $R_{16}-R_{4}$ \\
\midrule
CommonGen kw $k{=}3$ & 78 & $+0.2$ & $+0.5$ & +8.5 & +10.7 & +14.9 & $+6.4$ $[+2.9,+10.1]$ \\
CommonGen kw $k{=}4$ & 80 & $+0.2$ & $+0.5$ & +6.7 & +10.1 & +14.3 & $+7.6$ $[+4.3,+11.0]$ \\
CommonGen kw $k{=}5$ & 87 & $+0.1$ & $+0.5$ & +9.1 & +10.8 & +14.7 & $+5.6$ $[+2.9,+8.8]$ \\
CommonGen kw $k{=}10$ & 108 & $+0.2$ & $+0.5$ & +8.9 & +11.7 & +18.2 & $+9.3$ $[+6.2,+12.5]$ \\
\addlinespace
CCG length & 84 & $0$ & $0$ & +2.9 & +3.8 & +2.0 & $-0.8$ $[-2.7,+1.0]$ \\
CCG sentiment & 49 & $+0.5$ & $+0.6$ & +7.0 & +8.9 & +11.9 & $+4.9$ $[-0.3,+9.8]$ \\
CCG multi & 129 & $+0.1$ & $+0.3$ & +6.1 & +7.8 & +11.7 & $+5.5$ $[+3.5,+7.8]$ \\
\addlinespace
IFEval kw-freq & 58 & $0$ & $0$ & +4.1 & +5.2 & +4.5 & $+0.4$ $[-1.2,+1.9]$ \\
IFEval letter-freq & 42 & $+0.2$ & $-0.1$ & +0.6 & +3.0 & +2.6 & $+2.0$ $[+0.2,+4.8]$ \\
IFEval capital-freq & 35 & $+0.2$ & $+0.3$ & +11.0 & +13.0 & +11.5 & $+0.6$ $[-2.4,+3.6]$ \\
IFEval no-comma & 147 & $0$ & $+0.3$ & +1.4 & +1.4 & +3.3 & $+1.9$ $[+0.6,+3.3]$ \\
\addlinespace
COLLIE c07 & 8 & -- & -- & -- & -- & -- & -- \\
COLLIE c06a & 0 & -- & -- & -- & -- & -- & -- \\
\midrule
pooled & 905 & $-0.2$ & $+0.4$ & +5.7 & +7.4 & +10.0 & $+4.3$ $[+3.5,+5.1]$ \\
\bottomrule
\end{tabular}
\caption{Per-subtask breakdown of the same grid: the width-dependence is family-structured,
not a uniform property of parallel filling. $n$ is the number of handoff prompts (the
aligned population; COLLIE c06a carries none and c07 eight, below the reporting floor of
30). \emph{Cost closes at} is the earliest offset from which the paired
$K$-vs-$K{=}4$ gap of Table~\ref{tab:gate-k-shared}(b) stays within
$2$ points at that offset and every later one. \emph{Residual}
$R_K = \mathrm{SR}_K(t{=}0.9) - \mathrm{SR}_K(\hat a^{*}{+}0.1)$ is the success still
unclaimed one grid step past the horizon, computed per prompt and then averaged, so the
population is fixed across $K$ and no offset-dependent censoring enters; the last column is
its within-prompt $K{=}16$ vs.\ $K{=}4$ difference with a $95\%$ bootstrap interval
over prompts. Widening from $K{=}4$ to $K{=}16$ leaves more success unclaimed past
the horizon on 7 of 11 subtasks with a resolvable estimate, but the
sizes separate sharply: 5 of them carry $+5.5$ to $+9.3$ points
(CommonGen kw k=3, CommonGen kw k=4, CommonGen kw k=5, CommonGen kw k=10, CCG multi), 2 are positive but at most
$+2.0$ points, and 4 have intervals covering zero
(CCG length, CCG sentiment, IFEval kw-freq, IFEval capital-freq). Read with Table~\ref{tab:gate-k-shared}: the
handoff licence and the parallel-safety licence are separable, and the distance between them
is a property of the constraint family as well as of $K$.}
\label{tab:gate-k-bytask}
\end{table*}

\paragraph{How the cost depends on the width.} The sweep above fixes $K{=}8$;
widening the parallel step is a second axis, and the two licences need not
move together. We therefore cross the gate with $K \in \{4,8,16\}$ over all
13 subtasks and all $200$ prompts each ($2600$ prompts $\times$ $30$ arms),
decoding one serial prefix per prompt and branching every gate and width off
it, so gates and widths are paired within prompt
(Table~\ref{tab:gate-k-shared}; per-subtask breakdown in
Table~\ref{tab:gate-k-bytask}). The cost of widening, measured against the
$K{=}4$ arm of the same prompt at the same gate, is graded in $K$ and decays
with offset: $K{=}8$ never exceeds $1.8$ points anywhere from $d{=}-0.2$ on,
whereas $K{=}16$ still costs $6.1$ points at the horizon itself, $3.3$ two
grid steps later, and first falls within $2$ points at $d{=}+0.4$. The
independently drawn absolute-gate sweep replicates the $K{=}8$ columns to
$+0.0001$ $[-0.0008, +0.0010]$ over $26000$ paired cells.

Success and fluency separate along this axis, and wider steps separate them
further. At $d{=}+0.4$ the $K{=}16$ success cost is down to $1.6$ points
while its perplexity is still $143$ against $63$ at $K{=}4$; pooled on the
absolute axis the same asymmetry is visible without alignment, since at
$t{=}0.7$ the $K{=}4$--$K{=}16$ success spread has closed to about a point
while the perplexity ratio is still $2.2\times$, and it is $1.2\times$ only
at $t{=}0.9$. The horizon licenses parallel \emph{success}; it does not
license parallel \emph{fluency}, and the two costs vanish at different
places.

% Generated by analysis_utilities/injection_sharpening/fig_gate_k_abs.py
% (figs/fig_gate_k_abs_sr.pdf).  Main-text figure for Sec 5.6.2.  Its PPL
% companion is fig_gate_k_abs_ppl (appendix), and the numbers behind both are
% tab_gate_k_abs_sr / tab_gate_k_abs_ppl (appendix).

\begin{table*}[t]
\centering
\footnotesize
\setlength{\tabcolsep}{2pt}
\begin{tabular}{l r cccccccccc | c}
\toprule
Task & cens. & $0.0$ & $0.1$ & $0.2$ & $0.3$ & $0.4$ & $0.5$ & $0.6$ & $0.7$ & $0.8$ & $0.9$ & $1.0$ \\
 & \% & \multicolumn{10}{c}{gate $t$ \quad (cell: $K{=}4/8/16$)} & serial \\
\midrule
Keywords-10 & 46 & 31/15/5 & 46/43/30 & 47/46/39 & \cellcolor{black!10}53/50/41 & \cellcolor{black!10}60/56/51 & 63/60/54 & 65/64/61 & 68/67/66 & 70/69/68 & 72/72/72 & 74 \\
CommonGen $k3$ & 61 & 64/51/37 & 66/64/56 & \cellcolor{black!10}68/67/63 & \cellcolor{black!10}72/69/63 & \cellcolor{black!10}73/71/69 & 74/71/66 & 77/76/74 & 79/79/78 & 82/82/80 & 83/83/83 & 84 \\
CommonGen $k4$ & 60 & 49/38/26 & 55/51/45 & \cellcolor{black!10}60/57/54 & \cellcolor{black!10}63/60/54 & \cellcolor{black!10}63/61/58 & 64/62/57 & 67/66/64 & 72/71/70 & 73/73/70 & 74/74/74 & 74 \\
CommonGen $k5$ & 56 & 32/26/15 & 39/37/31 & \cellcolor{black!10}42/40/38 & \cellcolor{black!10}47/45/39 & \cellcolor{black!10}49/47/47 & 50/49/46 & 54/54/52 & 58/58/57 & 60/59/58 & 61/61/61 & 62 \\
Multi-constr. & 36 & 49/30/13 & 62/57/43 & \cellcolor{black!10}64/62/55 & \cellcolor{black!10}70/66/56 & \cellcolor{black!10}74/71/69 & 78/76/73 & 82/81/80 & 84/83/83 & 84/85/84 & 85/85/85 & 87 \\
Length & 58 & 45/52/60 & 59/63/66 & \cellcolor{black!10}71/72/75 & \cellcolor{black!10}73/74/76 & 76/77/78 & 77/77/80 & 77/78/79 & 78/79/79 & 80/80/80 & 82/82/82 & 81 \\
Sentiment & 76 & 57/57/54 & \cellcolor{black!10}61/60/59 & \cellcolor{black!10}61/60/56 & \cellcolor{black!10}62/60/56 & \cellcolor{black!10}64/62/59 & 63/61/58 & 63/60/59 & 61/61/60 & 62/62/61 & 63/63/63 & 64 \\
IFE kw-freq & 71 & 87/87/87 & 88/89/88 & \cellcolor{black!10}92/93/93 & \cellcolor{black!10}94/94/93 & \cellcolor{black!10}95/95/94 & 96/95/95 & 95/95/95 & 95/95/95 & 94/94/94 & 94/94/94 & 93 \\
IFE letter & 79 & 66/67/67 & 61/64/64 & \cellcolor{black!10}68/69/70 & \cellcolor{black!10}69/70/70 & 70/70/71 & 70/70/71 & 70/70/70 & 70/70/70 & 70/70/70 & 70/70/70 & 70 \\
IFE cap-freq & 82 & 47/47/47 & 59/56/57 & \cellcolor{black!10}60/59/60 & \cellcolor{black!10}62/61/61 & \cellcolor{black!10}65/63/64 & \cellcolor{black!10}67/66/65 & \cellcolor{black!10}68/67/68 & 69/69/70 & 70/71/70 & 72/72/72 & 73 \\
IFE no-comma & 26 & 71/84/68 & 49/55/47 & \cellcolor{black!10}67/69/64 & \cellcolor{black!10}75/76/72 & \cellcolor{black!10}78/78/74 & 79/78/76 & 79/79/77 & 80/79/78 & 80/79/79 & 80/80/80 & 80 \\
COLLIE c07 & 96 & 96/91/80 & 98/95/91 & 97/97/94 & \cellcolor{black!10}97/96/95 & 96/96/96 & 97/96/95 & 97/97/96 & 95/96/95 & 96/95/95 & 96/96/96 & 96 \\
COLLIE c06a & 100 & 5/5/6 & 7/11/11 & 13/13/14 & 7/7/6 & 4/4/5 & 6/4/3 & 1/2/2 & 5/5/2 & 5/4/2 & 4/3/3 & 4 \\
\midrule
\emph{pooled} & 65 & 54/50/44 & 58/57/53 & \cellcolor{black!10}62/62/60 & \cellcolor{black!10}65/64/60 & \cellcolor{black!10}67/66/64 & 68/67/64 & 69/68/67 & 70/70/69 & 71/71/70 & 72/72/72 & 72 \\
\bottomrule
\end{tabular}
\caption{Success rate ($\times 100$) on the absolute gate axis. Serial CFG ($w{=}2$, 1 tok/step) until gate $t$, then guidance-free parallel filling at $K$ tokens/step; $n{=}200$ census prompts per subtask, 2600 total, nothing excluded.  The $t{=}1.0$ column is full serial CFG: no parallel segment survives, so it is $K$-independent and holds a single number.  Shading marks the interquartile band of the per-prompt $\hat{a}^{*}$ over the subtask's handoff prompts; ``cens.'' is the percentage with $\hat{a}^{*}{=}T$, which have no $\hat{a}^{*}$ but do have curves.  $\hat{a}^{*}$ was resolved on the census switch grid $\{.10,.15,.22,.30,.45,.60\}$, so no band extends right of $0.6$ and the $0.6$ bin is a pile-up at that cap.  \textbf{A row is a pooled unaligned curve} and smears the per-prompt onsets: read levels here, never a plateau -- plateau claims rest on the $\hat{a}^{*}$-aligned increments (Tab.~\ref{tab:gate-k-shared}).}
\label{tab:gate-k-abs-sr}
\end{table*}

\begin{sidewaystable}[t]
\centering
\footnotesize
\setlength{\tabcolsep}{2pt}
\begin{tabular}{l r cccccccccc | c}
\toprule
Task & cens. & $0.0$ & $0.1$ & $0.2$ & $0.3$ & $0.4$ & $0.5$ & $0.6$ & $0.7$ & $0.8$ & $0.9$ & $1.0$ \\
 & \% & \multicolumn{10}{c}{gate $t$ \quad (cell: $K{=}4/8/16$)} & serial \\
\midrule
Keywords-10 & 46 & 102/242/384 & 43/67/120 & 44/68/136 & \cellcolor{black!10}45/68/143 & \cellcolor{black!10}48/69/121 & 51/77/145 & 49/67/107 & 52/64/95 & 52/63/91 & 53/61/62 & 51 \\
CommonGen $k3$ & 61 & 105/281/693 & 59/114/273 & \cellcolor{black!10}54/87/199 & \cellcolor{black!10}53/86/214 & \cellcolor{black!10}55/93/176 & 59/101/240 & 55/83/160 & 58/78/138 & 55/76/136 & 53/66/66 & 49 \\
CommonGen $k4$ & 60 & 95/242/669 & 53/94/213 & \cellcolor{black!10}54/83/178 & \cellcolor{black!10}53/84/197 & \cellcolor{black!10}54/90/161 & 61/99/227 & 56/84/161 & 59/78/133 & 57/78/131 & 54/68/69 & 49 \\
CommonGen $k5$ & 56 & 80/201/549 & 50/88/200 & \cellcolor{black!10}52/81/171 & \cellcolor{black!10}51/84/190 & \cellcolor{black!10}53/89/151 & 57/95/202 & 53/80/137 & 54/71/124 & 54/74/123 & 54/67/66 & 51 \\
Multi-constr. & 36 & 184/400/811 & 71/121/248 & \cellcolor{black!10}66/103/217 & \cellcolor{black!10}60/98/222 & \cellcolor{black!10}61/95/175 & 64/99/210 & 61/82/145 & 61/77/121 & 59/74/120 & 56/67/67 & 54 \\
Length & 58 & 354/709/2.0k & 216/314/577 & \cellcolor{black!10}188/221/343 & \cellcolor{black!10}188/221/346 & 171/234/233 & 200/255/328 & 169/211/285 & 178/202/208 & 165/185/234 & 180/188/194 & 251 \\
Sentiment & 76 & 175/487/2.4k & \cellcolor{black!10}60/74/154 & \cellcolor{black!10}45/63/118 & \cellcolor{black!10}43/64/132 & \cellcolor{black!10}46/63/101 & 51/68/121 & 50/60/92 & 55/61/82 & 52/61/83 & 55/61/61 & 54 \\
IFE kw-freq & 71 & 36/69/115 & 33/51/96 & \cellcolor{black!10}35/55/119 & \cellcolor{black!10}35/57/138 & \cellcolor{black!10}37/62/116 & 40/64/140 & 38/57/106 & 39/53/90 & 39/51/80 & 39/45/45 & 35 \\
IFE letter & 79 & 37/81/136 & 34/53/99 & \cellcolor{black!10}40/62/137 & \cellcolor{black!10}45/73/166 & 50/81/155 & 56/90/202 & 55/81/159 & 61/83/146 & 62/82/125 & 61/72/71 & 57 \\
IFE cap-freq & 82 & 41/100/159 & 39/60/123 & \cellcolor{black!10}47/68/148 & \cellcolor{black!10}52/81/180 & \cellcolor{black!10}57/94/164 & \cellcolor{black!10}64/101/212 & \cellcolor{black!10}61/91/169 & 64/87/151 & 63/83/132 & 61/74/75 & 54 \\
IFE no-comma & 26 & 38/144/175 & 35/51/99 & \cellcolor{black!10}48/71/156 & \cellcolor{black!10}56/88/201 & \cellcolor{black!10}65/101/198 & 73/120/248 & 72/106/208 & 76/105/179 & 77/102/156 & 73/89/89 & 67 \\
COLLIE c07 & 96 & 191/556/1.1k & 119/265/702 & 112/169/404 & \cellcolor{black!10}93/173/485 & 70/158/352 & 74/164/500 & 62/120/288 & 70/106/232 & 61/96/203 & 62/85/82 & 55 \\
COLLIE c06a & 100 & 3.1k/5.3k/8.0k & 83/195/556 & 85/171/624 & 89/245/632 & 89/252/807 & 103/273/1.1k & 121/263/819 & 152/273/627 & 178/287/509 & 190/239/241 & 155 \\
\midrule
\emph{pooled} & 65 & 98/265/604 & 48/77/164 & \cellcolor{black!10}53/80/171 & \cellcolor{black!10}54/87/196 & \cellcolor{black!10}58/93/168 & 64/103/213 & 61/87/159 & 63/84/141 & 60/81/132 & 59/72/72 & 55 \\
\bottomrule
\end{tabular}
\caption{GPT-2 perplexity (median over prompts) on the absolute gate axis, companion to Tab.~\ref{tab:gate-k-abs-sr}. Serial CFG ($w{=}2$, 1 tok/step) until gate $t$, then guidance-free parallel filling at $K$ tokens/step; $n{=}200$ census prompts per subtask, 2600 total, nothing excluded.  The $t{=}1.0$ column is full serial CFG: no parallel segment survives, so it is $K$-independent and holds a single number.  Shading marks the interquartile band of the per-prompt $\hat{a}^{*}$ over the subtask's handoff prompts; ``cens.'' is the percentage with $\hat{a}^{*}{=}T$, which have no $\hat{a}^{*}$ but do have curves.  $\hat{a}^{*}$ was resolved on the census switch grid $\{.10,.15,.22,.30,.45,.60\}$, so no band extends right of $0.6$ and the $0.6$ bin is a pile-up at that cap.  \textbf{A row is a pooled unaligned curve} and smears the per-prompt onsets: read levels here, never a plateau -- plateau claims rest on the $\hat{a}^{*}$-aligned increments (Tab.~\ref{tab:gate-k-shared}).}
\label{tab:gate-k-abs-ppl}
\end{sidewaystable}

% Generated by analysis_utilities/injection_sharpening/fig_gate_k_abs.py
% (figs/fig_gate_k_abs_ppl.pdf).  Main-text companion to fig_gate_k_abs_sr,
% both in Sec 5.6.2.  Separate figure, never a twin axis on the SR panels: the
% two quantities are on different scales and answer different questions.
% Two full-width figure* floats in one subsubsection is a real space cost --
% if the page budget bites, this is the one to move to the appendix (it is a
% \input, so it moves in one line and its caption does not assume a location).
\begin{figure*}[t]
\centering
\includegraphics[width=\linewidth]{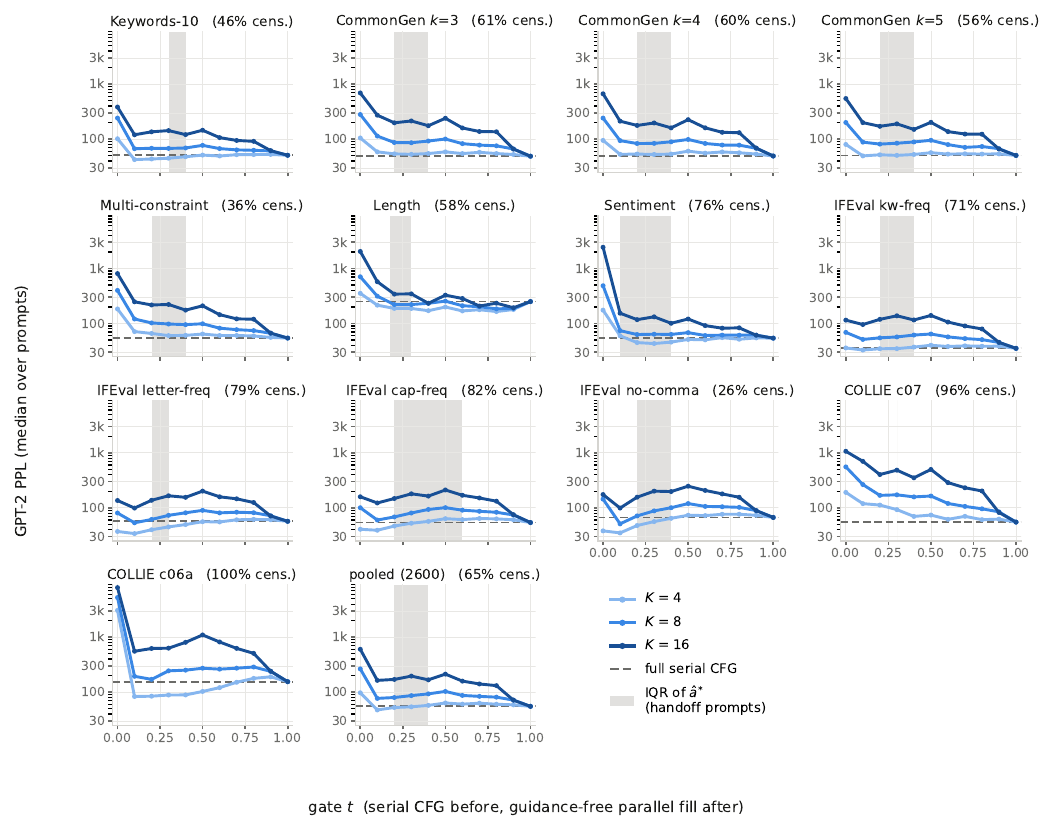}
\caption{Fluency companion to Figure~\ref{fig:gate-k-abs-sr}: GPT-2 perplexity (median over
prompts, log axis) on the same gate $\times$ width grid, same $2600$ prompts, same arms and
same conventions. Perplexity is a fluency proxy on a separate axis from success and is never
combined with it. The panels make the asymmetry quoted in Section~5.6.2 visible across
subtasks: the widths converge in success well before they converge in perplexity, so the
width at which handing off is free for one criterion is not the width at which it is free
for the other. As in Figure~\ref{fig:gate-k-abs-sr} the curves are pooled and unaligned and
are read as levels only.}
\label{fig:gate-k-abs-ppl}

\end{figure*}

Figure~\ref{fig:gate-k-abs-sr} shows the whole grid on the absolute gate
axis, one panel per subtask. It is the one view here that carries the
censored prompts: the gate schedule is prompt-independent, so $a^{*}$ is
not in the design and the $1695$ prompts with $a^{*}{=}T$ have complete
curves even though no aligned analysis can use them. Read as levels, the
panels show how unevenly the width axis bites---three lines lie on top of one
another for IFEval letter-frequency and COLLIE c07, and open by more than
twenty points at low $t$ for Keywords-10 and the CommonGen counting
family---and COLLIE c06a, censored for every one of its $200$ prompts, stays
near zero at every gate and width, which is what the hopeless fate looks like
from this side. The shaded band is the interquartile range of $a^{*}$
over the subtask's handoff prompts, not a located transition; it is drawn as
a band because every task-level scalar we tried fails, including the two that
look most defensible. The median $a^{*}$ over all prompts is $1.0$ in
$11$ of $13$ subtasks under censoring, and the earliest gate whose paired
deficit against full serial CFG clears a $0.03$ non-inferiority margin lands
at $t \ge 0.8$ in $10$ of $13$ subtasks and barely moves with $K$. That is
not an artefact of those estimators: a curve here is a mixture over prompts
with different horizons, and a mixture reaches parity only when its slowest
component does, which is the same reason the unaligned curve has no plateau.
The horizon is not recoverable from the aggregate that averages over it.
Figure~\ref{fig:gate-k-abs-ppl} reads the same grid on the fluency axis,
where the widths converge visibly later than they do in success.

\subsection{The Predicted Gate}
\label{app:cheap-gate}

This subsection backs Section~\ref{sec:results:predicted}, which quotes
only the parity contrast and the two qualifications. Table~\ref{tab:cheap-gate}
is the receipt; the accuracy metrics, the policy rule, the sentiment exception
and the matched-budget analysis are below. The comparator discipline this leg
taught us is stated once and applies to the whole block: judged against a
per-subtask constant the predicted gate looks like a failure, judged against
the oracle---the actual deployment question, can a cheap gate replace the one
we paid rollouts for?---it does not. Both are reported here, each with the
question it answers.

% tab_cheap_gate MOVED to Sec 5.4 (2026-07-28): the body cites it for every
% number in that block, so the receipt now sits with the claim. Do not re-input
% it here -- duplicate label. tab_astar_transfer arrived here from Sec 6.2 in
% the same pass: it varies only how the switch time is CHOSEN (classifier /
% family cut / oracle), which is this block's question.
% Auto-generated by analysis_utilities/injection_sharpening/make_astar_transfer_table.py
% from results/q3_classifier/Q3_RESULTS_13{,_POOLED}.json -- do not edit by hand.
\begin{table*}[t]
\centering
\small
\begin{tabular}{l r c c c c c c}
\toprule
 & & \multicolumn{3}{c}{Success rate} & \multicolumn{3}{c}{CFG-on fraction} \\
Constraint & $n$ & Classifier & Family cut & Oracle $a^{*}$ & Classifier & Family cut & Oracle $a^{*}$ \\
\midrule
CommonGen kw $k{=}3$ & 78 & 0.75 & 0.75 & 0.74 & 0.58 & 0.60 & 0.32 \\
CommonGen kw $k{=}4$ & 80 & 0.69 & 0.69 & 0.69 & 0.57 & 0.60 & 0.36 \\
CommonGen kw $k{=}5$ & 87 & 0.53 & 0.54 & 0.54 & 0.57 & 0.60 & 0.44 \\
CommonGen kw $k{=}10$ & 108 & 0.62 & 0.62 & 0.62 & 0.56 & 0.60 & 0.47 \\
CCG length & 84 & 0.76 & 0.76 & 0.76 & 0.59 & 0.60 & 0.30 \\
CCG sentiment & 49 & 0.66 & 0.66 & 0.72 & 0.15 & 0.15 & 0.17 \\
CCG multi & 129 & 0.82 & 0.83 & 0.82 & 0.56 & 0.60 & 0.35 \\
IFEval kw-freq & 58 & 0.95 & 0.96 & 0.97 & 0.39 & 0.45 & 0.18 \\
IFEval letter-freq & 42 & 0.69 & 0.70 & 0.69 & 0.52 & 0.60 & 0.28 \\
IFEval capital-freq & 35 & 0.71 & 0.71 & 0.70 & 0.56 & 0.60 & 0.31 \\
IFEval no-comma & 147 & 0.79 & 0.80 & 0.78 & 0.55 & 0.60 & 0.37 \\
COLLIE c07 & 8 & 0.98 & 0.98 & 0.98 & 0.07 & 0.10 & 0.07 \\
COLLIE c06a & -- & 0.10 & 0.10 & 0.10 & 0.30 & 0.30 & 0.30 \\
\midrule
Pooled & 905 & 0.73 & 0.74 & 0.73 & 0.52 & 0.56 & 0.34 \\
\bottomrule
\end{tabular}
\caption{Deploying the horizon on unseen prompts: three gate sources on the
handoff prompts of each family ($n$ per row), evaluated on held-out census
rollouts with the same freeze-branch grid. \emph{Family cut} is a single
train-derived switch time per family; \emph{classifier} predicts a
per-prompt tercile from rollout-free trajectory features of the prompt's own
run (confidence/entropy channel statistics and commit-order shape) and
deviates from the family cut only when its fate head is confident;
\emph{oracle} switches at the prompt's measured $a^{*}$ (ceiling). The
classifier and the family cut agree on success within one point on every
family, and the oracle adds success only on sentiment ($+0.06$, the overshoot
family where a late per-prompt horizon needs a later cut): the horizon's value
survives coarse deployment, and per-prompt prediction buys only a modestly
smaller CFG-on fraction (pooled $0.52$ versus $0.56$;
oracle headroom $0.34$). The features do not transfer across
families: under leave-one-task-out training the tercile accuracy drops from
$0.46$ (within-family) to $0.38$ (chance
$\approx 0.33$), so the gate source stays family-calibrated. Cross-trajectory
transfer of the \emph{recorded} per-prompt horizon to fresh rollouts is
Table~\ref{tab:policy-quant}; this table varies only how the switch time is
chosen. Not packaged as a decoding method.}
\label{tab:astar-transfer}
\end{table*}

\textbf{Design.} We fit a random forest on rollout-free features of a prompt's
own recorded run (confidence and entropy channel statistics and commit-order
shape), in a fate head and an ordinal head. The ordinal head regresses the
horizon and bins the prediction into terciles rather than classifying the bin
directly, which respects both the ordering of the bins and the noise in the
labels. Training is within-subtask with $\mathrm{GroupKFold}(5)$ grouped on
prompt index, so a test prompt is never seen in training; both heads are
$400$-tree forests with a minimum leaf size of $5$ and a fixed seed. Labels
come from the census, and because the recorded horizon is stable in the
guidance weight over this wave, a prompt contributes its runs at all eight
weights of the census grid ($w$ from $0.5$ to $4$) to the training pool. The
grouping keeps a prompt's weight variants inside one fold. Two constants are fixed in
advance and are the only tuned quantities: the fate-head confidence
$\tau{=}0.6$ at which the policy may depart from the constant cut, and a
$+0.02$ shift added to the regressed horizon before binning, which biases the
gate late on the grounds that releasing guidance early is the costlier error.
Each gate source is then scored by reading, from the gate-by-$K$ sweep, the arm
it would have selected on fresh trajectories, paired prompt by prompt: the
\emph{oracle} (the recorded $\astar$), the \emph{raw} predictor $\hat a^{*}$, a
deployable \emph{policy} (default to a per-subtask constant cut, deviate to the
upper edge of the predicted tercile only where the fate head is confident), and
the \emph{family cut} alone. The comparison that matters is the one against the
oracle: \emph{can a cheap gate replace the one we paid rollouts for?}

\textbf{Finding 1: the predictor is inaccurate by every accuracy measure.}
Within-subtask tercile accuracy averages $0.449$ over the 12 subtasks with
survivors against a chance rate of $0.333$, and horizon MAE is
$0.075$--$0.184$---pooled, $0.136$ against $0.139$ for a degenerate baseline
that predicts zero for every prompt. Under leave-one-task-out training the
pipeline collapses to $0.333$, exactly chance, so the features are
family-specific and the gate source must stay family-calibrated. We therefore
say the predictor \emph{orders the horizon's tercile above chance within a
subtask}, and never that it predicts $\astar$.

\textbf{Finding 2: accuracy is not what the gate needs.} Read against the
oracle at $K{=}8$, the raw predictor---the same one that barely beats
predicting zero---induces a gate whose success is statistically
\emph{indistinguishable} from the rollout-measured horizon: $+0.015$
$[-0.001, +0.031]$ pooled over $905$ survivors, at $46.5$ forward evaluations
per sequence against the oracle's $43.1$, an $8\%$ difference. That parity is
the claim. The deployable policy scores higher still ($+0.092$
$[+0.075, +0.108]$ pooled, significant on 8 of the 12 subtasks with survivors,
and the same ordering holds intention-to-treat over $1826$ gated prompts), but
it gates later and spends $64.6$ forwards, half as much again, so part of that
margin is bought with guided steps rather than won by prediction. We rest the
block on parity, not on the excess: what a cheap gate demonstrates is that the
rollout-measured horizon can be \emph{replaced} at matched cost, not that
predicting it improves on it. A quantity can be worth defining, and worth
measuring once, without being worth predicting precisely.

\textbf{Finding 3: the exception identifies where per-prompt information is
load-bearing.} CCG sentiment inverts the pattern: there the policy is
significantly \emph{below} the oracle ($-0.082$ $[-0.144, -0.020]$ at $K{=}8$,
and negative at every $K$). Sentiment is precisely the overshoot family in
which handoff is \emph{strictly superior} to full CFG in
Table~\ref{tab:noninferiority}---the family where guidance past the horizon
actively hurts rather than merely failing to help. Where sustained guidance is
redundant, the switch curve is flat past the horizon and any reasonable gate
lands on the plateau; where it is harmful, the curve turns \emph{down},
flatness fails, and the per-prompt location is what a policy must get right.
This also explains what the policy is doing: it defaults to the constant cut
and deviates only when confident, so most of its advantage comes from the
constant, and it inherits the constant's error exactly where the constant is
wrong.

\textbf{Scope, stated plainly.} We are not proposing a decoder, and the
comparison against the other baseline runs the other way: per-prompt gating
does \emph{not} beat a per-subtask constant, and gating at the recorded
$\astar$ loses to a later constant on 10 of 12 subtasks ($-0.116$
$[-0.131, -0.101]$ pooled). That contrast does not cancel in the pool, so we
quote it pooled rather than stratified. But the constant does not
\emph{locate} the switch better---it switches later and buys success with
guided steps. It lands on the last point of the freeze grid on 9 of 13
subtasks, and read on the wider sweep that argmax is a \emph{corner}: the
paired $0.6 \rightarrow 0.9$ increment is significantly positive in 26 of 39
subtask-by-$K$ cells and moves the argmax to the new edge in 25 of them, at a
cost of $+35$ to $+39$ forward evaluations per sequence. Nor is the trade
evadable by parallelizing harder: holding the budget fixed and pairing arms
whose mean forward counts agree to within $5\%$, a rule that buys extra guided
steps by committing more positions per step loses success in 84 of 195 matched
pairs and wins in 17. So ``a later constant beats gating at $\astar$'' is a
statement about how many guided steps a deployment is willing to spend, not
about where the boundary is. Compute is reported throughout as model forward
evaluations, descriptively: we make no wall-clock claim and none of a net gain
over existing schedulers. The horizon is a boundary in the committor field;
nothing here proposes it as a tuned hyperparameter.

% ============================================================================
% Moved out of Sec 5.3 on 2026-07-28 (6th pass, user): the repair half was
% ~740 words in the body and is now ~170. Everything below came out of the
% body verbatim except the lead paragraph, which was compressed into the
% body's "Repair by reopening" run-in. The BODY still states both negative
% results and the artifact in one clause each -- this appendix carries the
% numbers, the eight cells, the selector-agreement rates and the audit.
% Do NOT let the two withdrawals get softened here: they are the paper's own
% retractions and Sec 6.1 dropped a receipt because of them.
% ============================================================================
\section{Repair by Reopening: Design, Cells, and the Length Audit}
\label{app:remask}

This appendix backs the ``Repair by reopening'' paragraph of
Section~\ref{sec:results:repair}, which states its findings without effect
sizes: the per-cell recovery numbers are here, as are both negative results
--- repair is \emph{not} type-specific, and the selection hierarchy does not
reproduce --- and the length-matching artifact whose correction reverses the
fluency reading. Section~\ref{sec:discussion} withdraws the two claims those
results overturn; this section is the evidence it withdraws them on. This leg is the one place in
the paper with no counterpart in the theory of Section~\ref{sec:transport}: it
is a descriptive characterization of the post-commitment field, not a
consequence of the transport law.

\textbf{Design.} We separate trajectories into \emph{hopeless} (the committor
is born on the floor and never lifts) and \emph{collapse} (it forms a peak and
then breaks). Reopening is the only repair available: under absorbing
unmasking a committed position is never re-decided, so in-place correction is
impossible by construction (Appendix~\ref{app:macrostate}). The intervention
population is the whole census: all $196$
collapse prompts plus $200$ sampled hopeless prompts, $396$ over $12$ subtasks.
Each prompt is intervened at two anchors---\emph{pre}, half-way from the start
to $\astar$, and \emph{post}, three steps past the committor peak---under two
seeds, at $K \in \{1,3\}$ reopened positions, with five selection signals
ordered by their committor content (a \emph{restart} arm reopening every
committed position is run but not reported; see the table caption). Every arm
resumes guided decoding over $10$ continuations; contrasts are paired within
prompt and bootstrapped with prompts as clusters.

\textbf{Finding 1: repair is broad, and not collapse-specific.} All five
selection signals beat the no-reopen control in all eight
(anchor $\times$ type $\times$ $K$) cells, with every interval excluding zero
(Table~\ref{tab:remask-recovery}).
At the pre anchor and $K{=}3$, reopening lifts success from $0.369$ to $0.545$
on collapse and from $0.091$ to $0.243$ on hopeless: the absolute recovery is
comparable across the two geometries ($+0.176$ against $+0.152$), and it is
positive in all $12$ subtasks. \emph{This overturns the double dissociation we
previously reported from a $10$-prompt single-carrier wave}---failure type is
not what discriminates the interventions---and the anchor gradient shrinks with
it ($+0.176$ against $+0.170$ at $K{=}3$), so the earlier ``pre-$\astar$ repair
is worth about twice post-peak repair'' does not hold at scale either.

\textbf{Finding 2: which positions are reopened barely matters.} The selection
contrast---full counterfactual committor ordering against commit
confidence---is indistinguishable from zero in six of the eight cells, and the
two cells that do reach significance point in \emph{opposite} directions
($-0.056$ $[-0.104,-0.007]$, confidence ahead, at the post anchor on collapse
at $K{=}3$; $+0.036$ $[+0.009,+0.065]$, committor ahead, at the post anchor on
hopeless at $K{=}1$). The ordering by committor content does not reproduce: the
rollout-free guidance-disagreement signal ranks \emph{first} at the pre anchor
on collapse at $K{=}3$ ($+0.219$), ahead of the full committor ($+0.176$), and
per subtask the contrast is positive in seven subtasks and negative in five.
This is a stronger statement than ``the cheap proxy is good enough'': at
$K{=}1$ the committor and confidence orderings pick the \emph{same} position
only $14$--$36\%$ of the time, so the signals genuinely disagree about where to
intervene and recover the same amount anyway. What carries the effect is that a
committed position is reopened at all, not which one. We therefore withdraw the
claim that rollout is structurally irreducible here.
% tab_remask_recovery MOVED to Sec 5.3 (2026-07-28, "Repair by reopening"):
% the SR+PPL receipt now sits in the body next to the parallel figure. Do not
% re-input it here -- duplicate label.

\textbf{A measurement artifact that had to be removed first.} The rollout that
scores an arm draws a fixed per-step token budget computed on a full generation
block, so it is blind to how many positions the arm actually reopened. For an
untouched snapshot the budget matches the mask count exactly (verified on all
$1584$ interventions), so the control fills its sequence; a reopened snapshot
carries $K$ extra masks against the same budget and ends with $K$ positions
still masked, which the detokenizer deletes silently. The arm's text is
therefore not short at the tail but \emph{punctured}, and the exposure is
unequal by construction. We re-ran every outcome arm with the budget repaired,
reusing the recorded selections; the control reproduces bit for bit
($3168/3168$), which certifies the two runs are otherwise identical. Repairing
it moves the reopen arms by $0.015$ on average and reverses the fluency reading
entirely: reopened continuations are \emph{no less} fluent than the control by
GPT-2 perplexity, and often more so ($183$ against $264$ at the pre anchor on
collapse at $K{=}3$). The pre-registered length check this leg committed to
is thus discharged, with the confound found, quantified and removed rather than
merely bounded.

\section{Constraint Families and Success Predicates}
\label{app:constraints}

Every predicate is deterministic and judge-free, evaluated on the terminal text $y$;
none uses an LLM judge. For the continuous committor used by the transport estimator
(Section~\ref{sec:theory:cov}) we replace the binary indicator by the graded family
value in brackets below, since the binary predicate gives $\sigma(q') = 0$ at the
band edges.

\paragraph{Keywords and keywords\_multi.} Given a required key set
$\{k_1, \dots, k_J\}$, each key is matched by a word-boundary, case-insensitive regular
expression ($\backslash\mathrm{b}\,k_j\,\backslash\mathrm{b}$) against $y$. Success
requires all $J$ present, $S(y) = \mathbf{1}\big[\sum_{j} \mathbf{1}\{k_j \in y\} =
J\big]$; \emph{keywords} uses small $J$ and \emph{keywords\_multi} uses $J \ge 2$
jointly. [Graded family: fraction $\ge m$ of $J$ present.] Prompts and key sets are
drawn from CommonGen.

\paragraph{Length.} With source word count $w \ge 1$ and a target ratio band
$[\ell_{\mathrm{lo}}, \ell_{\mathrm{hi}}]$ (default $[0.4, 0.8]$), let
$r(y) = |{\rm split}(y)| / w$. Success is $S(y) = \mathbf{1}[r(y) \in [\ell_{\mathrm{lo}},
\ell_{\mathrm{hi}}]]$. [Graded family: band relaxed by $\pm 0.1, \pm 0.2$.]

\paragraph{Sentiment.} A fixed SST-2 sentiment classifier returns
$P_{\mathrm{pos}}(y) \in [0,1]$; success is $S(y) = \mathbf{1}[P_{\mathrm{pos}}(y) \ge
0.5]$ for the positive target. [Graded family: thresholds $0.5, 0.8, 0.9$.] This is the
only family whose predicate is a learned classifier rather than a surface match, which
is the source of its "single stiff axis" geometry (Section~\ref{sec:results}).

\paragraph{Avoidance (Cognac, within-task banding leg).} A WordNet dual constraint:
$S(y) = \mathbf{1}[\text{on\_topic} \wedge \neg\,\text{violated}]$, where the topic and
the forbidden sub-branch each expand to their WordNet leaf sets (plus plurals) and are
matched as substrings in $y$. This exclusion predicate is the non-fabricable axis used
to isolate the $\sigma(q')$ (room) factor within a single task.

\section{\texorpdfstring{$\astar$}{a*} Is Relational: Single-State Signals Do
Not Decode It}
\label{app:relational}
\label{sec:discussion:relational}

The horizon is defined by comparing two continuations. The remaining guidance
value $\Vval_t = \qguid_t - \qbase_t$ is a \emph{difference of committors}, and
that relational character has an empirical consequence: it cannot be read off a
single state. A bake-off of cheap single-state proxies (KL divergence between
guided and base steps, predictive entropy, top-token confidence and margin,
together with a hidden-state probe and a paired-difference probe) fails to
decode $\astar$: none is monotone in the true handoff across families. The
estimators that do work are themselves \emph{paired}: they compare two
continuation policies rather than read one state. The freeze estimator of
Appendix~\ref{app:census} is the expensive instance, and a truncated version of
the same comparison is cheap enough to be practical: on that readout, t
$k{=}8$ truncated committor selects repair positions as well as the untruncated
one at roughly an eighth of its cost (Table~\ref{tab:remask-recovery}).
therefore report $\astar$ as a relational order parameter, a property of the
pair (guided dynamics, base dynamics) rather than of any instantaneous
of the state. The negative is scoped to the signals tested rather than asserted
as impossibility, and it is informative: it rules out the confidence-st
stopping rules a practitioner would reach for first, and it is why
Appendix~\ref{app:cheap-gate} reaches for a trained predictor rather th
signal. A third, interventional receipt we previously offered, namely that
repair success is ordered by the committor content of the signal choosi
targets, does \emph{not} replicate at census scale
(Appendix~\ref{app:remask}); the claim therefore rests on the bake-off
the structure of the definition, with no third receipt behind it.

\end{document}